\documentclass{article}

     \PassOptionsToPackage{numbers, compress}{natbib}

     \usepackage[preprint]{neurips_2022}

\usepackage[utf8]{inputenc} 
\usepackage[T1]{fontenc}    
\usepackage{hyperref}       
\usepackage{url}            
\usepackage{booktabs}       
\usepackage{amsfonts}       
\usepackage{nicefrac}       
\usepackage{microtype}      
\usepackage[table]{xcolor}         
\usepackage{caption}

\usepackage{multicol}

\usepackage{amssymb}
\usepackage{amsmath}
\usepackage{mathtools}
\usepackage{mathabx}
\usepackage{ mathrsfs }
\usepackage{amsthm}
\usepackage{tikz}
\usepackage{verbatim}
\usepackage{bbold}
\usepackage{wallpaper}
\usepackage{tikz-cd}
\usepackage{epigraph}
\usepackage{dsfont}

\usepackage{floatrow}

\usepackage{wrapfig,booktabs}

\usepackage{pict2e}
\usepackage{cleveref}
\usepackage{autonum}	
\usepackage{tikz-cd}
\usepackage{url}
\usepackage{tensor}
\usepackage{url} 
\usepackage{enumitem}
\usepackage{multirow}
\newtheoremstyle{definition_style}
{15pt}
{10pt}
{\itshape}
{}
{\bfseries}
{ }
{\newline}
{}

\newtheorem{Th}{Theorem}[section]

\newtheorem{Cor}[Th]{Corollary}

\theoremstyle{definition}
\newtheorem{Def}[Th]{Definition}
\newtheorem{Thm}[Th]{Theorem}

\newtheorem{Lem}[Th]{Lemma}
\newtheorem{Rem}[Th]{Remark}

\newtheorem{Not}[Th]{Notation}

\title{Graph Scattering beyond Wavelet Shackles}

\author{Christian Koke\\
Technical University of Munich \& \\
	Ludwig Maximilian University Munich\\
	\texttt{christian.koke@tum.de} \\
 \And
   Gitta Kutyniok\\
  Ludwig Maximilian University Munich \&\\
  University of Troms\o\\
  \texttt{kutyniok@math.lmu.de} \\
}

\begin{document}

\maketitle

\begin{abstract}
	This work develops a flexible and mathematically sound framework for the design and analysis of graph scattering networks with variable branching ratios and generic functional calculus filters.
	Spectrally-agnostic stability guarantees for node- and graph-level perturbations are derived; the vertex-set non-preserving case is treated by utilizing recently developed mathematical-physics based tools.
Energy propagation through the network layers is investigated and related to truncation stability.
	New methods of graph-level feature aggregation are introduced and stability 
	of the resulting composite scattering architectures is established. Finally, 
	scattering transforms
	are extended to edge- and higher order tensorial input. Theoretical results are complemented by numerical investigations:
	Suitably chosen scattering networks conforming to the developed theory perform better than traditional graph-wavelet based scattering approaches in social network graph classification tasks and
	significantly outperform 
	other graph-based learning approaches to regression of quantum-chemical energies on QM$7$.
\end{abstract}

\section{Introduction}\label{Int}
Euclidean wavelet scattering networks \citep{Mallat2012group,bruna2012invariant} are deep convolutional architectures where output-features are generated in each layer. Employed filters are designed rather than learned and derive from a fixed (tight) wavelet frame, resulting in a tree structured network with constant branching ratio. Such networks 
provide state of the art methods in settings with limited data availability and
serve as a mathematically tractable model of standard convolutional neural networks (CNNs). Rigorous investigations --- establishing remarkable invariance- and stability properties of wavelet scattering networks ---
were initially carried out in \citep{Mallat2012group}. The extensive mathematical analysis \citep{wt}
generalized the term 'scattering network' to include tree structured networks with varying branching rations
and frames of convolutional filters,  thus significantly narrowing the conceptual gap
to general CNNs.

With increasing interest in data on graph-structured domains, well performing networks generalizing Euclidean CNNs to this geometric setting emerged \citep{Kipf, BrunaOrig,Bresson}. If efficiently implemented, such graph convolutional networks (GCNs)  replace Euclidean convolutional filters by functional calculus filters; i.e. scalar functions applied to a suitably chosen graph-shift-oprator
capturing the geometry of the underlying graph \citep{Kipf, GWLVSGT, Bresson}. Almost immediately, 
proposals aimed at extending the success story of Euclidean scattering networks
to the graph convolutional setting began appearing: 
In \citep{Zou_2020}, the authors utilize dyadic graph wavelets (see e.g. \citep{GWLVSGT}) based on the non-normalized graph Laplacian resulting in a  norm preserving graph wavelet scattering transform. 
In \cite{gama2018diffusion}, diffusion wavelets (see e.g. \citep{COIFMAN200653})
are used to construct a graph scattering transform enjoying spectrum-dependent stability guarantees to graph level perturbations. 
For scattering transforms with $N$ layers and $K$ distinct functional calculus filters, the work \citep{gama2019stability} derives node-level stability bounds of $\mathcal{O}(K^{N/2})$  and conducts corresponding numerical experiments choosing diffusion wavelets, monic cubic wavelets  \citep{GWLVSGT}  and tight Hann wavelets \citep{tight} as filters.
In \citep{gao2019geometric} the authors, following \citep{COIFMAN200653}, construct so called geometric wavelets and establish the expressivity of a scattering transform based on such a frame through extensive numerical experiments. A theoretical analysis of this and a closely related wavelet based scattering transform is the main focus of \citep{Understanding}.
Additionally, graph-wavelet based scattering transforms have been extended to the spatio-temporal domain \citep{stgst}, utilized to overcome the problem of oversmoothing in GCNs \cite{oversmooth} and pruned to deal with their exponential (in network depth) increase in needed resources \cite{Ioannidis2020Pruned}.

Common among all these contributions is the focus on graph wavelets, which are generically understood to derive in a scale-sampling procedure from a common wavelet generating kernel function $g: \mathds{R}\rightarrow\mathds{R}$ satisfying various properties \cite{GWLVSGT}.
Established stability or expressivity properties --- especially to structural perturbations --- are then generally linked to the specific choice of the wavelet kernel $g$ and utilized graph shift operator \cite{gama2018diffusion,Understanding}. 
%
This  severely limits the diversity of available filter banks in the design of scattering networks and
draws into question their validity as models for more general GCNs whose filters generically do not derive from a wavelet kernel.

A primary focus of this work is to provide alleviation in this situation: After reviewing the graph signal processing setting in Section \ref{GSPFW}, we introduce a general framework for the construction of (generalized) graph scattering transforms beyond the wavelet setting in Section \ref{GGST}.  Section \ref{clubbedtodeath} establishes spectrum-agnostic stability guarantees on the node signal level and for the first time also for graph-level perturbations. To handle the vertex-set non-preserving case, a new 'distance measure' for operators capturing the geometry of varying graphs is utilized.
After providing conditions for energy decay (with the layers) and relating it to truncation stability, we consider graph level feature aggregation and higher order inputs in Sections  \ref{GLFA} and \ref{HOS} respectively. In Section \ref{Trinity} we then provide numerical results indicating that general functional calculus filter based scattering is at least as expressive as standard wavelet based scattering in graph classification tasks and outperforms leading graph neural network approaches to regression of quantum chemical energies on QM$7$.

\section{Graph Signal Processing}\label{GSPFW}
Taking a signal processing approach, we consider signals on graphs as opposed to graph embeddings:
\paragraph{Node-Signals:} Given a graph $(G,E)$, we are primarily interested in node-signals, which are functions from the node-set $G$ to the complex numbers, modelled as elements of $\mathds{C}^{|G|}$. We equip this space with an inner product according to $\langle f,g \rangle = \sum^{|G|}_{i=1}\overline{f_{i}}g_{i} \mu_{i} $ (with all vertex weights $\mu_{i} \geq 1$) and denote the resulting inner product space by $\ell^2(G)$.
We forego considering  arbitrary inner products on $\mathds{C}^{|G|}$  solely in the interest of increased readability.
%
%
%
\paragraph{Functional Calculus Filters:} Our fundamental objects in investigating node-signals will be functional calculus filters based on a normal operator $\Delta: \ell^2(G)\rightarrow \ell^2(G)$. Prominent examples include the adjacency matrix $W$, the degree matrix $D$,  normalized $(\mathds{1}-D^{-\frac12}WD^{-\frac12})$ or  un-normalized ($\mathcal L := D-W$) graph Laplacians 
Writing normalized eigenvalue-eigenvector pairs of $\Delta$ as $(\lambda_i,\phi_i)_{i=1}^{|G|}$, the filter obtained from applying $g:\mathds{C}\rightarrow\mathds{C}$ is given by $
g(\Delta)f = \sum_{i=1}^{|G|} g(\lambda_i)\langle\phi_i,f\rangle_{\ell^2(V)}\phi_i
$.
The operator we utilize in our numerical investigations of Section \ref{HOS}, is given by \textbf{ $\mathscr{L} := \mathcal L / \lambda_{\max} (\mathcal L)$}. We divide by the largest eigenvalue to ensure that the spectrum  $\sigma(\mathscr{L})$ is contained in the interval $[0,1]$, which aids in the choice of functions from which filters are derived.

\paragraph{Generalized Frames:} We are most interested in filters that arise from a collection of functions adequately covering the spectrum of the operator to which they are applied. To this end we call a collection $\{g_{i }(\cdot)\}_{i \in I}$ of functions a \textbf{generalized frame} if it satisfies the \textbf{generalized frame condition}
$	A \leq \sum_{i \in  I} |g_i(c)|^2 \leq B$ for any $c$ in $\mathds{C}$  for constants $A;B > 0$. As proved in Appendix \ref{opfrpf}, this condition is sufficient to guarantee that the associated operators form a frame: 
\begin{Thm}\label{opfr} Let $\Delta : \ell^2(G) \rightarrow \ell^2(G)$ be normal. 	If the family $ \{g_{i}(\cdot)\}_{i \in  I}$
	of bounded  functions satisfies $	A \leq \sum_{i \in  I} |g_i(c)|^2 \leq B$ for all $c$ in the spectrum $\sigma(\Delta)$,  we have ($\forall f \in \ell^2(G)$)
	\begin{align}
	A \|f\|^2_{\ell^2(G)} \leq  \sum\limits_{i \in  I} \|g_i(\Delta)f\|^2_{\ell^2(G)} \leq B \|f\|_{\ell^2(G)}^2.
	\end{align}
\end{Thm}
Notably, the functions $\{g_i\}_{i\in I}$ need not be continuous: In fact, in our numerical implementations, we will -- among other mappings -- utilize the function $\delta_0(\cdot)$, defined by $\delta_0(0)=1$ and $\delta_0(c)=0$ for $c\neq 0$ as well as a modified cosine, defined by $\overline{\cos}(0)=0$ and  $\overline{\cos}(c)= \cos(c)$ for $c\neq 0$.

%

\section{The Generalized Graph Scattering Transform}\label{GGST}

A generalized graph scattering transform is a non-linear map $\Phi$  based on a tree structured multilayer graph convolutional network with constant branching factor in each layer. For an input signal $f \in \ell^2(G)$, outputs are generated in each layer of such a scattering network, and then concatenated to form a feature vector in a feature space $\mathscr{F}$. The network is built up from three ingredients:

\paragraph{Connecting Operators:} To allow intermediate signal representations in the 'hidden' network layers to be further processed with functional calculus filters based on varying operators, which might not all be normal for the same choice of node-weights, we allow these intermediate representations to live in varying graph signal spaces. In fact, we do not even assume that these signal spaces are based on a common vertex set. This is done to allow for modelling of recently proposed networks where  input- and 'processing' graphs are decoupled  (see e.g. \citep{AlonaTal, BronsteinInBottle}), as well as architectures incorporating graph pooling \cite{gpooling}. Instead, we associate one signal space $\ell^2(G_n)$ to each layer $n$. Connecting operators are then (not necessarily linear) operators $P_n: \ell^2(G_{n-1}) \rightarrow \ell^2(G_{n})$ connecting the signal spaces of subsequent layers.
We assume them to be Lipschitz continuous ($\|P(f)-P(g)\|_{\ell^2( G_{n-1})}\leq R^+\|f-g\|_{\ell^2( G_n)})$ and triviality preserving ($P(0) = 0$).  For our original node-signal space we also write $\ell^2(G) \equiv \ell^2(G_0)$ . 

\paragraph{Non-Linearities:} To each layer, we also associate a (possibly) non-linear function $\rho_n:\mathds{C}\rightarrow\mathds{C}$ acting poinwise on signals in $\ell^2(G_n)$. Similar to connecting operators, we assume $\rho_n$ preserves zero  and is Lipschitz-continuous with Lipschitz constant denoted by $L^+_n$. This definition allows for the absolute value non-linearity, but also ReLu or -- trivially -- the identity function.

\paragraph{Operator Frames:} Beyond these ingredients, the central building block of our scattering architecture is comprised of a family of functional calculus filters in each layer. That is, we assume that in each layer, the node signal space $\ell^2(G_n)$ carries a normal operator $\Delta_n$ and an associated collection of functions comprised of an \textbf{output generating function} $\chi_n(\cdot)$ as well as a \textbf{filter bank} $\{g_{\gamma_n}(\cdot)\}_{\gamma_n \in  \Gamma_n}$  indexed by an index set $\Gamma_n$. As the network layer $n$ varies (and in contrast to wavelet-scattering networks) we allow the index set $\Gamma_n$ as well as the collection $\{\chi_n(\cdot)\}\bigcup\{g_{\gamma_n}(\cdot)\}_{\gamma_n \in  \Gamma_n} $ of functions to vary. We only demand  that in each layer the functions in the filter bank together with the output generating function constitute a generalized frame with frame constants $A_n,B_n \geq 0$.\\
\ \\
We refer to the collection of functions $\Omega_N:= (\rho_n, \{\chi_n(\cdot)\}\bigcup\{g_{\gamma_n}(\cdot)\}_{\gamma_n \in  \Gamma_n}  )_{n=1}^N$ as a \textbf{module sequence}  and call $\mathscr{D}_N:=(P_n,\Delta_n)_{n=1}^N$  our \textbf{operator collection}. 	The generalized scattering transform is then constructed iteratively:

\begin{minipage}{0.52\textwidth}
	
	\includegraphics[scale=0.4]{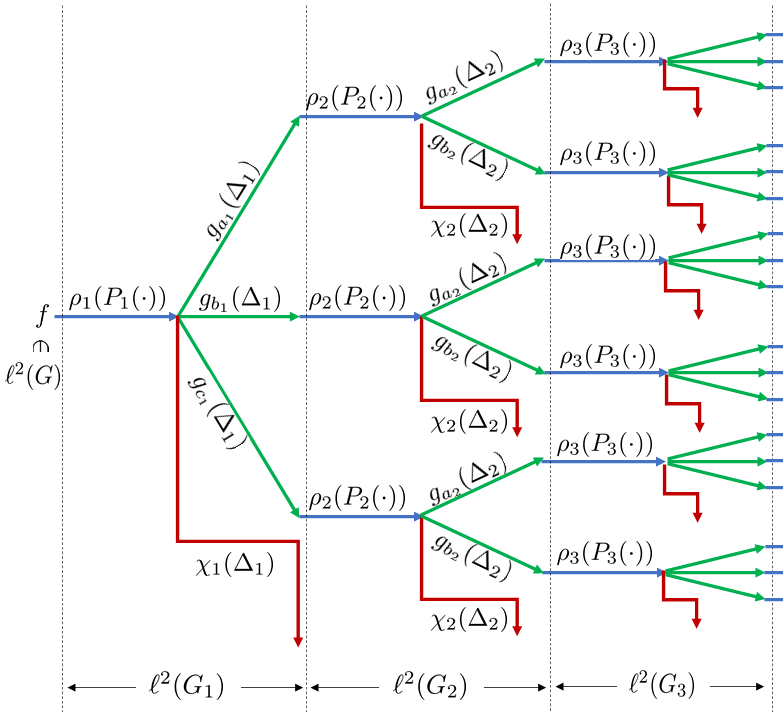}
	\captionof{figure}{Schematic Scattering Architecture} 
	\label{SKA}
\end{minipage}\hfill
\begin{minipage}{0.48\textwidth}
	
	To our initial signal $f \in \ell^2(G)$ we first apply the connecting operator $P_1$, yielding a signal representation in $\ell^2(G_1)$.  Subsequently, we apply the pointwise non-linearity $\rho_1$. Then we apply our graph filters $\{\chi_1(\Delta_1)\} \bigcup \{g_{\gamma_1}(\Delta_1)\}_{\gamma_1 \in  \Gamma_1}$ to $\rho_1(P_1(f))$ yielding the 
	output $V_1(f) := \chi_1(\Delta_1)\rho_1(P_1(f))$ as well as the intermediate hidden representations $\{U_1[\gamma_1](f) := g_{\gamma_1}(\Delta_1)\rho_1(P_1(f))\}_{\gamma_1 \in \Gamma_1} $ obtained in the first layer.
	Here we have introduced the \textbf{one-step scattering propagator} $U_n[\gamma_n]: \ell^2(G_{n-1}) \rightarrow \ell^2(G_{n})$ mapping $f \mapsto g_{\gamma_n}(\Delta_n)\rho_n(P_n(f))$  as well as the \textbf{output generating operator} $V_{n}: \ell^2(G_{n-1})\rightarrow \ell^2(G_n)$ mapping $f$ to $\chi_n(\Delta_n)\rho_n(P_n(f))$.  Upon defining the \textbf{set} $\Gamma^{N-1} := \Gamma_{N-1} \times...\times\Gamma_1$ \textbf{of paths} of length $(N-1)$  terminating in layer $N-1$ (with $\Gamma^0$ taken to be the one-element set) and iterating the above procedure, we see that the outputs generated in the $N^{\text{th}}$-layer  are indexed by paths $\Gamma^{N-1}$ terminating in the previous layer.
	
\end{minipage}


%

Outputs generated in the $N^\textit{th}$ layer are thus given by $\{V_N\circ U[\gamma_{N-1}]\circ...\circ U[\gamma_1](f)\}_{(\gamma_{N-1},...,\gamma_1)\in \Gamma^{N-1}}$.
Concatenating the features obtained in the various layers of a network 
with depth $N$,  our full feature vectors thus live in the feature space
\begin{equation}\label{FTSP}
\mathscr{F}_N = \oplus_{n=1}^N \left(  \ell^2(G_n)  \right)^{|\Gamma^{n-1}|}.
\end{equation}
The associated canonical norm is denoted $\|\cdot\|_{\mathscr{F}_N}$. For convenience, a brief review of direct sums of spaces, their associated norms and a discussion of corresponding direct sums of maps is provided in Appendix \ref{LA}.
We denote the hence constructed generalized scattering transform of length $N$, based on a module sequence $\Omega_N$ and operator collection $\mathscr{D}_N$ by $\Phi_N$.\\
\begin{minipage}{0.68\textwidth}
In our numerical experiments in Section \ref{Trinity}, we consider two particular instantiations of the above general architecture. In both cases the utilized shift-operator is $\mathscr{L} := \mathcal L / \lambda_{\max} (\mathcal L)$, node weights satisfy $\mu_i =1$, the  branching ratio in each layer is chosen as $4$ and the depth is set to $N=4$ as well. The connecting operators are set to the identity and non-linearities are set to the modulus ($|\cdot|$).  The two architectures differ in the utilized filters, which are repeated in each layer and depicted in Fig. \ref{BothArchitectures}. Postponing a discussion of other parameter-choices, we note here that the filters $\{\sin(\pi/2\cdot), \cos(\pi/2\cdot )\}$ provide a high and a low pass filter on the spectrum $\sigma(\mathscr L) \subseteq [0,1]$, while $\{\sin(\pi\cdot), \cos(\pi\cdot )\}$ provides a spectral refinement of the former two filters. The inner two elements of the filter bank in Architecture II thus separate an input signal into high- and low-lying spectral 
\end{minipage}\hfill
\begin{minipage}{0.3\textwidth}
	\includegraphics[scale=0.4]{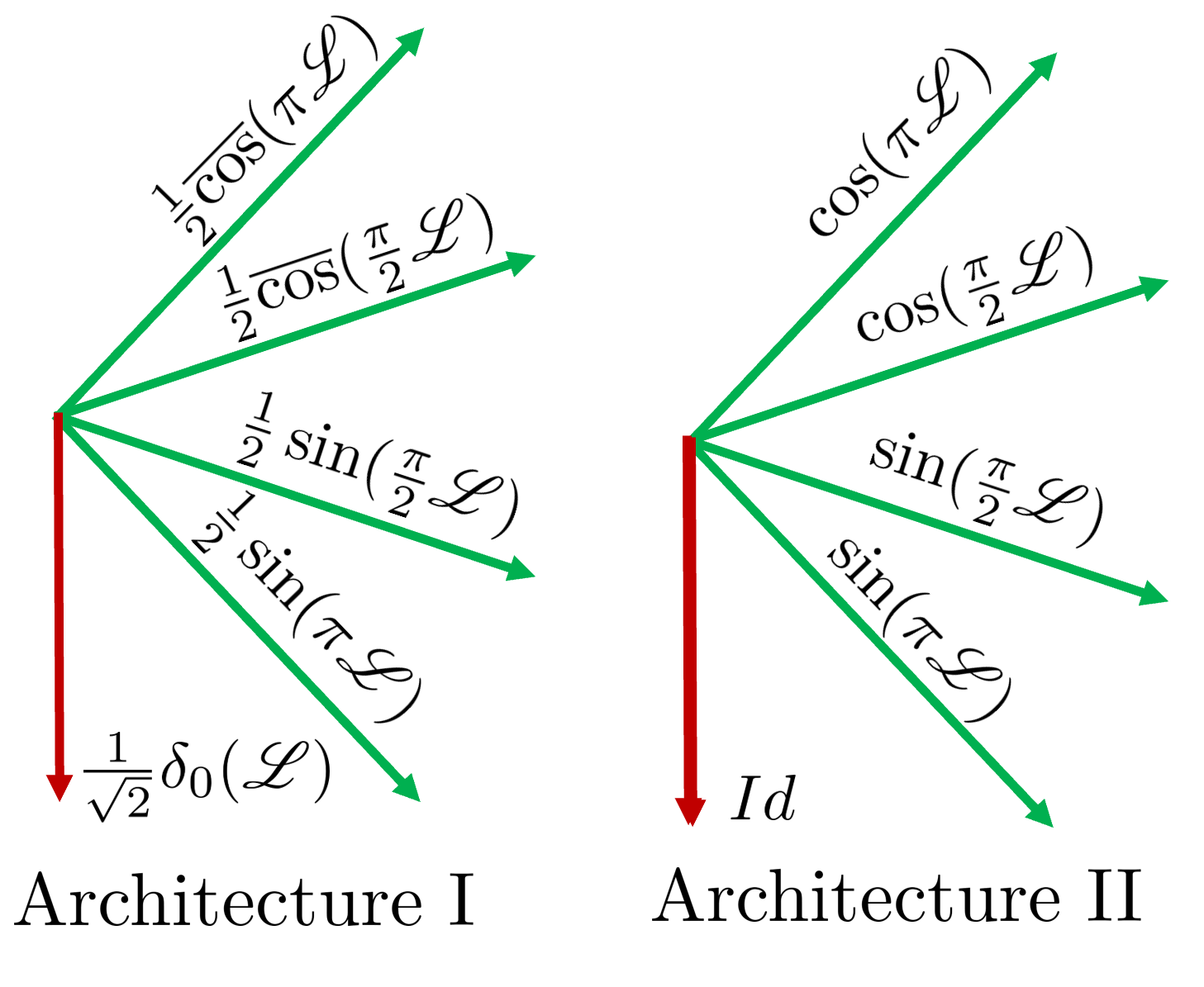}
	\captionof{figure}{Filters of tested Architectures} 
	\label{BothArchitectures}
\end{minipage}
components. The outer two act similarly at a higher spectral scale. Additionally Architecture I -- utilizing   $\overline{\cos}$ and $\delta_0$ as introduced Section \ref{GSPFW} -- prevents the lowest lying spectral information from propagating. Instead it is extracted via $\delta_0(\cdot)$ in each layer. Note that $Id $ arises from applying the constant-$1$ function to $\mathscr L$. Normalizations are chosen to generate frames with upper bounds $B \lessgtr 1$.
%
%
%
%
%
%

\section{Stability Guarantees}\label{clubbedtodeath}
In order to produce meaningful signal representations, a small change in input signal should produce a small change in the output of our generalized scattering transforms. This property is captured in the result below, which is proved in Appendix \ref{sptcy}.
\begin{Thm}\label{signalpertcty}
	With the notation of Section \ref{GGST}, we have for all $f,h\in\ell^2(G)$:
	\begin{align}
	\|\Phi_N(f) - \Phi_N(h)\|_{\mathscr{F}_N} \leq \left(1+ \sum\limits_{n=1}^N\max\{[B_n-1],[B_n(L_n^+R_n^+)^2-1],0\}\prod\limits_{k=1}^{n-1}B_k\right)^\frac12\|f-h\|_{\ell^2(G)}
	\end{align}
\end{Thm}
In the case where upper frame bounds $B_n$ and Lipschitz constants $L_n^+$ and $R_n^+$ are all smaller than or equal one, this statement reduces to the much nicer inequality:
\begin{equation}\label{nicerbound}
\|\Phi_N(f) - \Phi_N(h)\|_{\mathscr{F}_N} \leq \|f-h\|_{\ell^2(G)}.
\end{equation}
Below, we always assume $R^+_n,L^+_n \leq 1$ as this easily achievable through rescaling. We will keep $B_n$ variable to demonstrate how  filter size influences stability results.
%
%
%
%
As for our experimentally tested architectures (cf. Fig. \ref{BothArchitectures}), we note for Architecture I that  $B_n = 1/2$ for all $n$, so that (\ref{nicerbound}) applies. For Architecture II we have $B_n = 3$, which yields a stability constant of $\sqrt{1+2\cdot3+2\cdot3^2+2\cdot3^3} = 9$. Similar to other constants derived in this section, this bound is however not necessarily tight.

Operators capturing graph geometries might only be known approximately in real world tasks; e.g. if edge weights are only known to a certain level of precision. Hence it is important that our scattering representation be insensitive to small perturbations in the underlying normal operators in each layer, which is captured by our next result, proved in Appendix \ref{sdimoppertpf}.
Smallness here is measured in Frobenius norm $\|\cdot\|_F$, which for convenience is briefly reviewed in Appendix \ref{LA}).
\begin{Thm}\label{sdimoppert}
	Let $\Phi_N$ and $\widetilde{\Phi}_N$ be two  scattering transforms  based on the same module sequence $\Omega_N$ and operator sequences
	$\mathscr{D}_N,\widetilde{\mathscr{D}}_N$  with the same connecting operators ($P_n = \widetilde P_n$) in each layer. Assume $R^+_n,L^+_n\leq 1$ and $B_n\leq B$ for some $B$ and $n \leq N$. Assume that the respective normal operators satisfy $\|\Delta_n - \widetilde \Delta_n\|_F \leq \delta$ for some $\delta >0$. Further assume that  the  functions $\{g_{\gamma_n}\}_{\gamma_n \in \Gamma_n}$ and $\chi_n$ in each layer  are Lipschitz continuous with associated Lipschitz constants
	satisfying $L_{\chi_n}^2+\sum_{\gamma_n \in \Gamma_n} L_{g_{\gamma_n}}^2\leq D^2$ for all $n \leq N$ and some $D>0$. Then we have 
	\begin{equation}
	\|\widetilde \Phi_N(f)  -\Phi_N(f) \|_{\mathscr{F}_N} \leq  \sqrt{2(2^{N}-1)  }\cdot\sqrt{(\max\{B,1/2\})^{N - 1}} \cdot D \cdot \delta \cdot	 \|f\|_{\ell^2(G)}
	\end{equation}
for all $f\in \ell^2(G)$. If $B \leq 1/2$, the stability constant improves to $\sqrt{2(1 - B^N)/(1 - B)} \cdot D \leq 2 \cdot D $.
\end{Thm}
The condition $B\leq \frac12$ is e.g. satisfied by our Architecture I, but  --strictly speaking--  we may  not apply Theorem \ref{sdimoppert}, since not all utilized filters are Lipschitz continuous. Remark \ref{stabremA} in Appendix \ref{sdimoppertpf} however shows, that the above stability result remains applicable for this architecture as long as we demand that $\Delta$ and $\widetilde \Delta$ are (potentially rescaled) graph Laplacians. For Architecture II we note that $D = \pi\sqrt{10}/2 $ and thus the stability constant is given by $\sqrt{2(2^{4}-1) } \cdot \sqrt{3^3 } \cdot \pi\sqrt{10}/2  =45 \pi$.

We are also interested in perturbations that change the vertex set of the graphs in our architecture. This is important for example in the context of social networks, when passing from nodes representing individuals to nodes representing (close knit) groups of individuals.  To investigate this setting, we utilize tools originally developed within the mathematical physics community \citep{PostBook}:
\begin{Def}\label{TFDef}
	Let $\mathcal{H}$ and $\widetilde{\mathcal{H}}$ be two finite dimensional Hilbert spaces. Let $\Delta$ and $\widetilde \Delta$ be normal operators on these spaces.  Let $J:\mathcal{H}\rightarrow\widetilde{\mathcal{H}}$ and $\widetilde J:\widetilde{\mathcal{H}}\rightarrow\mathcal{H}$ be linear maps --- called \textbf{identification operators}.   We call the two  spaces \textbf{$\delta$-quasi-unitarily-equivalent} (with $\delta \geq 0$) if for any $f \in \mathcal{H}$ and $u \in \widetilde{\mathcal{H}}$ we have
	\begin{align}
	\|Jf\|_{\widetilde {\mathcal{H}}} \leq 2 \|f\|_{\mathcal{H}},& \ \ \ \ \|(J - \widetilde {J}^* )f\|_{\widetilde {\mathcal{H}}} \leq \delta  \|f\|_{\mathcal{H}},\\
	\| f -  \widetilde{J} Jf\|_{\mathcal H} \leq \delta \sqrt{\|f\|_{\mathcal{H}}^2 + \langle f, |\Delta|\ f\rangle_{\mathcal{H}}   }, & \ \ \ \   \|u - J \widetilde J u\|_{\widetilde {\mathcal{H}}} \leq \delta \sqrt{\|u\|_{\widetilde {\mathcal{H}}}^2 + \langle u,|\widetilde{\Delta}|\ u\rangle_{\widetilde {\mathcal{H}}}}.
	\end{align}
	If, for some $w \in \mathds{C}$ 
	 the resolvent $R := (\Delta-\omega)^{-1} $  satisfies
	$
	\| (\widetilde R J  - J R) f\|_{\widetilde{\mathcal{H}}} \leq \delta\|f\|_{\mathcal{H}}
	$ for all $f \in \mathcal{H}$,
	we say that $\Delta$ and $\widetilde \Delta$ are   \textbf{$\omega$-$\delta$-close} with identification operator $J$.
\end{Def}


\begin{minipage}{0.68\textwidth}
	Absolute value $|\Delta|$ and adjoint $\widetilde J^\asterisk$ of operators are briefly reviewed in Appendix \ref{LA}. While the above definition might seem fairly abstract at first, it is in fact a natural setting to investigate structural perturbations as Figure \ref{prettygraphdrawings} exemplifies.
	In our current setting, the Hilbert spaces in Definition \ref{TFDef} are node-signal spaces $\mathcal{H} = \ell^2(G)$, $\widetilde{\mathcal{H}} = \ell^2(\widetilde G)$ of  different graphs. 
	The notion of $\omega$-$\delta$-closeness is then useful, as it allows to compare filters  defined on different graphs but obtained from applying the same function to the respective graph-operators:

	\begin{Lem} \label{cpl}
		In the setting of Definition \ref{TFDef} let $\Delta$ and $\widetilde \Delta$ be $\omega$-$\delta$-close and satisfy $\|\Delta\|_\textit{op},\|\widetilde \Delta\|_\textit{op}\leq K$ for some $K>0$.
		If $g:\mathds{C} \rightarrow\mathds{C}$  is holomorphic on the disk $B_{K+1}(0)$ of radius $(K+1)$, there is a constant $C_g \geq 0$ so that
		\begin{equation}
		\|g(\widetilde\Delta )J - J g(\Delta)\|_\text{op} \leq C_g \cdot \delta
		\end{equation} 
		with $C_g$ depending on $g$, $\omega$ and $K$. 
	\end{Lem}
	An explicit characterization of $C_g$  together with a proof of this result   is presented in  Appendix \ref{cplpf}.
	Lemma \ref{cpl} is our main tool in establishing our next result, proved in Appendix \ref{BigTpf}, which captures stability under vertex-set non-preserving perturbations:
	
\end{minipage}\hfill
\begin{minipage}{0.3\textwidth}
	
	\includegraphics[scale=0.7]{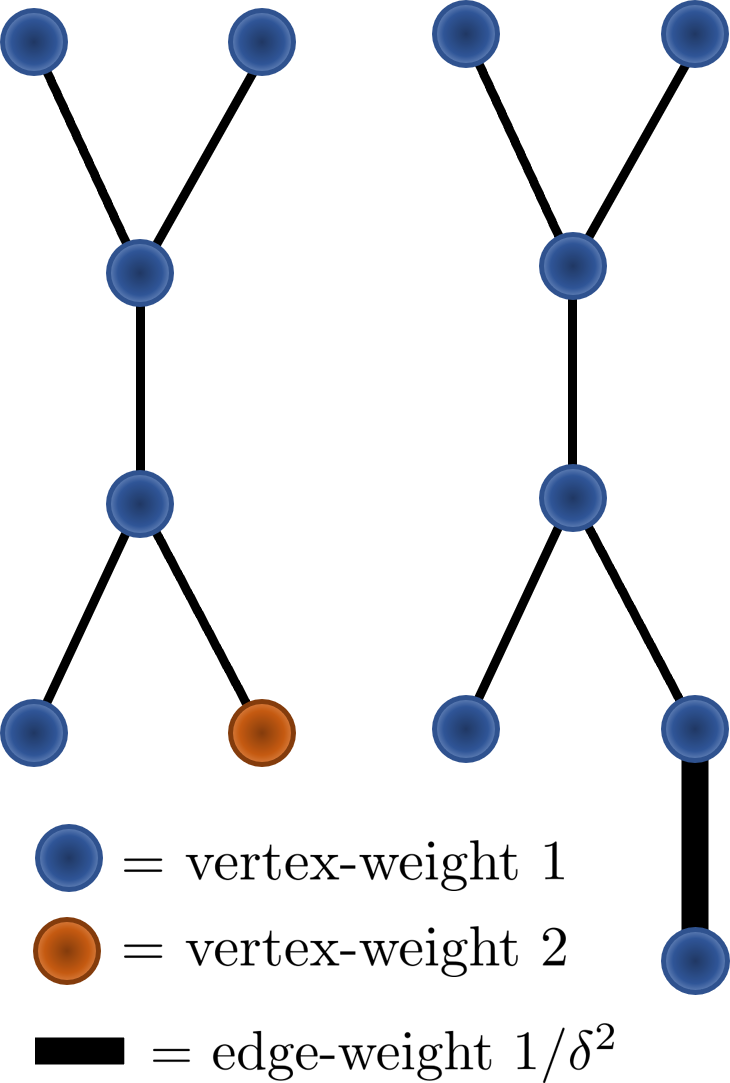}
	\captionof{figure}{Prototypical Example of $\delta$-unitary-equivalent Node Signal Spaces with  $(-1)$-$12\delta$-close Laplacians. Details in Appendix \ref{delta}.} 
	\label{prettygraphdrawings}
\end{minipage}

\begin{Thm}\label{BigT} 
	Let $\Phi_N, \widetilde{\Phi}_N$ be  scattering transforms based on a common module sequence $\Omega_N$ and differing operator sequences $\mathscr{D}_N, \widetilde{\mathscr{D}}_N$.  Assume $R^+_n,L^+_n\leq 1$ and $B_n\leq B$ for some $B$ and $n \geq 0$.
	Assume that there are identification operators $J_n: \ell^2(G_n) \rightarrow \ell^2(\widetilde G_n) $, $\widetilde{J}_n: \ell^2(\widetilde{G}_n) \rightarrow \ell^2( G_n) $ ($0\leq n \leq N$) so that the respective signal spaces are $\delta$-unitarily equivalent, the respective normal operators $\Delta_n,\widetilde{\Delta}_n$ are $\omega$-$\delta$-close as well as bounded (in norm) by $K>0$ and the connecting operators satisfy $\|\widetilde P_nJ_{n-1}f - J_n P_n f\|_{\ell^2(\widetilde G_n)}= 0$. For the common module sequence $\Omega_N$ assume  that the non-linearities  satisfy $\| \rho_n(J_{n}f) - J_n \rho_n(f)\|_{\ell^2(\widetilde G_n)}= 0$ and that the constants $C_{\chi_n}$ and $\{C_{g_{\gamma_n}}\}_{\gamma_n \in \Gamma_N}$	associated through Lemma \ref{cpl} to the functions of the generalized frames in each layer satisfy $ C^2_{\chi_n} + \sum_{\gamma_n\in \Gamma_N} C^2_{g_{\gamma_n}} \leq D^2$ for some $D > 0$. Denote the operator that the family $\{J_n\}_n$ of identification operators induce on $\mathscr{F}_N$ through concatenation by
	$\mathscr{J}_N: \mathscr{F}_N \rightarrow \widetilde{\mathscr{F}}_N$. 
	Then, with $K_N = \sqrt{ (2^N-1) 2 D^2 \cdot B^{N-1}}$ if $B > 1/2$  and $K_N = \sqrt{  2 D^2 \cdot (1 - B^N)/(1 - B)} $ if $B \leq 1/2$:
	\begin{align}
	\|\widetilde{\Phi}_N(J_0f) - \mathscr{J}_N \Phi_N(f) \|_{\widetilde{\mathscr{F}}_N}	\leq K_N  \cdot \delta \cdot \|f\|_{\ell^2(G},\ \  \ \forall f \in \ell^2(G).
	\end{align}
\end{Thm}
The stability result persists with slightly altered stability constants, if identification operators only {\it almost} commute with non-linearities and/or connecting operators, as Appendix \ref{BigTpf}  further elucidates. Theorem \ref{BigT} is not applicable to Architecture I, where filters are not all holomorphic, but is directly applicable to Architecture II. Stability constants can be calculated in terms of $D$ and $B$ as before.

Beyond these results, stability under truncation of the scattering transform is equally desirable: 
Given the energy  $W_N := \sum_{(\gamma_N,...,\gamma_1) \in \Gamma^{N}}\|U[\gamma_{N}]\circ...\circ U[\gamma_1](f)\|^2_{\ell^2(G_{N})} $ stored in the network at layer $N$, it is not hard to see that after extending $\Phi_N(f)$ by zero to match dimensions with $\Phi_{N+1}(f)$ we have
$
	\| \Phi_N(f) - \Phi_{N+1}(f) 	\|^2_{\mathscr{F}_{N + 1}} \leq \left(R^+_{N+1} L^+_{N+1} \right)^2B_{N+1} \cdot W_N
$ (see Appendix \ref{somelabelpf} for more details).
A bound for $W_N$ is then given as follows:
\begin{Thm}\label{EDTM} Let $\Phi_\infty$ be a generalized  graph scattering transform based on a  an operator sequence $\mathscr{D}_\infty = (P_n,\Delta_n)_{n=1}^\infty$ and a module sequence $\Omega_\infty$ with each $\rho_n(\cdot)\geq 0$. Assume in each layer $n \geq 1$ that there is an eigenvector $\psi_n$ of $\Delta_n$ with solely positive entries; denote the smallest entry by $m_n:=\min_{i \in G_n}\psi_n[i] $ and the eigenvalue corresponding to $\psi_n$ by $\lambda_n$. Quantify the 'spectral-gap' opened up at this eigenvalue through neglecting the output-generating function by $\eta_n := \sum_{\gamma_n \in \Gamma_n} |g_{\gamma_n}(\lambda_n)|^2  $ and assume $B_nm_n \geq \eta_n$. We then have (with $
	C^+_N := \prod_{i=1}^N \max\left\{ 1,B_i(L_i^+R_i^+)^2\right\}$)
	\begin{align}\label{productstuff}
		W_N(f) \leq C_N^+\cdot \left[\prod\limits_{n=1}^{N}\left(1-\left(m_n-\frac{\eta_n}{B_n}\right)\right)\right]\cdot\|f\|^2_{\ell^2(G)}.
	\end{align}
\end{Thm}
The product in (\ref{productstuff}) decays if  $C^+_N\rightarrow C^+$ converges  and $\sum_{n=1}^N (m_n-\eta_n/B_n)\rightarrow \infty$ diverges as $N \rightarrow \infty$. The positivity-assumptions on the eigenvectors $\psi_n$ can e.g. always be ensured if they are chosen to lie in the lowest lying eigenspace of a graph Laplacian or normalized graph Laplacian (irrespective of the connectedness of the underlying graphs). As an example, we note that if we extend our Architecture I to infinite depth (recall from Section \ref{GGST} that we are using the same filters, operators, etc. in each layer) we have upon choosing $\lambda_n = 0$ and $\psi_n$ to be the constant normalized vector that  $\eta_n = 0$, $C_N = 1$ and $m_n = 1/\sqrt{|G|}$, for a graph with $|G|$ vertices. On a graph with $16$ vertices, we then e.g. have $W_N \leq (3/4)^N \|f\|^2_{\ell^2(G)} $ and thus $
\| \Phi_N(f) - \Phi_{N+1}(f) 	\|_{\mathscr{F}_{N + 1}} \leq  (3/4)^N \cdot \|f\|^2_{\ell^2(G)}/2
$.\\
 As detailed in Appendix \ref{somelabelpf}, Theorem \ref{EDTM} also implies that under the given assumptions the scattering transform has trivial 'kernel' for $N \rightarrow \infty$, mapping only $0$ to $0$.

\section{Graph-Level Feature Aggregation}\label{GLFA}
To solve tasks such as graph classification or regression over multiple graphs, we need to represent graphs of varying sizes 
in a common feature space. Given a scattering transform $\Phi_N$, we thus need to find a stability preserving map  from the feature space $\mathscr{F}_N$ to some Euclidean space that is independent of any vertex set cardinalities. Since  $\mathscr{F}_N$ is a large direct sum of smaller spaces (cf. (\ref{FTSP})), we simply construct such maps on each summand independently and then concatenate them.
\paragraph{General non-linear feature aggregation:}\label{aggro1}
Our main tool in passing 
to graph-level features is a non-linear map $N^G_p:\ell^2(G) \rightarrow \mathds{R}^p$ given as
\begin{equation}\label{NGP}
N^G_p(f) = \frac{1}{\sqrt{p}}(\|f\|_{\ell^1(G)}/\sqrt{\mu_G},\|f\|_{\ell^2(G)},\|f\|_{\ell^3(G)},...,\|f\|_{\ell^p(G)})^\top,
\end{equation}
with  $\mu_G:=\sum_{i\in G}\mu_i$ and
$\|f\|_{\ell^q(G)} := (\sum_{i \in G}|f_i|^q \mu_i)^{1/q}$. 
Our inspiration to use this map stems from the standard case where all $\mu_i=1$: 
For $p \geq |G|$, the vector $|f| = ((|f_1|,...,|f_G|)^\top$ can then be recovered from $N^G_p(f)$ up to permutation of indices \cite{Sturmfels}. Hence, employing $N^G_p$ (with $p \geq |G|$) to aggregate node-information into graph-level information, we lose the minimal necessary information about node permutation
(clearly $N^G_p(f) = N^G_p(\Pi f)$ for any permutation matrix $\Pi$)
and beyond that only information about the complex phase (respectively the sign in the real case) in each entry of $f$. 


\begin{minipage}{0.42\textwidth}
	
	\includegraphics[scale=0.5]{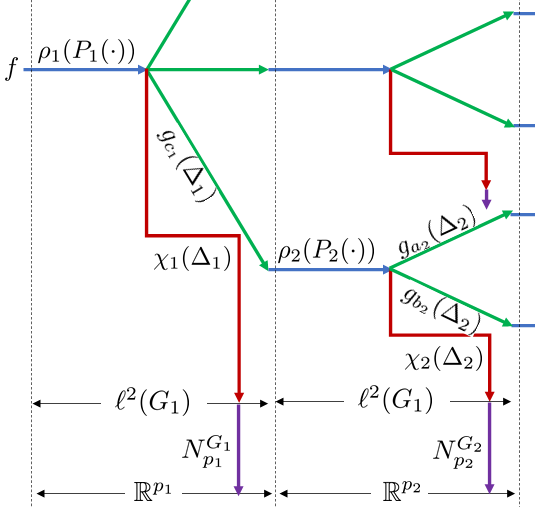}
	\captionof{figure}{Graph Level Scattering } 
	\label{SKA2}
\end{minipage}\hfill
\begin{minipage}{0.58\textwidth}
	Given a scattering transform $\Phi_N$ mapping from $\ell^2(G)$ to the feature space $	 \mathscr{F}_N = \oplus_{n=1}^N \left(  \ell^2(G_n)  \right)^{|\Gamma^{n-1}|}$, we  obtain a corresponding map $\Psi_N$ mapping from $\ell^2(G)$ to  $	\mathscr{R}_N = \oplus_{n=1}^N \left(  \mathds{R}^{p_n} \right)^{|\Gamma^{n-1}|}$ by concatenating the feature map $\Phi_N$ with the operator that the family of non-linear maps $\{  N^{p_n}_{G_n} \}_{n=1}^N$  induces on $\mathscr{F}_N$ by concatenation. Similarly we obtain the map $\widetilde{\Psi}_N:\ell^2(\widetilde G)\rightarrow \mathscr{R}_N $ by concatenating the map $\widetilde{\Phi}_N:\ell^2(\widetilde G) \rightarrow \oplus_{n=1}^N \left(  \ell^2(\widetilde G_n)  \right)^{|\Gamma^{n-1}|}$ with the operator induced by the family $\{  N^{p_n}_{\widetilde G_n} \}_{n=1}^N$. The feature space $\mathscr{R}_N$ is completely determined by path-sets $\Gamma^N$ and used maximal $p$-norm indices $p_n$.
	It no longer depends on cardinalities of vertex sets of any graphs, allowing to compare (signals on) varying graphs with each other.   Most of the results of the previous sections  then readily transfer to the   graph-level-feature setting (c.f. Appendix \ref{Awildsectionappears}).	
\end{minipage}
\paragraph{Low-pass feature aggregation:}\label{Lp}
The spectrum-free aggregation scheme of the previous paragraph is especially adapted to settings where there are no high-level spectral properties remaining constant under graph perturbations.
 However, many commonly utilized operators, such as normalized and un-normalized graph Laplacians,
have a somewhat 'stable' spectral theory:  Eigenvalues are always real, non-negative, the lowest-lying eigenvalue equals zero and simple (if the graph is connected).
In this section we shall thus assume that each mentioned normal operator $\Delta_n$ ($\widetilde{\Delta}_n$) has these spectral properties. We denote the lowest lying normalized eigenvector (which is generically determined up to a complex phase) by $\psi_{\Delta_n}$ and denote by $M^{|\langle\cdot,\cdot\rangle|}_{G_n}:\ell^2(G_n)\rightarrow\mathds{C}$ the map given by $M^{|\langle\cdot,\cdot\rangle|}_{G_n}(f) = |\langle \psi_{\Delta_n},f\rangle_{\ell^2(G_n)}|$. The absolute value around the inner product is introduced to absorb the phase-ambiguity in the choice of $\psi_{\Delta_n}$.
Given a scattering transform $\Phi_N$ mapping from $\ell^2(G)$ to the feature space $\mathscr{F}_N$, 
we  obtain a corresponding map $\Psi^{ |\langle\cdot,\cdot\rangle|}_N$ 
mapping from $\ell^2(G)$ to  $	\mathscr{C}_N = \oplus_{n=1}^N   \mathds{C} ^{|\Gamma^{n-1}|}$ by concatenating the feature map $\Phi_N$ with the operator that the family of maps $\{  M^{|\langle\cdot,\cdot\rangle|}_{G_n} \}_{n=1}^N$  induces on $\mathscr{F}_N$ by concatenation. As detailed in Appendix \ref{J2}, this map inherits stability properties in complete analogy to the discussion of Section \ref{clubbedtodeath}.

\section{Higher Order Scattering}\label{HOS}
Node signals capture information about nodes in isolation. However, one might
 be interested in binary, ternary or even higher order relations between nodes such as distances or angles in graphs representing molecules. In this section we focus on binary relations -- i.e. edge level input -- as this is the instantiation we also test in our regression experiment in Section \ref{Trinity}. Appendix \ref{ktwo} provides more details and extends these considerations beyond the binary setting. We equip the space of edge inputs with an inner product according to  $\langle f,g \rangle = \sum^{|G|}_{i,j=1}\overline{f_{ij}}g_{ij} \mu_{ij} $ and  denote the resulting inner-product space by $\ell^2(E)$ with $E = G  \times G $ the set of edges.
Setting e.g. node-weights $\mu_i$ and edge weights $\mu_{ik}$ to one, the adjacency matrix $W$ as well as normalized or un-normalized graph Laplacians constitute self-adjoint operators on $\ell^2(E)$, where they act by matrix multiplication.
Replacing the $G_n$ of Section \ref{GGST} by $E_n$, we can then follow the recipe laid out there in constructing $2^{\textit{nd}}$-order scattering transforms; all that we need are a module sequence $\Omega_N$ and an operator sequence $\mathscr{D}^2_N:=(P^2_n,\Delta^2_n)_{n=1}^N$, where now $P^2_n:\ell^2(E_{n-1})\rightarrow \ell^2(E_{n})$ and $\Delta^2_n:\ell^2(E_{n}) \rightarrow \ell^2(E_{n})$.
We denote the resulting feature map by $\Phi_N^2$
and write $\mathscr{F}^2_N$ for the corresponding feature space. 
%
The map $N^G_p$ introduced in (\ref{NGP}) can also be adapted to aggregate higher-order  features into graph level features:
With  $\|f\|_q := (\sum_{ij \in G}|f_{ij }|^q \mu_{ij} )^{1/q} $ and $\mu_{E}:= \sum_{ij=1}^{|G|}\mu_{ij}$, we define $
N^{E}_p(f) = (\|f\|_{\ell^1(E)}/\sqrt{\mu_{E}},\|f\|_{\ell^2(E)},\|f\|_{\ell^3(E)},...,\|f\|_{\ell^p(E)})^\top/\sqrt{p}$.
Given a feature map $\Phi^2_N$ with feature space $	 \mathscr{F}^2_N = \oplus_{n=1}^N \left(  \ell^2(E_n)  \right)^{|\Gamma^{n-1}|}$, we  obtain a corresponding map $\Psi^2_N$ mapping from $\ell^2(E)$ to  $	\mathscr{R}_N = \oplus_{n=1}^N \left(  \mathds{R}^{p_n} \right)^{|\Gamma^{n-1}|}$ by concatenating  $\Phi^E_N$ with the map that the family of non-linear maps $\{  N_{p_n}^{E_n} \}_{n=1}^N$  induces on $\mathscr{F^2}_N$ by concatenation.  The stability
results of the preceding sections then readily translate to $\Phi^2_N$ and $\Psi^2_N$ 
(c.f. Appendix \ref{HLSTB}).

\section{Experimental Results}\label{Trinity}
We showcase that even upon selecting the fairly simple Architectures I and II introduced in Section \ref{GGST} (c.f. also Fig. \ref{BothArchitectures}), our generalized graph scattering networks  are able to outperform both wavelet-based scattering transforms and  leading graph-networks under different circumstances. To aid visual clarity when comparing results, we colour-code the best-performing method in green, the second-best performing in yellow and the third-best performing method in orange respectively.
\paragraph{Social Network Graph Classification:}
To facilitate contact between our generalized graph scattering networks, 
and the wider literature, we combine a  network conforming to our general theory namely Architecture I in Fig. \ref{BothArchitectures} (as discussed in Section \ref{GGST} with depth $N=4$, identity as connecting operators and $|\cdot|$-non-linearities) with the low pass aggregation scheme of Section \ref{Lp} and a Euclidean support vector machine with RBF-kernel (GGSN+EK). The choice $N=4$ was made to keep computation-time palatable, while  aggregation scheme and non-linearities were chosen to facilitate comparison with standard {\it wavelet}-scattering approaches.  For this hybrid architecture (GGSN+EK), classification accuracies under the standard choice of $10$-fold cross validation on five common social network graph datasets are compared with performances of popular graph kernel approaches, leading deep learning methods as well as geometric wavelet scattering (GS-SVM) \cite{gao2019geometric}. More details are provided in Appendix \ref{AddEx}.
As  evident from Table \ref{classify}, our network consistently achieves higher accuracies than the geometric wavelet scattering transform of \cite{gao2019geometric}, with the performance gap becoming significant on the more complex REDDIT datasets, reaching a relative mean performance increase of more than $10\%$ on REDDIT-$12$K. This indicates the liberating power of transcending the graph wavelet setting.
While on comparatively smaller and somewhat simpler datasets there is a performance gap between our static architecture and fully trainable networks, this gap closes on more complex datasets: While P-Poinc e.g. outperforms our method on IMDB datasets, the roles are reversed on REDDIT datasets. On REDDIT-B our approach trails only GIN; with  difference in accuracies insignificant. On  REDDIT-$5$K our method comes in third, with the gap to the second best method (GIN) being statistically insignificant. On REDDIT-$12$K we generate state of the art results.
\begin{table}[H]
	
	\small
	\centering
	\caption{Classification Accuracies on Social Network Datasets }
	\scalebox{0.85}{
		\begin{tabular}{ p{2.6cm}|p{1.6cm}p{1.6cm}p{1.6cm}p{1.6cm}p{1.8cm}p{1.9cm}  }
			\hline
			\textbf{Method} &	\multicolumn{6}{c}{\textbf{Classification Accuracies} $[\%]$ } \\
			\hline
			\ 			& \textbf{COLLAB} 	&\textbf{IMDB- B}&\textbf{IMDB-M}&		 \textbf{REDDIT-B} & \textbf{REDDIT-5K}&\textbf{REDDIT-12K} \\
			\hline
			WL  \cite{WL} & $77.82\pm 1.45$  		&  $71.60\pm 5.16$  & N/A & 		$78.52 \pm 2.01$	&	$50.77\pm 2.02$ & $34.57\pm 1.32$\\
			
			Graphlet \cite{Graphlet} &  			$73.42\pm 2.43$     &	$65.40\pm 5.95$  & N/A&	$77.26 \pm 2.34 $	&$39.75 \pm 1.36 $ &$25.98 \pm 1.29$\\

			
			
			DGK \cite{DGK} &  			$73.00\pm 0.20$     &	$66.90\pm0.50$  & $44.50 \pm 0.50$&	$78.00 \pm 0.30 $	& $41.20 \pm 0.10$ & $32.20 \pm 0.10 $ \\

			DGCNN \cite{DGCNN} &  			$73.76\pm 0.49$     &	$70.03\pm0.86$  & $47.83 \pm0.85$&	N/A	& $ 48.70 \pm4.54$ & N/A \\


			PSCN \cite{PSCN} &  			$72.60 \pm 2.15$     &	$71.00\pm2.29$  & $45.23\pm2.84$ &	$86.30\pm1.58$	& $ 49.10 \pm 0.70$ & $41.32\pm0.42$ \\	
				
			P-Poinc \cite{kyriakis2021learning}& N/A & \cellcolor{green} $81.86 \pm 4.26$ & \cellcolor{green} $57.31 \pm 4.27$   & $79.78 \pm 3.21$ & $51.71 \pm 3.01$ & $42.16 \pm 3.41$\\
			
		S2S-N2N-PP \cite{S2S} &  		\cellcolor{yellow}	$81.75 \pm 0.80$     &	$  73.80 \pm 0.70$  &  $51.19 \pm 0.50$ &	$86.50 \pm 0.80$	& $ 52.28 \pm 0.50$ & $42.47\pm0.10$\\

			
			GSN-e \cite{bouritsas2022improving} & \cellcolor{green} $85.5 \pm 1.2$   & \cellcolor{yellow}  $77.8 \pm 3.3$ & \cellcolor{yellow} $54.3 \pm 3.3$ & N/A &N/A & N/A \\

			WKPI-kC\cite{zhao2019learning}&	N/A  &  \cellcolor{orange} $75.1 \pm 1.1$ &  $49.5 \pm 0.4$ &  N/A & \cellcolor{green} $59.5\pm 0.6$ & \cellcolor{yellow} $48.4\pm 0.5$\\	
			
			GIN \cite{GIN} &  		$80.20\pm1.90$     & \cellcolor{orange}	$75.10 \pm 5.10$  & \cellcolor{orange} $52.30\pm2.80$ &	\cellcolor{green} $92.40\pm2.50$	& \cellcolor{yellow} $ 57.50\pm 1.50$ & N/A\\

			\hline
			GS-SVM \cite{gao2019geometric} &  			$79.94\pm1.61$     &	$71.20\pm3.25$  & $48.73\pm2.32$ & \cellcolor{orange}	$89.65\pm1.94$	&  $ 53.33\pm1.37$ & \cellcolor{orange} $45.23\pm1.25$\\

			GGSN+EK [OURS]&  			\cellcolor{orange} 	$80.34\pm 1.68$     &	$73.20\pm 3.76$  &  $49.47 \pm 2.27$& \cellcolor{yellow}	$91.60 \pm 1.97 $	& \cellcolor{orange} $56.89 \pm 2.24 $ &\cellcolor{green}$49.03\pm 1.58$\\

			\hline
		\end{tabular}
	}
	
	\label{classify}
\end{table}

\paragraph{Regression of Quantum Chemical Energies:}
In order to showcase the prowess of both our higher order scattering scheme and our spectrum-agnostic aggregation method of Section \ref{aggro1}, we combine these building blocks into a hybrid architecture  which we then apply in combination with kernel methods ($2$GGST + EK) to the task of atomization energy regression on QM$7$. This is a comparatively small dataset of $7165$ molecular graphs, taken from the $970$ million strong molecular database GDB-$13$ \cite{blum}. Each graph in QM$7$ represents an organic molecule, with nodes corresponding to individual atoms. Beyond the node-level information of atomic charge, there is also edge level information characterising interaction strengths between individual nodes/atoms available. This is encoded into so called Coulomb matrices (see e.g. \cite{rupp} or Appendix \ref{AddEx}) of molecular graphs, which for us serve a dual purpose: On the one hand we consider a Coulomb matrix as an edge-level input signal on a given graph. On the other hand, we also treat it as an adjacency matrix  from which we build up a graph Laplacian $\mathcal{L}$. Our normal operator is then chosen as $\mathscr{L} = \mathcal{L}/\lambda_{\textit{max}}(\mathcal{L})$ again.  Connecting operators 
are set to the identity, while non-linearities are fixed to $\rho_{n \geq 1}(\cdot) = |\cdot|$.  Filters are chosen as  $(\sin(\pi/2\cdot \mathscr{L}),$ $ \cos(\pi/2\cdot \mathscr{L}) , \sin(\pi\cdot \mathscr{L}), \cos(\pi\cdot \mathscr{L}))  $ acting 
through matrix multiplication. Output generating functions are set to the identity and depth is $N=4$, so that we essentially recover Architecture II of 
\begin{minipage}{0.5\textwidth}
	\includegraphics[scale=0.43]{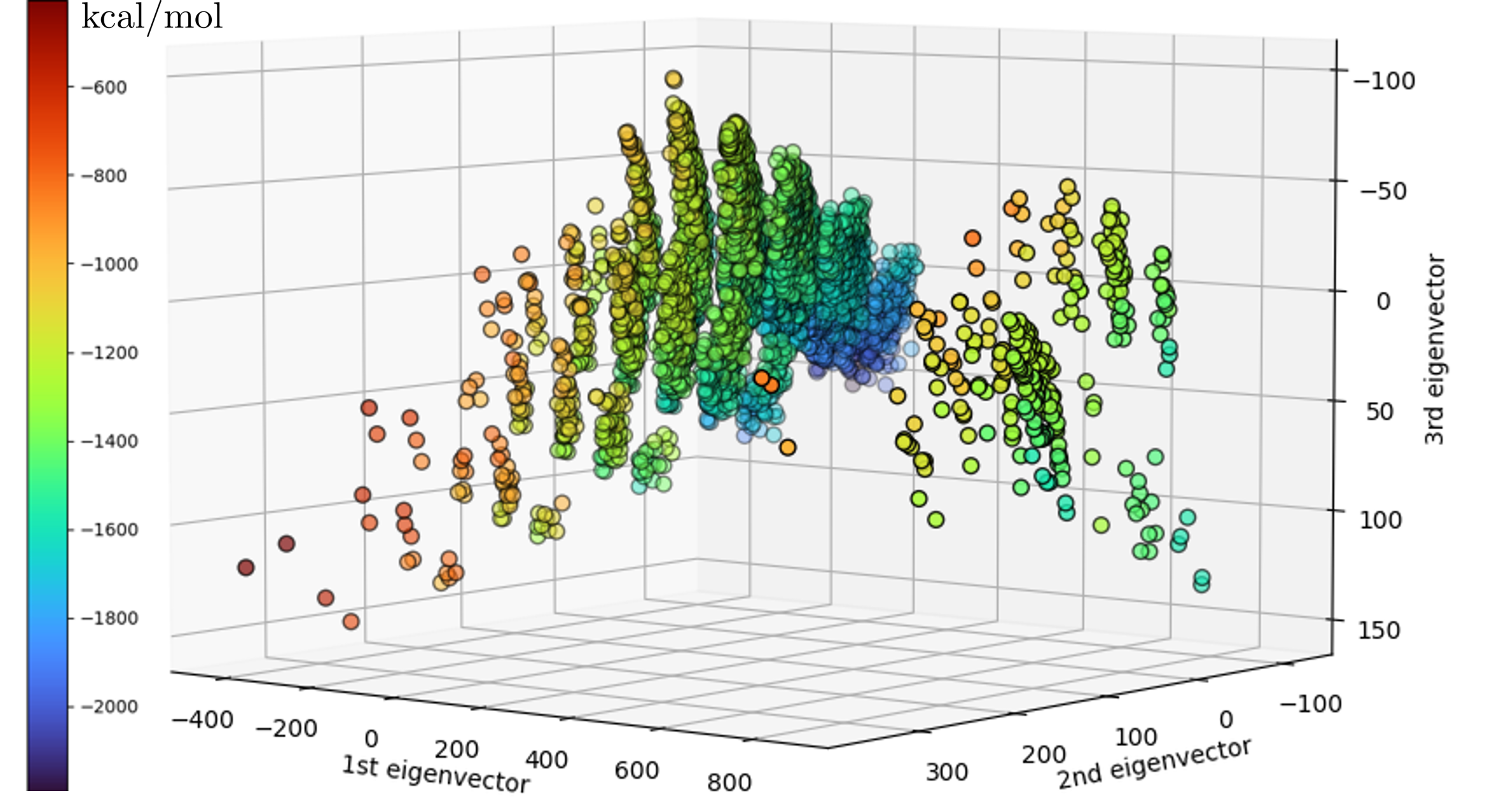}
	\captionof{figure}{Atomization Energy as a Function of primary Principal Components of  Scattering Features } 
	\label{QM7ScatterPlot}
\end{minipage}
\hfill
\begin{minipage}{0.45\textwidth}
	 

Fig. \ref{BothArchitectures}; now applied to edge-level input. Graph level features are aggregated via the map $N^{E}_5$ of Section \ref{HOS}. We chose $p=5$  (and not $p\gg 5$) for $N^{E}_p$ to avoid overfitting. Generated feature vectors are combined with node level scattering features obtained from applying Architecture II of Fig. \ref{BothArchitectures} to the input signal of atomic charge into composite feature vectors; plotted in  Figure \ref{QM7ScatterPlot}. As is visually evident, even when reduced to the low-dimensional subspace of their first three principal components, the generated scattering features are able to aptly   resolve the atomization energy of the molecules.  
This aptitude is also reflected in 	Table  \ref{regs},  comparing our 
\end{minipage}
approach with leading graph-based learning methods trained with ten-fold cross validation on node  
 \begin{minipage}{0.61\textwidth}
 and (depending 	on the model) edge level information. 
Our method  is the best performing.  We significantly outperform the next best model (DTNN), producing less than half of its mean absolute error (MAE). Errors of other methods are at least one --- sometimes two --- orders of magnitude greater. In part, this performance discrepancy might be explained by the hightened suitability of our scattering transform for environments with somewhat limited training-data availability. Here we speculate that the additional performance gap might be explained by the fact that our graph shift  operator $\Delta$ carries the same information as the Coulomb matrix (a proven molecular graph descriptor in itself \cite{rupp}). Additionally, our filters  being infinite series' in powers of the underlying normal operator allows for rapid dispersion of information  across underlying molecular graphs, as opposed to e.g. the filters in GraphConv
\end{minipage}\hfill
\begin{minipage}{0.37\textwidth}
	\begin{table}[H]
		\small
		\centering
		\caption{Comparison of Methods }
		\scalebox{0.9}{
			\begin{tabular}{ p{2.7cm}p{2.3cm}  }
				\hline
				\textbf{Method} &	\textbf{MAE} [kcal/mol] \\
				\hline
				AttentiveFP \cite{pushing} &  			$66.2 \pm 2.8$     \\
				DMPNN \cite{ALMR} &  		$	105.8 \pm 13.2$     \\	
				DTNN \cite{molnet} &  	\cellcolor{yellow} 		$ 8.2 \pm 3.9$     \\
				GraphConv \cite{Kipf}   & $118.9 \pm 20.2 $  	\\
				
				GROVER (base)\cite{grover} &  		$	72.5\pm 5.9 $     \\	
				MPNN \cite{mpnncm} &  		$	113.0 \pm 17.2 $     \\				
				
				N-GRAM\cite{ngram} &  		$	125.6 \pm 1.5 $     \\		
				PAGTN (global) \cite{patgn} &  			$47.8\pm 3.0 $     \\	
				
				PhysChem \cite{quantmet} &  			$59.6\pm 2.3 $     \\

				SchNet \cite{sch} &  			$74.2 \pm 6.0 $     \\
				
				Weave \cite{kearnes2016molecular} &  			$59.6\pm 2.3 $     \\
				
				GGST+EK [OURS] &  \cellcolor{orange}	$	11.3 \pm 0.6$     \\	
				
				$2$GGST+EK [OURS] &  	\cellcolor{green} 	$	3.4 \pm 0.3$     \\

				\hline
			\end{tabular}
		}
		
		\label{regs}
	\end{table}
\end{minipage}
 or SchNet, which 
do not incorporate such higher powers. To quantify the effect of including second order scattering coefficients, we also include the result of performing kernel-regression solely on first order features generated through Architecture II of Fig. \ref{BothArchitectures} 
(GGST + EK). While results are still better than those of all but one leading approach, incorporating higher order scattering improves performance significantly.


%
%
%
%
%
%
%
\section{ Discussion}\label{conc}
Leaving behind the traditional reliance on graph wavelets, we developed a theoretically well founded framework for the design and analysis of (generalized) graph scattering networks; allowing for varying branching rations, non-linearities and filter banks. We provided spectrum independent stability guarantees,
covering changes in input signals  and for the first time also arbitrary normal perturbations in the underlying graph-shift-operators. After introducing a new framework to quantify vertex-set non-preserving changes in graph  domains, we obtained spectrum-independent stability guarantees for this setting too. We provided conditions for energy decay and discussed implications for truncation stability. Then we introduced a new method of graph-level feature aggregation and extended scattering networks to higher order input data. Our numerical experiments showed that a simple scattering transform conforming to our framework is able to outperform the traditional graph-wavelet based approach to graph scattering in social network graph classification tasks. On complex datasets our method is also competitive with current fully trainable methods, ouperforming all competitors on REDDIT-$12$K. Additionally, higher order graph scattering transforms significantly outperform current leading graph-based learning methods in predicting atomization energies on QM$7$.
A reasonable critique of  scattering networks as tractable models for general graph convolutional networks is their inability to emulate non-tree-structured network topologies. While transcending the wavelet setting has arguably diminished the conceptual gap between the two architectures, this structural difference persists. Additionally we note that despite a provided promising example, it is not yet clear whether the newly introduced graph-perturbation framework can aptly provide stability guarantees to all reasonable coarse-graining procedures. Exploring this question is the subject of ongoing work.

\section*{Broader Impact}
We caution against an over-interpretation of established mathematical guarantees: Such guarantees do not negate biases that may be inherent to utilized datasets.

\section*{Disclosure of Funding}
Christian Koke acknowledges support from the German Research Foundation through the MIMO II-project (DFG SPP 1798, KU 1446/21-2).
Gitta Kutyniok acknowledges support from the ONE Munich Strategy Forum (LMU Munich, TU Munich, and the Bavarian Ministery for Science and Art), the Konrad Zuse School of Excellence in Reliable AI (DAAD), the Munich Center for Machine Learning (BMBF) as well as the German Research Foundation under Grants DFG-SPP-2298, KU 1446/31-1 and KU
1446/32-1 and under Grant DFG-SFB/TR 109, Project C09 and the Federal Ministry of Education and Research under Grant MaGriDo.



%
%



\bibliographystyle{plain}
\bibliography{Bibliography}
\medskip

\section*{Checklist}


\begin{enumerate}

\item For all authors...
\begin{enumerate}
  \item Do the main claims made in the abstract and introduction accurately reflect the paper's contributions and scope?
    \answerYes{A discussion of how and in which order the main claims made in  Abstract and Introduction were substantiated within the paper is a main focus of Section \ref{conc}}
  \item Did you describe the limitations of your work?
    \answerYes{This is a second focus of Section \ref{conc}.}
  \item Did you discuss any potential negative societal impacts of your work?
    \answerYes{This is part of Section \ref{conc}.}
  \item Have you read the ethics review guidelines and ensured that your paper conforms to them?
    \answerYes{}
\end{enumerate}

\item If you are including theoretical results...
\begin{enumerate}
  \item Did you state the full set of assumptions of all theoretical results?
  \answerYes{Every Theorem and Lemma is stated in a mathematically precise way, with all underlying assumptions included. Additionally, Appendix \ref{LA} briefly reviews some terminology that might not be immediately present in every readers mind, but is utilized in order to be able to state theoretical results in a precise and concise manner.  }
        \item Did you include complete proofs of all theoretical results?
 \answerYes{This is the Focus of Appendices \ref{opfrpf}, \ref{sptcy}, \ref{sdimoppertpf}, \ref{cplpf}, \ref{BigTpf}, \ref{somelabelpf} and  \ref{Aggregation}. Results in Appendix \ref{HLSTB} are merely stated, as statements and corresponding proofs are in complete analogy (in fact almost verbatim the same) to previously discussed statements and proofs.}
\end{enumerate}

\item If you ran experiments...
\begin{enumerate}
  \item Did you include the code, data, and instructions needed to reproduce the main experimental results (either in the supplemental material or as a URL)?
   \answerYes{Yes; please see the supplementary material.}
  \item Did you specify all the training details (e.g., data splits, hyperparameters, how they were chosen)?
   \answerYes{This is the main focus of Appendix \ref{AddEx}.}
        \item Did you report error bars (e.g., with respect to the random seed after running experiments multiple times)?
  \answerYes{Errors for results of conducted experiments are included in Table \ref{classify} and Table \ref{regs} respectively.}
        \item Did you include the total amount of compute and the type of resources used (e.g., type of GPUs, internal cluster, or cloud provider)?
     \answerYes{This is described at the beginning of Appendix \ref{AddEx}}
\end{enumerate}

\item If you are using existing assets (e.g., code, data, models) or curating/releasing new assets...
\begin{enumerate}
  \item If your work uses existing assets, did you cite the creators?
    \answerYes{All utilized datasets were matched to the papers that introduced them (cf. Section \ref{AddEx}). Additionally,  we partially built on code corresponding to \cite{gao2019geometric} which we mentioned in Appendix \ref{AddEx}. }
  \item Did you mention the license of the assets?
    \answerYes{We mentioned in Appendix \ref{AddEx} that the code corresponding to \cite{gao2019geometric} is freely available under an Apache License.}
  \item Did you include any new assets either in the supplemental material or as a URL?
    \answerYes{Please see supplementary material.}
  \item Did you discuss whether and how consent was obtained from people whose data you're using/curating?
    \answerNA{}
  \item Did you discuss whether the data you are using/curating contains personally identifiable information or offensive content?
    \answerYes{For the utilized social network datasets, we mention in Appendix \ref{AddEx} that neither personally identifyable data nor content that might be considered offensive is utilised.}
\end{enumerate}

\item If you used crowdsourcing or conducted research with human subjects...
\begin{enumerate}
  \item Did you include the full text of instructions given to participants and screenshots, if applicable?
    \answerNA{}
  \item Did you describe any potential participant risks, with links to Institutional Review Board (IRB) approvals, if applicable?
    \answerNA{}
  \item Did you include the estimated hourly wage paid to participants and the total amount spent on participant compensation?
    \answerNA{}
\end{enumerate}

\end{enumerate}


\appendix

\appendix

\section{Some Concepts in Linear Algebra }\label{LA}
In the interest of self-containedness, we provide a brief review of some concepts from linear algebra utilized in this work that might potentially be considered more advanced. Presented results are all standard; a very thorough reference is \cite{RS}.

\paragraph{Hilbert Spaces:} To us, a Hilbert space --- often denoted by $\mathcal{H}$ --- is a vector space over the complex numbers which also has an inner product --- often denoted by $\langle\cdot,\cdot\rangle_{\mathcal{H}}$. Prototypical examples are given by the Euclidean spaces $\mathds{C}^d$ with inner product $
\langle x,y\rangle_{\mathds{C}^d}:= \sum_{i=1}^d \overline{x}_iy_i $. Associated to an inner product is a norm, denoted by $\|\cdot\|_{\mathcal{H}}$ and defined by $\|x\|_{\mathcal{H}} := \sqrt{\langle x,x\rangle_{\mathcal{H}}}$ for $x \in \mathcal{H}$.

\paragraph{Direct Sums of Spaces:} Given two potentially different Hilbert spaces $\mathcal{H}$ and $\widehat{\mathcal{H}}$, one can form their direct sum $\mathcal{H} \oplus \widehat{\mathcal{H}}$. Elements of $\mathcal{H} \oplus \widehat{\mathcal{H}}$ are vectors of the form $(a,b)$, with $a\in \mathcal{H}$ and $b\in \widehat{\mathcal{H}}$. Addition and scalar multiplication  are defined in the obvious way by
\begin{equation}
(a,b)+\lambda (c,d) := (a+\lambda c, b+ \lambda d)
\end{equation}
for $a,c \in \mathcal{H}$, $b,d\in \widehat{\mathcal{H}}$ and $ \lambda \in \mathds{C}$. The inner product on the direct sum is defined by
\begin{equation}
\langle (a,b),(c,d)\rangle_{ \mathcal{H} \oplus \widehat{\mathcal{H}}} := \langle a,c\rangle_{\mathcal{H}} + \langle b,d\rangle_{\widehat{\mathcal{H}}}.
\end{equation}
As is readily checked, this implies that the norm $\|\cdot\|_{\mathcal{H} \oplus \widehat{\mathcal{H}}}$ on the direct sum is given by
\begin{equation}
\| (a,b)\|^2_{ \mathcal{H} \oplus \widehat{\mathcal{H}}} := \| a \|^2_{\mathcal{H}} + \| b\|^2_{\widehat{\mathcal{H}}}.
\end{equation}

Standard examples of direct sums are again the Euclidean spaces, where one has $\mathds{C}^d = \mathds{C}^n \oplus \mathds{C}^m$ if $m+n=d$, as is easily checked. One might also consider direct sums with more than two summands, writing $\mathds{C}^d = \oplus_{i=1}^d \mathds{C}$ for example. In fact, one might also consider infinite sums of Hilbert spaces: The space $\oplus_{i=1}^\infty\mathcal{H}_i$ is made up of those elements $
a =(a_1,a_2,a_3,...) $
with $a_i \in \mathcal{H}_i$ for which the norm
\begin{equation}
\|a\|^2 _{\oplus_{i =1}^\infty\mathcal{H}_i} := \sum\limits_{i=1}^\infty \|a_i\|^2_{\mathcal{H}_i}
\end{equation}
is finite. This means for example that the vector $(1,0,0,0,...)$ is in $\oplus_{i=1}^\infty\mathds{C}$, while $(1,1,1,1,...)$ is not.

\paragraph{Direct Sums of Maps:} Suppose we have two collections of Hilbert spaces $\{\mathcal{H}_i\}_{i=1}^\Gamma$, $\{\widetilde{\mathcal{H}}_i\}_{i=1}^\Gamma$ with $\Gamma \in \mathds{N}$ or $\Gamma = \infty$. Suppose further that for each $i\leq \Gamma$ (resp. $i < \Gamma$) we have a (not necessarily linear) map $J_i: \mathcal{H}_i \rightarrow \widetilde{\mathcal{H}}_i$. Then the collection $\{J_i\}_{i=1}^\Gamma$ of these 'component' maps induce a 'composite' map 
\begin{equation}
\mathscr{J}: \oplus_{i=1}^\Gamma \mathcal{H}_i \longrightarrow  \oplus_{i=1}^\Gamma \widetilde{\mathcal{H}}_i
\end{equation}
between the direct sums. Its value on an element $a = (a_1,a_2,a_3,...) \in \oplus_{i=1}^\Gamma \mathcal{H}_i$ is defined by
\begin{equation}
\mathscr{J} (a) = (J_1(a_1),J_2(a_2),J_3(a_3),...) \in \oplus_{i=1}^\Gamma \widetilde{\mathcal{H}}_i.
\end{equation}
Strictly speaking, one has to be a bit more careful in the case where $\Gamma = \infty$ to ensure that $\|\mathscr{J}(a)\|_{\oplus_{i=1}^\infty \widetilde{\mathcal{H}}_i} \neq \infty$. This can however be ensured if we have $\|J_i(a_i)\|_{\widetilde{\mathcal{H}}_i} \leq C \|a_i\|_{\mathcal{H}_i} $ for all $1\leq i $ and some $C$ independent of all $i$, since then  $\|\mathscr{J}(a)\|_{\oplus_{i=1}^\infty \widetilde{\mathcal{H}}_i} \leq C \|a\|_{\oplus_{i=1}^\infty \mathcal{H}_i}  \leq \infty$. If each $J_i$ is a linear operator, such a $C$ exists precisely if the operator norms (defined below) of all $J_i$ are smaller than some constant.

\paragraph{Operator Norm:} Let $J: \mathcal{H} \rightarrow \widetilde{\mathcal{H}}$ be a linear operator between Hilbert spaces. We measure its 'size' by what is called the operator norm, denoted by $\|\cdot\|_{op}$ and defined by
\begin{equation}
\|J\|_{op} := \sup\limits_{\psi \in \mathcal{H},\|\psi\|_{\mathcal{H}}=1}\frac{\|A\psi\|_{\widetilde{\mathcal{H}}}}{\|\psi\|_\mathcal{H}}.
\end{equation}

\paragraph{Adjoint Operators}
Let $J: \mathcal{H} \rightarrow \widetilde{\mathcal{H}}$ be a linear operator from the Hilbert space $\mathcal{H}$ to the Hilbert space $\widetilde{\mathcal{H}}$. Its adjoint $J^*: \widetilde{\mathcal{H}} \rightarrow \mathcal{H}$ is an operator mapping in the opposite direction. It is uniquely determined by demanding that
\begin{equation}
\langle Jf,u \rangle_{\widetilde{\mathcal{H}}} = \langle f, J^*u \rangle_{\mathcal{H}}
\end{equation}
holds true for arbitrary $f \in \mathcal{H}$ and $u \in \widetilde{\mathcal{H}}$.

\paragraph{Normal Operators:} If a linear operator $\Delta:\mathcal{H}\rightarrow\mathcal{H}$ maps from and to the same Hilbert space, we can compare it directly with its adjoint.  If $\Delta \Delta^* = \Delta^* \Delta$, we say that the operator $\Delta$ is normal. Special instances of normal operators are self-adjoint operators, for which  we have the stronger property $\Delta = \Delta^*$. If an operator is normal, there are unitary maps $U: \mathcal{H} \rightarrow \mathcal{H}$ diagonalizing $\Delta$ as
\begin{equation}
U^*\Delta U = \text{diag}(\lambda_1,...\lambda_n),
\end{equation}
with eigenvalues in $\mathds{C}$. We call the collection of eigenvalues the spectrum $\sigma(\Delta)$ of $\Delta$. If $\dim \mathcal{H} = d$, we may  write $\sigma(\Delta) = \{\lambda\}_{i=1}^d$. It is a standard exercise to verify that each eigenvalue satisfies $|\lambda_i|\leq \|\Delta\|_{op}$.  Associated to each eigenvalue is an eigenvector $\phi_i$. The collection of all (normalized) eigenvectors forms an orthonormal basis of $\mathcal{H}$. We may then write
\begin{equation}
\Delta f = \sum\limits_{i=1}^d \lambda_i\ \langle \phi_i,f\rangle_{\mathcal{H}} \phi_i.
\end{equation}

\paragraph{Resolvent of a (normal) Operator:}
Given a normal operator $\Delta$ on some Hilbert space $\mathcal{H}$, we have that the operator $(\Delta - z): \mathcal{H} \rightarrow \mathcal{H} $ is invertible precisely if $z \neq \sigma(\Delta)$. In this case we write
\begin{equation}
R(z,\Delta)  =  (\Delta - z)^{-1}
\end{equation}
and call this operator the \textbf{resolvent} of $\Delta$ at $z$. It can be proved that the norm of the resolvent satisfies
\begin{equation}
\|R(z,\Delta)\|_{op} = \frac{1}{\textit{dist}(z,\sigma(\Delta))},
\end{equation}
where $\textit{dist}(z,\sigma(\Delta))$ denotes the minimal distance between $z$ and any eigenvalue of $\Delta$.

\paragraph{Functional Calculus:} Given a normal operator $\Delta: \mathcal{H} \rightarrow \mathcal{H}$ on a Hilbert space of dimension $d$ and a complex function $g: \mathds{C} \rightarrow \mathds{C}$,
we can define another normal operator obtained from applying the function $g$ to $\Delta$ by
\begin{equation}
g(\Delta)f = \sum_{i=1}^{f} g(\lambda_i)\langle\phi_i,f\rangle_{\mathcal{H}}\phi_i.
\end{equation}
For example if $g(\cdot)=|\cdot|$,  we obtain the absolute value $|\Delta|$ of $\Delta$ by specifying for all $f \in \mathcal{H}$ that
\begin{equation}
|\Delta|f = \sum_{i=1}^{d} |\lambda_i|\langle\phi_i,f\rangle_{\mathcal{H}}\phi_i.
\end{equation}
Similarly we find (if $z \notin \sigma(\Delta)$ and for $f \in \mathcal{H}$) 
\begin{equation}
\frac{1}{\Delta - z} = \sum_{i=1}^{d} \frac{1}{\lambda_i - z}\langle\phi_i,f\rangle_{\mathcal{H}}\phi_i = (\Delta-z)^{-1} = R(z,\Delta)
\end{equation}
where we think of the left-hand-side as applying a function to $\Delta$, while we think of the right-hand-side as inverting the operator $(\Delta - z)$.\
This now allows us to apply tools from complex analysis also to operators:
If a function $g$ is analytic (i.e. can be expanded into a power series), we have
\begin{equation}
g(\lambda) = -\frac{1}{ 2 \pi i}\oint_{S} \frac{g(z)}{\lambda - z} dz
\end{equation}
for any circle $S \subseteq \mathds{C}$ encircling $\lambda$ by Cauchy's integral formula. Thus, if we chose $S$ large enough to encircle the entire spectrum $\sigma(\Delta)$, we have
\begin{equation}
g(\Delta)f = -\sum\limits_{i=1}^d\frac{1}{ 2 \pi i}\oint_{S} \frac{g(z)}{\lambda_i - z} dz \langle\phi_i,f\rangle_{\mathcal{H}}\phi_i = -\frac{1}{2 \pi i} \oint_{S} g(z) R(z,\lambda)dz.
\end{equation}

\paragraph{Frobenius Norm:}
Given a finite dimensional Hilbert space $\mathcal{H}$ with inner product $\langle\cdot , \cdot\rangle_{\mathcal{H}}$, and an orthonormal basis $\{\phi_i\}_{i=1}^d$, we define the trace of an operator $A:\mathcal{H}\rightarrow\mathcal{H}$ as
\begin{equation}
\textit{Tr}(A):=\sum\limits_{k=1}^d\langle\phi_k,A\phi_k\rangle_{\mathcal{H}}.
\end{equation}
It is a standard exercise to show that this is independent of the choice of orthonormal basis. The associated Frobenius inner product on the space of operators is then given as
\begin{equation}
\langle B,A\rangle_F:= \textit{Tr}(B^*A)\sum\limits_{k=1}^d\langle\phi_k,B^*A\phi_k\rangle_{\mathcal{H}}.
\end{equation}
Hence the Frobenius norm of an operator is determined by
\begin{equation}
\|A\|_F^2 = \textit{Tr}(A^*A)=\sum\limits_{k=1}^d\langle\phi_k,A^*A\phi_k\rangle_{\mathcal{H}}.
\end{equation}
It is a standard exercise to verify that we have $\|A\|_{op}\leq \|A\|_{F}$. 
Since the trace is independent of the choice of orthonormal basis, the Frobenius norm is invariant under unitary transformations. More precisely, if $U,V:\mathcal H \rightarrow \mathcal H$ are unitary, we have
\begin{equation}
\|UAV\|_F^2 = \|A\|_F^2.
\end{equation}

Frobenius norms can be used to transfer Lipschitz continuity properties of complex functions to the setting of functions applied to normal operators:

\begin{Lem}\label{A1result}
	Let $g:\mathds{C}\rightarrow\mathds{C}$ be Lipschitz continuous with Lipschitz constant $D_g$.  This implies
	\begin{equation}
	\|g(X)-g(Y)\|_F\leq D_g\cdot\|X-Y\|_F.
	\end{equation}
	for normal operators $X,Y$ on $\mathcal{H}$.
\end{Lem}

\begin{proof}
This proof is taken (almost) verbatim from \cite{Wihler}.	For an operator $A:\mathcal{H}\rightarrow\mathcal{H} $ denote by $A_{ij}$ its matrix representation with respect to the orthonormal basis $\{\phi_i\}_{i=1}^d$:
	\begin{equation}
	A_{ij} := \langle\phi_i,A\phi_j\rangle_{\mathcal{H}}.
	\end{equation}
	We then have
	\begin{align}
	\|A\|^2_F = \sum\limits_{i,j=1}^d|A_{ij}|^2
	\end{align}
	as a quick calculation shows. Let now $U,W$ be unitary (with respect to the inner product $\langle\cdot,\cdot\rangle_{\mathcal{H}}$) operators diagonalizing the normal operators $X$ and $Y$ as
	\begin{align}
	V^*XV = \text{diag}(\lambda_1,...\lambda_n)=:D(X)\\
	W^*YW = \text{diag}(\mu_1,...\mu_n)=:D(Y).
	\end{align}
	Since the Frobenius norm is invariant under unitary transformations we find
	\begin{align}
	\|g(X) - g(Y)||_F^2 &=||g(VD(X)V^*) - g(WD(Y)W^*)\|_F^2\\
	& = \|Vg(D(X))V^* - Wg(D(Y))W^*\|_F^2\\
	& = \|W^*Vg(D(X)) - g(D(Y))W^*V\|_F^2\\
	&=  \sum\limits_{i,j=1}^d \left|(W^*Vg(D(X)) - g(D(Y))W^*V)_{ij} \right|^2 \\
	&=  \sum\limits_{i,j=1}^d \left|\sum\limits_{k=1}^n[W^*V]_{ik}[g(D(X))]_{kj} - [g(D(Y))]_{ik}[W^*V]_{kj} \right|^2 \\
	&= \sum\limits_{i,j=1}^d\left| [W^*V]_{ij}\right|^2 |g(\lambda_j) - g(\mu_i)|^2\\
	&\leq \sum\limits_{i,j=1}^d\left| [W^*V]_{ij}\right|^2 D^2_g|\lambda_j - \mu_i|^2\\
	&= D^2_g \sum\limits_{i,j=1}^d \left|\sum\limits_{k=1}^n[W^*V]_{ik}[D(X)]_{kj} - [D(Y)]_{ik}[W^*V]_{kj} \right|^2 \\
	&=D^2_g \|X-Y \|_F^2.
	\end{align}

\end{proof}

\section{Proof of Theorem \ref{opfr}}\label{opfrpf}

\begin{Thm} Let $\Delta : \ell^2(G) \rightarrow \ell^2(G)$ be normal. 	If the family $ \{g_{i}(\cdot)\}_{i \in  I}$
	of bounded  functions satisfies $	A \leq \sum_{i \in  I} |g_i(c)|^2 \leq B$ for all $c$ in the spectrum $\sigma(\Delta)$,  we have ($\forall f \in \ell^2(G)$)
	\begin{align}
	A \|f\|^2_{\ell^2(G)} \leq  \sum\limits_{i \in  I} \|g_i(\Delta)f\|^2_{\ell^2(G)} \leq B \|f\|_{\ell^2(G)}^2.
	\end{align}
\end{Thm}
\begin{proof}
	Writing the normalized eigenvalue-eigenvector sequence of $\Delta$ as $(\lambda_i,\phi_i)_{i=1}^{|G|}$,	we simply note 
	\begin{equation}
	\sum\limits_{i \in  I} \sum\limits_{k=1}^{|G|}| \langle g_i(\lambda_k)\phi_k,f\rangle_{\ell^2(G)}|^2 =  \sum\limits_{k=1}^{|G|}\left(\sum\limits_{i \in I}  |g_i(\lambda_k)|^2\right)|\langle\phi_k,f\rangle_{\ell^2(G)}|^2.
	\end{equation}
	Now under the assumption, we can estimate the sum in brackets by $A$ from below and by $B$ from above. Then we need only use Bessel's (in)equality to prove
	\begin{equation}
	A||f||^2 \leq \sum\limits_{i \in \widehat I} \sum\limits_{k=1}^{|G|} |\langle g_i(\lambda_k)\phi_k,f\rangle_{\ell^2(G)}|^2 \leq B ||f||^2.
	\end{equation}	
\end{proof}

\section{Proof of Theorem \ref{signalpertcty}}\label{sptcy}
\begin{Thm}
	With the notation of Section \ref{GGST} and setting $B_0=1$, we have:
	\begin{align}
	\|\Phi_N(f) - \Phi_N(h)\|^2_{\mathscr{F}_N} \leq \left(1+ \sum\limits_{n=1}^N\max\{[B_n(L_n^+R_n^+)^2-1],0\}\prod\limits_{k=0}^{n-1}B_k(R^+_k L^+_k)^2\right)\|f-h\|^2_{\ell^2(G)}
	\end{align}
\end{Thm}
To streamline the argumentation let us first introduce some notation:
\begin{Not}\label{cc2} Let us denote paths in $\Gamma^N$ as $
	q := (\gamma_N,...,\gamma_1)$.
	For $f \in \ell^2(G)$ let us write
	\begin{equation}
	f_q := U[\gamma_N]\circ...\circ U[\gamma_1](f).
	\end{equation}
	
\end{Not}

\begin{proof}
	By Definition, we have
	\begin{align}
	\|\Phi_N(f) - \Phi_N(g)\|^2_{\mathscr{F}_N} =& \sum\limits_{n=1}^N \left(\sum\limits_{q \in \Gamma^{n-1}} \|V_n(f_q) - V_n(h_q) \|^2_{\ell^2(G_n)}\right)\\
	=&\sum\limits_{n=1}^N \underbrace{\left(\sum\limits_{q \in \Gamma^{n-1}} \|\chi_n(\Delta_n)\rho_n(P_n(f_q)) - \chi_n(\Delta_n)\rho_n(P_n(h_q)) \|^2_{\ell^2(G_n)}\right)}_{=:a_n}.
	\end{align}

	We proceed in two steps:\\
	Our initial goal is to upper bound $a_n$ as
	\begin{equation}\label{firstblood}
	a_n \leq B_n(L^+_n  R^+_n)^2 \cdot b_{n-1} - b_n \equiv (b_{n-1}-b_n) + \left[B_n(L^+_n  R^+_n)^2 -1 \right] \cdot b_{n-1}
	\end{equation}
	for 
	$
	b_{n} := \sum_{q \in \Gamma^n} \|f_q - h_q\|^2_{\ell^2(G_n)}$ with $b_0 = \|f-h\|^2_{\ell^2(G)} $. 
	To achieve this we note that (\ref{firstblood}) is equivalent to
	\begin{equation}
	a_n + b_n \leq B_n(L^+_n  R^+_n)^2 \cdot b_{n-1}
	\end{equation}
	which upon unraveling definitions may be written as
	\begin{equation}\label{toprove}
	\begin{split}
	&\sum\limits_{q \in \Gamma^{n-1}} \|\chi_n(\Delta_n)\rho_n(P_n((f_q))) - \chi_n(\Delta_n)\rho_n(P_n(h_q) \|^2_{\ell^2(G_n)} + \sum_{\widehat{q} \in \Gamma^n} \|f_{\widehat{q}} - h_{\widehat{q}}\|^2_{\ell^2(G_n)}\\
	\leq& B_n(L^+_n  R^+_n)^2\sum_{q \in \Gamma^{n-1}} \|f_q - h_q\|^2_{\ell^2(G_{n-1})} .
	\end{split}
	\end{equation}
	To establish (\ref{toprove}), we note, that in the sum over paths of length $n$, any $\widehat{q}\in \Gamma^n$ can uniquely be written as $\widehat{q} = (\gamma_n,q)$, with the path $q \in \Gamma^{n-1}$ of length $(n-1)$ determined by
	\begin{equation}
	\widehat{q} = (\gamma_n,\underbrace{\gamma_{n-1},...,\gamma_1}_{=:q}).
	\end{equation}
	With this we find
	\begin{align}
	\sum_{\widehat{q} \in \Gamma^n} \|f_{\widehat{q}} - h_{\widehat{q}}\|^2_{\ell^2(G_n)} = \sum\limits_{\gamma_n \in \Gamma_n}\sum\limits_{q \in \Gamma^{n-1}} \|g_{\gamma_n}(\Delta_n)\rho_n(P_n((f_q))) - g_{\gamma_n}(\Delta_n)\rho_n(P_n(h_q)) \|^2_{\ell^2(G_n)}.
	\end{align}
	Thus we can rewrite the left hand side of (\ref{toprove}) as
	\begin{align}
	&\sum\limits_{q \in \Gamma^{n-1}} \|\chi_n(\Delta_n)\rho_n(P_n((f_q))) - \chi_n(\Delta_n)\rho_n(P_n(h_q) \|^2_{\ell^2(G_n)} + \sum_{\widehat{q} \in \Gamma^n} \|f_{\widehat{q}} - h_{\widehat{q}}\|^2_{\ell^2(G_n)}\\
	=&\sum\limits_{q \in \Gamma^{n-1}}\bigg(\|\chi_n(\Delta_n)\rho_n(P_n(f_q)-\chi_n(\Delta_n)\rho_n(P_n(h_q) \|^2_{\ell^2(G_n)} \\
	&+\left.   \sum\limits_{\gamma_n \in \Gamma_n}  \|g_{\gamma_n}(\Delta_n)\rho_n(P_n((f_q))) - g_{\gamma_n}(\Delta_n)\rho_n(P_n(h_q)) \|^2_{\ell^2(G_n)}     \right)\\
	=:& \star
	\end{align}

	The fact that in each layer the function $\{\chi_n(\cdot)\}\bigcup\{g_{\gamma_n}(\cdot)\}_{\gamma_n \in  \Gamma_n} $ form a generalized frame with upper frame constant $B_n$ implies by Theorem \ref{opfr}, that we can further bound this as
	\begin{align}
	\star\leq B_n  \sum\limits_{q \in \Gamma^{n-1}}\|\rho_n(P_n(f_q)-\rho_n(P_n(h_q) \|^2_{\ell^2(G_n)} .
	\end{align}
	Using the Lipschitz continuity of $\rho_n$ and $P_n$, we arrive at the desired expression (\ref{toprove}).

	Having established that
	\begin{equation}
	a_n \leq (b_{n-1}-b_n) + \left[B_n(L^+_n  R^+_n)^2 -1 \right] \cdot b_{n-1}
	\end{equation}
	holds true, we note that we can establish 
	\begin{equation}
	b_{n-1} \leq \prod\limits_{k=1}^{n-1} B_k(L_k^+R_k^+)^2 b_{n-2}
	\end{equation}
	arguing similarly as in the case of (\ref{toprove}) by using (for $f \in \ell^2(G_{n-1})$)
\begin{equation}
\sum\limits_{\gamma_{n-1} \in  \Gamma_{n-1}} \|g_{\gamma_{n-1}}(\Delta_{n-1})f\|^2_{\ell^2(G_{n-1})} \leq \|\chi_{n-1}(\Delta_{n-1})f\|^2_{\ell^2(G_{n-1})} + \sum\limits_{\gamma \in  \Gamma} \|g_{\gamma_{n-1}}(\Delta_{n-1})f\|^2_{\ell^2(G_{n-1})}
\end{equation}
together with the frame property and Lipschitz continuities. We then iterate this inequality and recall that $b_0 = \|f-h\|^2_{\ell^2(G)} $.
	Using the fact that
	\begin{equation}
	\sum\limits_{n=1}^N (b_{n-1} - b_{n}) = b_0 - b_N \leq b_0,
	\end{equation}
	we finally find
	\begin{align}
	\|\Phi_N(f) - \Phi_N(h)\|^2_{\mathscr{F}_N} \leq \left(1+ \sum\limits_{n=1}^N\max\{[B_n(L_n^+R_n^+)^2-1],0\}\prod\limits_{k=0}^{n-1}B_k(R^+_k L^+_k)^2\right)\|f-h\|^2_{\ell^2(G)}.
	\end{align}

\end{proof}
\section{Proof or Theorem \ref{sdimoppert}}\label{sdimoppertpf}
\begin{Thm}
	Let $\Phi_N$ and $\widetilde{\Phi}_N$ be two  scattering transforms  based on the same module sequence $\Omega_N$ and operator sequences
	$\mathscr{D}_N,\widetilde{\mathscr{D}}_N$  with the same connecting operators ($P_n = \widetilde P_n$) in each layer. Assume $R^+_n,L^+_n\leq 1$ and $B_n\leq B$ for some $B$ and $n \leq N$. Assume that the respective normal operators satisfy $\|\Delta_n - \widetilde \Delta_n\|_F \leq \delta$ for some $\delta >0$. Further assume that  the  functions $\{g_{\gamma_n}\}_{\gamma_n \in \Gamma_n}$ and $\chi_n$ in each layer  are Lipschitz continuous with associated Lipschitz constants
	satisfying $L_{\chi_n}^2+\sum_{\gamma_n \in \Gamma_n} L_{g_{\gamma_n}}^2\leq D^2$ for all $n \leq N$ and some $D>0$. Then we have 
	\begin{equation}
	\|\widetilde \Phi_N(f)  -\Phi_N(f) \|_{\mathscr{F}_N} \leq  \sqrt{2(2^{N}-1)  }\cdot\sqrt{(\max\{B,1/2\})^{N - 1}} \cdot D \cdot \delta \cdot	 \|f\|_{\ell^2(G)}
	\end{equation}
	for all $f\in \ell^2(G)$. If $B \leq 1/2$, the stability constant improves to $\sqrt{2(1 - B^N)/(1 - B)} \cdot D \leq 2 \cdot D $.
\end{Thm}
%
\begin{Not}\label{ddd3}
	Let us denote scattering propagators based on operators $\Delta_n$ and connecting operators $P_n$ by $U_n$ and scattering propagators based on 	operators $\widetilde{\Delta}_n$ by $\widetilde{U}_n$. Similarly, to Notation \ref{cc2}, let us then write (with $q = (\gamma_N,...,\gamma_1)$)
	\begin{equation}
	\widetilde f_q := \widetilde{U}_n[\gamma_n]\circ...\circ\widetilde{U}_1[\gamma_1](f).	
	\end{equation}
\end{Not}
\begin{proof}
	By definition we have
	\begin{align}
	\|\Phi_N(f) - \widetilde{\Phi}_N \|_{\mathscr{F}_N}^2 =  \sum\limits_{n=1}^N \underbrace{\left(\sum\limits_{q \in \Gamma^{n-1}} \|\chi_n(\Delta_n)\rho_n(P_n((f_q))) - \chi_n(\widetilde\Delta_n)\rho_n(P_n(\widetilde f_q)) \|^2_{\ell^2(G_n)}\right)}_{=:a_n}.
	\end{align}
	We define $
	b_{n} := \sum_{q \in \Gamma^n} \|f_q - \widetilde{f}_q\|^2_{\ell^2(G_n)}$, with $b_0 = \|f-h\|^2_{\ell^2(G)} = 0$ and note 
	\begin{align}
	a_n + b_n =&\sum\limits_{q \in \Gamma^{n-1}}\bigg(\|\chi_n(\Delta_n)\rho_n(P_n(f_q)-\chi_n(\widetilde\Delta_n)\rho_n(P_n(\widetilde f_q) \|^2_{\ell^2(G_n)} \\
	&+\left.   \sum\limits_{\gamma_n \in \Gamma_n}  \|g_{\gamma_n}(\Delta_n)\rho_n(P_n((f_q))) - g_{\gamma_n}(\widetilde\Delta_n)\rho_n(P_n(\widetilde f_q)) \|^2_{\ell^2(G_n)}     \right).
	\end{align}
	Using (with $|a+b|^2\leq2(|a|^2+|b|^2)$)
	\begin{align}
	\frac12&\|g_{\gamma_n}(\Delta_n)\rho_n(P_n(f_q)) - g_{\gamma_n}(\widetilde{\Delta}_n)\rho_n(P_n(\widetilde{f}_q)) \|^2_{\ell^2(G_n)}\\ \leq& \|[g_{\gamma_n}(\Delta_n) - g_{\gamma_n}(\widetilde{\Delta}_n)]\rho_n(P_n(f_q)) \|^2_{\ell^2(G_n)}\\
	+& \|g_{\gamma_n}(\widetilde\Delta_n)[\rho_n(P_n((f_q))) - \rho_n(P_n(\widetilde{f}_q))] \|^2_{\ell^2(G_n)}\\
	\leq&\|[g_{\gamma_n}(\Delta_n) - g_{\gamma_n}(\widetilde{\Delta}_n)]\|^2_\infty\cdot\|\rho_n(P_n(f_q)) \|^2_{\ell^2(G_n)}\\
	+& \|g_{\gamma_n}(\widetilde\Delta_n)[\rho_n(P_n(f_q)) - \rho_n(P_n(\widetilde{f}_q))] \|^2_{\ell^2(G_n)},
	\end{align}
	and 
	\begin{equation}
	\|[g_{\gamma_n}(\Delta_n) - g_{\gamma_n}(\widetilde{\Delta}_n)]\|_\infty^2 \leq \|[g_{\gamma_n}(\Delta_n) - g_{\gamma_n}(\widetilde{\Delta}_n)]\|_F^2 \leq L^2_{g_\gamma}\cdot\delta^2
	\end{equation}
	 (c.f. Lemma \ref{A1result} ), we find
	\begin{align}
	a_n + b_n \leq & 2\sum_{q \in \Gamma^{n-1}} \left( L_{\chi_n}^2 + \sum\limits_{\gamma_n \in \Gamma_n} L_{g_{\gamma_n}^2}\right)(L^+_nR^+_n)^2\delta^2||\rho_n(P_n( f_q))||_{\ell^2(G_n)}^2\\
	+&2 \sum_{q \in \Gamma^{n-1}} B_n ||\rho_n(P_n( f_q))-\rho_n(P_n(\widetilde f_q))||_{\ell^2(G_n)}^2.
	\end{align}
	Using $L_{\chi_n}^2+\sum_{\gamma_n \in \Gamma_n} L_{\gamma_n}^2\leq D^2$, we then infer (using the  assumption $L^+_n,R^+_n\leq 1$)
	\begin{equation}
	a_n \leq (b_{n-1} -b_n ) + [2B - 1] b_{n-1} + B^{n-1}2 D^2\delta^2||f||_{\ell^2(G)}.
	\end{equation}
	Now if $B \leq \frac12$, we have 
	\begin{equation}
	a_n \leq (b_{n-1} -b_n ) +  B^{n-1}2 D^2\delta^2||f||_{\ell^2(G)}
	\end{equation}
	and results of geometric sums leads to the desired bound after summing over $n$.\\
Hence let us assume $B > \frac12$. 	Using similar arguments as before, we find
	\begin{align}
	b_{n-1} \leq& B^{n-2} 2  D^2 \delta^2 ||f||^2_{\ell^2(G)} + 2B b_{n-2} \leq  B^{n-2}2 D^2 \delta^2 ||f||^2_{\ell^2(G)} + B^{n-2} 4 D^2 \delta^2 ||f||^2_{\ell^2(G)}  + 4 b_{n-3}  \\
	\leq& B^{n-2} \left(\sum\limits_{k=1}^{n-1}2^k\right)D^2 \delta^2 ||f||_{\ell^2(G)}^2= B^{n-2} (2^n -2)D^2 \delta^2 ||f||_{\ell^2(G)}^2.
	\end{align} 
	Thus we now know
	\begin{equation}
	a_n \leq 2  D^2 \delta^2 B^{n - 1} ||f||_{\ell^2(G)}^2 + [2B - 1](2^n -2)D^2 \delta^2 B^{n - 2} ||f||_{\ell^2(G)}^2 +  (b_{n-1}-b_n)
	\end{equation}
	In total we find
	\begin{equation}
	\sqrt{\sum\limits_{n=1}^N a_n }\leq \sqrt{2(2^{N}-1)  }\cdot\sqrt{B^{N - 1}} \cdot D \cdot \delta \cdot	 \|f\|_{\ell^2(G)},
	\end{equation}
	where we have estimated the sum over $(b_{n-1}-b_n)$ by zero from above again. This establishes the claim.
\end{proof}

\begin{Rem}\label{stabremA}
	
	To see that this also holds for our Architecture I of Fig. \ref{BothArchitectures}, we note that the critical step is establishing that Lemma \ref{A1result} also applies to $\delta_0$ and $\overline{\cos}$, as defined in Section \ref{GSPFW}. Here we establish that
		\begin{equation}
	\|\delta_0(\Delta)-\delta_0(\widetilde \Delta)\|_F = 0
	\end{equation}
and 
		\begin{equation}
\|\overline{\cos}(\Delta)-\overline{\cos}(\widetilde \Delta)\|_F \leq D_{\cos} \|\Delta - \widetilde \Delta\|_F.
\end{equation}
Indeed, since $\Delta$ and $\widetilde \Delta$ are (possibly) rescaled graph Laplacians on the same graph, the spectral projections to their lowest lying eigen space, associated to the eigenvalue $\lambda_{\min}=0$ agree. Denoting this spectral projection by $\mathcal P$, we have
\begin{equation}
\overline{\cos}(\Delta)-\overline{\cos}(\widetilde \Delta) = [\cos(\Delta)-\mathcal P] - [\cos(\widetilde \Delta) - \mathcal P] = {\cos}(\Delta)-{\cos}(\widetilde \Delta)
\end{equation}
and we can apply Lemma \ref{A1result}. Similar considerations apply to $\delta_0$.
\end{Rem}

\section{Prototypical Example illustrating $\omega$-$\delta$ Closeness and $\delta$-Unitary Equivalence}\label{delta}

\begin{minipage}{0.68\textwidth}
	To investigate the example of Figure \ref{prettygraphdrawings}, we label the vertices of the respective graphs as depicted in Figure \ref{index}. We denote the left graph by $G$ and the right graph by $\widetilde G$.  The node-weights on $\widetilde G$ are given as $\widetilde \mu_i =1$ for $1\leq i \leq 7$, while on $G$ the weights are given as $\mu_i=1$ for $1\leq i \leq 5$ while $\mu_6 = 2$.
	We then consider the respective un-normalized graph Laplacians $\Delta: \ell^2(G) \rightarrow \ell^2(G)$ and $\widetilde \Delta: \ell^2(\widetilde G) \rightarrow \ell^2(\widetilde G)$, which for a given adjacency matrix $W$ on a graph signal space $\ell^2(G)$ with node weights $\{\mu_i\}_i$ is given as
	\begin{equation}
	(\Delta f)_i = \frac{1}{\mu_i}\sum\limits_{j}W_{ij}(f_i-f_j).
	\end{equation}
	Such operators are positive and hence $|\Delta|=\Delta$ (similarly for $\widetilde \Delta$).
	
	We now need to find operators $J:\ell^2(G)\rightarrow\ell^2(\widetilde G)$ and $\widetilde J:\ell^2(\widetilde G)\rightarrow\ell^2( G)$ satisfying the conditions of Definition \ref{TFDef}. 
	To construct $J$, we define a family $\{\psi_i\}_{i=1}^6$ of vectors on $\ell^2(\widetilde G)$ as 
	\begin{align}
	\psi_1 &= (1,0,0,0,0,0,0),\ \ \psi_2 = (0,1,0,0,0,0,0),\\
	\psi_3 &= (0,0,1,0,0,0,0),\ \ \psi_4 = (0,0,0,1,0,0,0),\\
	\psi_5 &= (0,0,0,0,1,0,0),\ \ \psi_6 = (0,0,0,0,0,1,1).\\
	\end{align}
\end{minipage}\hfill
\begin{minipage}{0.3\textwidth}
	
	\includegraphics[scale=0.7]{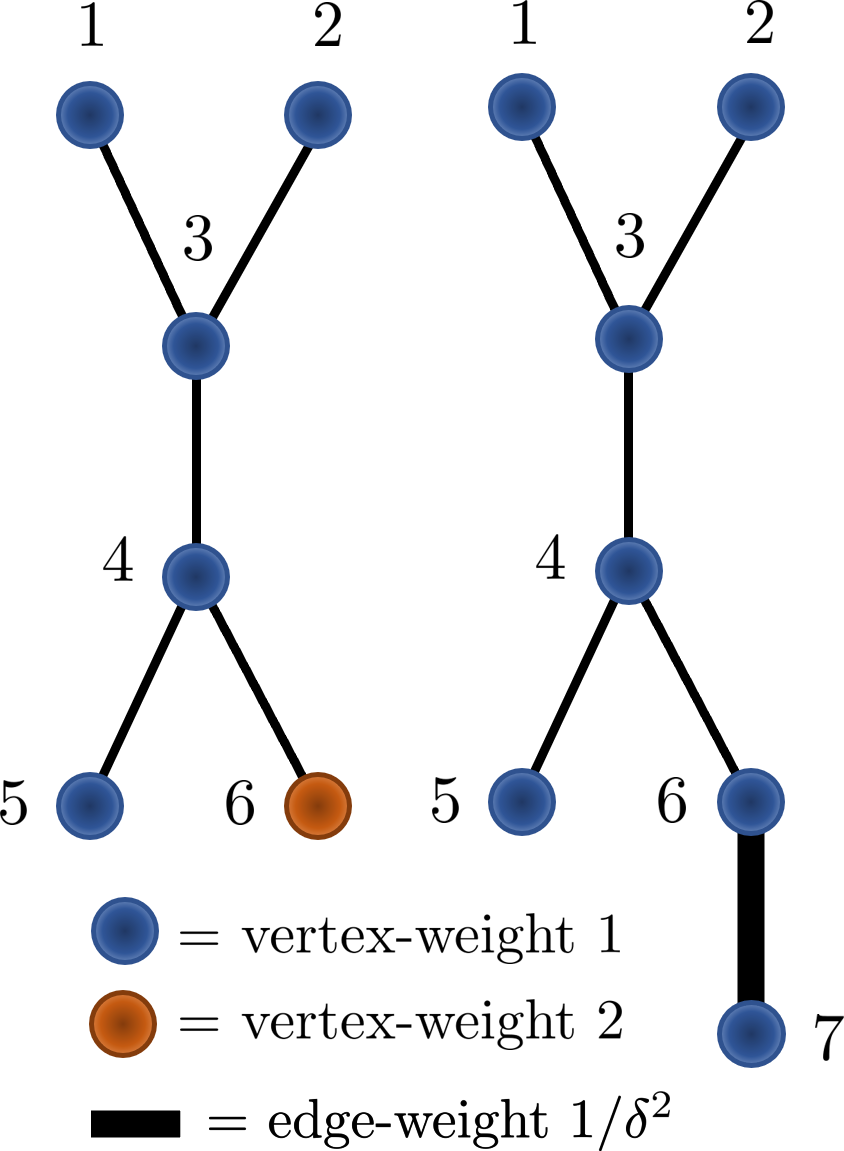}
	\captionof{figure}{Indexing on the respective graphs} 
	\label{index}
\end{minipage}
The map $J:\ell^2(G)\rightarrow\ell^2(\widetilde G)$ is then defined as 
\begin{equation}
Jf := \sum_{i=1}^6 f_i \psi_i,
\end{equation}
 for any $f \in \ell^2(G)$. We take $\widetilde J:\ell^2(\widetilde G)\rightarrow\ell^2( G)$ to be its adjoint ($\widetilde J := J^*$), which determined explicitly by
\begin{equation}
(\widetilde J u)_i = \frac{1}{\mu_i}\langle\psi_i,u\rangle_{\ell^2(\widetilde G)}
\end{equation}
for any $u \in \ell^2(\widetilde G)$ We shall now first check the conditions for $\delta$-quasi unitary equivalence, which we list again for convenience; now adapted to our current setting:
\begin{align}
\|Jf\|_{ \ell^2(\widetilde G)} \leq 2 \|f\|_{ \ell^2( G)},& \ \ \ \ \|(J - \widetilde {J}^* )f\|_{ \ell^2(\widetilde G)} \leq \delta  \|f\|_{ \ell^2( G)},\\
\| f - \widetilde J Jf\|_{ \ell^2( G)}^2 \leq \delta^2\left( \|f\|_{ \ell^2( G)}^2 + \langle f, \Delta,\ f\rangle_{\ell^2(G)} \right)  , & \ \ \ \   \|u - J \widetilde J u\|^2_{ \ell^2(\widetilde G)} \leq \delta^2 \left(\|u\|_{ \ell^2(\widetilde G)}^2 + \langle u,\widetilde{\Delta}\ u\rangle_{ \ell^2(\widetilde G)}\right).
\end{align}
We first note that since $\widetilde J = J^*$, we have $\|(J - \widetilde {J}^* )f\|_{ \ell^2(\widetilde G)} = 0$. Next we note
\begin{equation}
\|Jf\|_{ \ell^2(\widetilde G)}^2 = \sum\limits_{i=1}^7 |(Jf)_i|^2 = |f_6|^2 + \sum\limits_{i=1}^6 |f_i|^2  = \sum\limits_{i=1}^6 \mu_i = \|f\|_{ \ell^2( G)}^2.
\end{equation}
Furthermore we note
\begin{equation}
(\widetilde J Jf)_i = \sum\limits_{k=1}^6f_{k}\underbrace{\frac{1}{ \mu_i} \langle\psi_{i},\psi_k \rangle_{\ell^2(\widetilde G)}}_{=\delta_{ ik}} = f_i
\end{equation}
and hence $\| f - \widetilde J Jf\|_{ \ell^2( G)}^2 = 0$. It remains to control $\|u - J \widetilde J u\|^2_{ \ell^2(\widetilde G)}$. We note
\begin{align}
\widetilde J u = (u_1,u_2,u_3,u_4,u_5,(u_5+u_6)/2)^\top
\end{align}
and thus
\begin{align}
J\widetilde J u = (u_1,u_2,u_3,u_4,u_5,(u_6+u_7)/2,(u_6+u_7)/2)^\top,
\end{align}
Which implies
\begin{align}
u-J\widetilde J u = (0,0,0,0,0,(u_7-u_6)/2,(u_6-u_7)/2)^\top,
\end{align}
and thus 
\begin{equation}
\|u-J\widetilde J u\|_{\ell^2(\widetilde G)}^2 = 2\frac{|u_6-u_7|^2}{4} = \frac{|u_6-u_7|^2}{2}.
\end{equation}
We have
\begin{equation}
\langle u,\widetilde{\Delta}\ u\rangle_{ \ell^2(\widetilde G)} = \frac12 \sum\limits_{i,j=1}^d\widetilde W_{ij}|u_i-u_j|^2 .
\end{equation}
Since $\widetilde W_{67}= 1/\delta^2$ by assumption, we have 
\begin{align}
\|u-J\widetilde J u\|_{\ell^2(\widetilde G)}^2 &= \frac12|u_6-u_7|^2 = \frac12 \frac{\delta^2}{\delta^2}|u_6-u_7|^2 = \frac12 \delta^2 \widetilde W_{67}|u_6-u_7|^2\\
&\leq \frac12 \delta^2\sum\limits_{i,j=1}^d\widetilde W_{ij}|u_i-u_j|^2 = \delta^2 \langle u,\widetilde{\Delta}\ u\rangle_{ \ell^2(\widetilde G)} \\
&\leq \delta^2 \left(\|u\|_{ \ell^2(\widetilde G)}^2 + \langle u,\widetilde{\Delta}\ u\rangle_{ \ell^2(\widetilde G)}\right).
\end{align}
Thus we have proven $\delta$-unitary-equivalence and it remains to establish $(-1)$-$12\delta$ closeness. Combining Proposition 4.4.12. and Theorem 4.4.15 of \cite{PostBook}, instead of bounding $ 
\| (\widetilde R J  - J R) f\|_{\ell^2(\widetilde G)} \leq 12\delta\|f\|_{\ell^2(G)}
$ directly, we may instead establish that there are operators $J^1:\ell^2(G)\rightarrow \ell^2(\widetilde G) $, $\widetilde{J^1}:\ell^2(\widetilde G)\rightarrow \ell^2( G)$ satisfying
\begin{equation}\label{A}
\|J^1f - Jf\|_{\ell^2(\widetilde G)} \leq \delta^2\left(  \|f\|_{\ell^2(G)} + \langle f, \Delta,\ f\rangle_{\ell^2(G)}    \right),
\end{equation}
\begin{equation}\label{B}
\|\widetilde{J^1}u - \widetilde Ju\|_{\ell^2( G)} \leq \delta^2\left(  \|u\|_{\ell^2(\widetilde G)} + \langle f, \widetilde \Delta,\ u\rangle_{\ell^2(\widetilde G)}    \right),
\end{equation}
and 
\begin{equation}\label{C}
\langle J^1f, \widetilde \Delta\ u\rangle_{\ell^2(\widetilde G)}  = \langle f,  \Delta\ \widetilde{J^1}u\rangle_{\ell^2( G)}.
\end{equation}
We chose $J^1 = J$ and determine $\widetilde{J^1}$ by setting (for ($1\leq i \leq 6$))
\begin{equation}
(\widetilde{J^1}u)_i = u_i .
\end{equation}
Thus (\ref{A}) is clearly satisfied. For (\ref{B}) we note that we have
\begin{equation}
(\widetilde Ju - \widetilde{J^1}u) = (0,0,0,0,0,(u_7-u_6)/2).
\end{equation}
Thus we have
\begin{equation}
\|\widetilde{J^1}u - \widetilde Ju\|_{\ell^2( G)} = \frac12 |u_6-u_7|^2 \leq  \delta^2 \left(\|u\|_{ \ell^2(\widetilde G)}^2 + \langle u,\widetilde{\Delta}\ u\rangle_{ \ell^2(\widetilde G)}\right)
\end{equation}
as before. It remains to establish (\ref{C}). We have
\begin{align}
\langle f,  \Delta\ \widetilde{J^1}u\rangle_{\ell^2( G)} = \sum\limits_{i,j=1}^{6}\overline{f_i}W_{ij}(u_i-u_j), 
\end{align}
while we have
\begin{align}
\langle J^1f, \widetilde \Delta\ u\rangle_{\ell^2(\widetilde G)} &= \sum\limits_{i=1}^6 \overline{f_i}\cdot \langle \psi_i, \widetilde \Delta\ u\rangle_{\ell^2(\widetilde G)}\\
&=\sum\limits_{i,j=1}^{5} \overline{f_i}W_{ij}(U_j-u_i) + \overline{f_6}\cdot \langle \psi_6, \widetilde \Delta\ u\rangle_{\ell^2(\widetilde G)}.
\end{align}
We have (with all node-weights on $\ell^2(G)$ equal to unity)
\begin{align}
\langle \psi_6, \widetilde \Delta\ u\rangle_{\ell^2(\widetilde G)} = (\Delta\ u)_6 + (\Delta\ u)_7 &=  \left(\sum\limits_{j}W_{6j}(f_6-f_j) + \frac{1}{\delta^2}(f_6-f_7)\right) + \left(\frac{1}{\delta^2}(f_7-f_6)\right) \\
& = \left(\sum\limits_{j}W_{6j}(f_6-f_j) + \frac{1}{\delta^2}(f_6-f_7)\right)
\end{align}
And thus
\begin{equation}
\langle J^1f, \widetilde \Delta\ u\rangle_{\ell^2(\widetilde G)} = \sum\limits_{i,j=1}^{6}\overline{f_i}W_{ij}(u_i-u_j) = \langle f,  \Delta\ \widetilde{J^1}u\rangle_{\ell^2( G)}
\end{equation}
which proves the claim.
\section{Proof of Lemma \ref{cpl}}\label{cplpf}

\begin{Lem}
	In the setting of Definition \ref{TFDef} let $\Delta$ and $\widetilde \Delta$ be $\omega$-$\delta$-close and satisfy $\|\Delta\|_\textit{op},\|\widetilde \Delta\|_\textit{op}\leq K$ for some $K>0$.
	If $g:\mathds{C} \rightarrow\mathds{C}$  is holomorphic on the disk $B_{K+1}(0)$ of radius $(K+1)$, there is a constant $C_g \geq 0$ so that
	\begin{equation}
	\|g(\widetilde\Delta )J - J g(\Delta)\|_\text{op} \leq C_g \cdot \delta
	\end{equation} 
	with $C_g$ depending on $g$ , $\omega$ and $K$. 
\end{Lem}
\begin{proof}
	Without loss of generality, let us assume that $K > |\omega|$.
	Let us denote the circle of radius $r$ in $\mathds{C} $ by $S_r$.  For any holomorphic function $g$ and (normal) operator $\Delta$ whose spectrum is enclosed by the circle $S_r$, we can express the operator $g(\Delta)$ as
	\begin{equation}
	g(\Delta) = -\frac{1}{2 \pi i}\oint_{S_r} \frac{g(z)}{ \Delta - z} dz
	\end{equation}
as  discussed in Appendix \ref{LA} (see also \cite{Civin1960LinearOP} for more details). Note that in our case the resolvent $R(z,\Delta) = (\Delta - z)^{-1}$ is well defined for $|z| \geq K$, since with our assumptions all eigenvalues are within the circle of radius $K$. Additionally note that we have  
	\begin{equation}
	\textit{dist}(z ,\sigma(\Delta)) \geq \textit{dist}(z ,S_K) = |z| - K
	\end{equation} if $|z| \geq K$. The same holds true after replacing $\Delta$ with $\widetilde \Delta$. 
	Since for any normal operator $\Delta$ we have 
	\begin{equation}\label{fftheq}
	\|R(z,\Delta)\|_{\textit{op}} = 1/\textit{dist}(z ,\sigma(\Delta)),
	\end{equation} 
	we find
	\begin{equation}
	|R(z,\widetilde \Delta)\|_{op}, \ \|R(z,\Delta)\|_{op} \leq 1/(|z| - K) .
	\end{equation}
	To quantify the difference $\| R(z,\widetilde\Delta) J - J R(z,\Delta)\|_{op}$ in terms of the difference $\|\widetilde R(\omega) J - J R(\omega)\|_{op}\leq \delta$, we define the function
	\begin{equation}
	\gamma_0(z) := 1+\frac{|z - \omega|}{|z|-K},
	\end{equation}
	for which
	\begin{equation}
	\|  R(z,\widetilde \Delta) J - J  R(z, \Delta)\|_{op} \leq \gamma_0(z)^2 \| R(\omega,\widetilde \Delta) J - J  R(\omega,\widetilde \Delta)\|_{op} 
	\end{equation}
	holds, as proved (in more general form) in Lemma 4.5.9 in \cite{PostBook}. Since on $S_{K+1}$ we have and $|z-\omega|\leq 2K+1$ hence $\gamma_{0}(z) \leq 2(K+1) $, we find
	\begin{align}
	\|g(\widetilde\Delta )J - J g(\Delta)\|_\text{op} =& \left\| \frac{1}{2 \pi i}\oint_{S_{K+1}} g(z)\left( R(z,\widetilde \Delta) -  R(z, \Delta)\right)dz \right\|_{op}\\
	\leq&  \frac{1}{2 \pi }\oint_{S_{K+1}} |g(z)| \left\| R(z,\widetilde \Delta) -  R(z, \Delta) \right\|_{op}dz\\
	\leq& 2 \frac{(K+1)^2}{\pi}\left(\oint_{S_{K+1}} |g(z)| dz\right)\cdot \| R(\omega,\widetilde \Delta) J - J  R(\omega,\widetilde \Delta)\|_{op} .
	\end{align}
	Thus we may set 
	\begin{equation}
	C_g := 2\frac{(K+1)^2}{\pi}\oint_{S_{K+1}} |g(z)| dz.
	\end{equation}

\end{proof}

\section{Proof of Theorem \ref{BigT} }\label{BigTpf} 
We state and prove a somewhat more general theorem, incorporating also the case where the identification operators only almost commute with connecting operators or non-linearities. We also would like to point out that the constant $2$ in Definition \ref{TFDef} is arbitrary and any constant larger than one would suffice. Much more details are provided in Chapter IV of \citep{PostBook}.

\begin{Thm} 
	Let $\Phi_N, \widetilde{\Phi}_N$ be  scattering transforms based on a common module sequence $\Omega_N$ and differing operator sequences $\mathscr{D}_N, \widetilde{\mathscr{D}}_N$.  Assume $R^+_n,L^+_n\leq 1$ and $B_n\leq B$ for some $B$ and $n \geq 0$.
	Assume that there are identification operators $J_n: \ell^2(G_n) \rightarrow \ell^2(\widetilde G_n) $, $\widetilde{J}_n: \ell^2(\widetilde{G}_n) \rightarrow \ell^2( G_n) $ ($0\leq n \leq N$) so that the respective signal spaces are $\delta$-unitarily equivalent, the respective normal operators $\Delta_n,\widetilde{\Delta}_n$ are $\omega$-$\delta$-close as well as bounded (in norm) by $K>0$ and the connecting operators satisfy $\|\widetilde P_nJ_{n-1}f - J_n P_n f\|_{\ell^2(\widetilde G_n)} \leq \delta \|f\|_{\ell^2( G_{n-1})}$. For the common module sequence $\Omega_N$ assume  that the non-linearities  satisfy $\| \rho_n(J_{n}f) - J_n \rho_n(f)\|_{\ell^2(\widetilde G_n)} \leq \delta \| f\|_{\ell^2(G_n)}$ and that the constants $C_{\chi_n}$ and $\{C_{g_{\gamma_n}}\}_{\gamma_n \in \Gamma_N}$	associated through Lemma \ref{cpl} to the functions of the generalized frames in each layer satisfy $ C^2_{\chi_n} + \sum_{\gamma_n\in \Gamma_N} C^2_{g_{\gamma_n}} \leq D^2$ for some $D > 0$. Denote the operator that the family $\{J_n\}_n$ of identification operators induce on $\mathscr{F}_N$ through concatenation by
	$\mathscr{J}_N: \mathscr{F}_N \rightarrow \widetilde{\mathscr{F}}_N$. 
	Then we have with  $K_N = \sqrt{ (8^N-1) (2 D^2 +12B)/7\cdot B^{N-1}}$ if $B > 1/8$  and $K_N = \sqrt{  (2 D^2 +12B)\cdot (1 - B^N)/(1 - B)} $ if $B \leq 1/8$  that
	\begin{align}
		\|\widetilde{\Phi}_N(J_0f) - \mathscr{J}_N \Phi_N(f) \|_{\widetilde{\mathscr{F}}_N}	\leq K_N  \cdot \delta \cdot \|f\|_{\ell^2(G},\ \  \ \forall f \in \ell^2(G).
	\end{align}
	If additionally $\|\widetilde P_nJ_{n-1}f - J_n P_n f\|_{\ell^2(\widetilde G_n)} = 0$ or $ \| \rho_n(J_{n}f) - J_n \rho_n(f)\|_{\ell^2(\widetilde G_n)} = 0$ holds in each layer, then we have  $K_N = \sqrt{ (4^N-1) (2 D^2 +4B)/3\cdot B^{N-1}}$ if $B > 1/4$  and $K_N = \sqrt{  (2 D^2 +4B)\cdot(1 - B^N)/(1 - B)} $ if $B \leq 1/4$. If both additional equations hold, we have  $K_N = \sqrt{ (2^N-1) 2 D^2 \cdot B^{N-1}}$ if $B > 1/2$  and $K_N = \sqrt{  2 D^2 \cdot (1 - B^N)/(1 - B)} $ if $B \leq 1/2$. 
\end{Thm}

\begin{Not}
	Let us denote scattering propagators based on operators $\Delta_n$ and $P_n$ by $U_n$ and scattering propagators based on 	operators $\widetilde{\Delta}_n$ and $\widetilde P_n$ by $\widetilde{U}_n$. Similarly, to Notation \ref{ddd3} and , let us then write (with $q = (\gamma_N,...,\gamma_1)$)
	\begin{equation}
	\widetilde f_q := \widetilde{U}_n[\gamma_n]\circ...\circ\widetilde{U}_1[\gamma_1](J_0f).	
	\end{equation}
\end{Not}
\begin{proof}
	
	By definition we have
	\begin{align}\label{annsophie}
	\|\mathscr{J}\Phi_N(f) - \widetilde{\Phi}_N(J_0f) \|_{\widetilde{\mathscr{F}}_N}^2 =  \sum\limits_{n=1}^N \underbrace{\left(\sum\limits_{q \in \Gamma^{n-1}} \|J_n\chi_n(\Delta_n)\rho_n(P_n(f_q)) - \chi_n(\widetilde\Delta_n)\rho_n(P_n(\widetilde f_q)) \|^2_{\ell^2(\widetilde G_n)}\right)}_{=:a_n}.
	\end{align}
	We define $
	b_{n} := \sum_{q \in \Gamma^n} \|J_nf_q - \widetilde{f}_q\|^2_{\ell^2(\widetilde G_n)}$, with $b_0 = \|J_0f - J_0f\|^2_{\ell^2(\widetilde G)}= 0$ and note 
	\begin{align}
	a_n + b_n =&\sum\limits_{q \in \Gamma^{n-1}}\bigg(\|J_n\chi_n(\Delta_n)\rho_n(P_n(f_q)-\chi_n(\widetilde\Delta_n)\rho_n(P_n(\widetilde f_q) \|^2_{\ell^2(\widetilde G_n)} \\
	&+\left.   \sum\limits_{\gamma_n \in \Gamma_n}  \|J_ng_{\gamma_n}(\Delta_n)\rho_n(P_n((f_q))) - g_{\gamma_n}(\widetilde\Delta_n)\rho_n(P_n(\widetilde f_q)) \|^2_{\ell^2(\widetilde G_n)}     \right).
	\end{align}
	Using
	\begin{align}
	\frac12&\|J_ng_{\gamma_n}(\Delta_n)\rho_n(P_n((f_q))) - g_{\gamma_n}(\widetilde{\Delta}_n)\rho_n(P_n(\widetilde{f}_q)) \|^2_{\ell^2(\widetilde G_n)}\\ \leq& \|[J_ng_{\gamma_n}(\Delta_n) - g_{\gamma_n}(\widetilde{\Delta}_n)J_n]\rho_n(P_n(f_q)) \|^2_{\ell^2(\widetilde G_n)}\\
	+& \|g_{\gamma_n}(\widetilde\Delta_n)[J_n\rho_n(P_n((f_q))) - \rho_n(P_n(\widetilde{f}_q))] \|^2_{\ell^2(\widetilde G_n)}\\
	\leq&\|[J_ng_{\gamma_n}(\Delta_n) - g_{\gamma_n}(\widetilde{\Delta}_n)J_n]\|_{op}\cdot\|\rho_n(P_n(f_q)) \|^2_{\ell^2(\widetilde G_n)}\\
	+& \|g_{\gamma_n}(\widetilde\Delta_n)[J_n\rho_n(P_n((f_q))) - \rho_n(P_n(\widetilde{f}_q))] \|^2_{\ell^2(\widetilde G_n)},
	\end{align}
	and $\|[g_{\gamma_n}(\Delta_n) - g_{\gamma_n}(\widetilde{\Delta}_n)]\|_\infty \leq C^2_{g_\gamma}\cdot\delta^2$ (c.f. Lemma \ref{cpl}), we find
	\begin{align}
	a_n + b_n \leq & 2\sum_{q \in \Gamma^{n-1}} \left( C_{\chi_n}^2 + \sum\limits_{\gamma_n \in \Gamma_n} C_{g_{\gamma_n}^2}\right)(L^+_nR^+_n)^2\delta^2||\rho_n(P_n(\widetilde f_q))||_{\ell^2(\widetilde G_n)}^2\\
	+&2 \sum_{q \in \Gamma^{n-1}} B_n ||J_n\rho_n(P_n( f_q))-\rho_n(P_n(\widetilde f_q))||_{\ell^2(\widetilde G_n)}^2\\
	\leq&2\sum_{q \in \Gamma^{n-1}} \delta^2\left( C_{\chi_n}^2 + \sum\limits_{\gamma_n \in \Gamma_n} C_{g_{\gamma_n}^2}\right)(L^+_nR^+_n)^2||\rho_n(P_n(\widetilde f_q))||_{\ell^2(\widetilde G_n)}^2\\
	+&4B\cdot B^{n-1}||f||^2_{\ell^2(G)}\delta^2 + 8B\cdot B^{n-1}||f||^2_{\ell^2(G)}\delta^2 + 8Bb_{n-1},
	\end{align}
	where the second inequality arises from permuting the identification operator $J_n$ through non-linearity and connecting operator.		
	Using $C_{\chi_n}^2+\sum_{\gamma_n \in \Gamma_n} C_{\gamma_n}^2\leq D^2$, we then infer
	\begin{equation}
	a_n \leq (b_{n-1} -b_n ) + [8B-1]b_{n-1} +(2 D^2 +12 B) B^{n-1} \delta^2||f||^2_{\ell^2(G)}.
	\end{equation}
	If $B\leq\frac18$, summing over $n$ and using a geometric sum argument yields the desired stability constant.\\
	Hence let us assume $B > \frac18$. 	Using similar arguments as before, we find
	\begin{align}
	b_{n-1} \leq& (2 D^2 +12B) \delta^2 B^{n-2} ||f||^2_{\ell^2(G)} + 8B b_{n-2} \\
	\leq& \left(\sum\limits_{k=1}^{n-1}8^{k-1}\right)B^{n-2}(2 D^2 +12B)\delta^2 ||f||_{\ell^2(G)}^2=  \frac{1}{56}(8^n - 8) (2 D^2 +12)\delta^2 ||f||_{\ell^2(G)}^2.
	\end{align} 
	In total we find
	\begin{align}
	&\sum\limits_{n=1}^N a_n \\
	\leq &\underbrace{(b_0-b_N)}_{\leq 0}+(2 D^2 +12 B) B^{n-1} \delta^2||f||^2_{\ell^2(G)} +   (8B-1)(8^{n-1}-1)/7 B^{n-2}\cdot (2 D^2 +12 B)\delta^2||f||^2_{\ell^2(G)}  \\
 \leq & (8^N-1) (2 D^2 +12B)/7\cdot B^{N-1} ||f||_{\ell^2(G)}^2.
	\end{align}
	
	If one of the additional equations holds, we find
	\begin{equation}
	a_n + b_n \leq  (b_{n-1} -b_n ) + [4B - 1]b_{n-1} +(2 D^2 +4B ) \delta^2||f||^2_{\ell^2(G)}.
	\end{equation}
	and 
	\begin{align}
	b_{n-1} \leq& (2 D^2 + 4B) \delta^2 B^{n-2}||f||^2_{\ell^2(G)} + 4B b_{n-2} \\
	\leq& \left(\sum\limits_{k=1}^{n-1}4^{k-1}\right)B^{n-2}(2 D^2 +4)\delta^2 ||f||_{\ell^2(G)}^2=  \frac{1}{12}(4^n - 4) B^{n-2}(2 D^2 +4)\delta^2 ||f||_{\ell^2(G)}^2.
	\end{align} 
Arguing as previously yields the desired stability bounds.\\
If both additional equations are satisfied the proof is virtually the same as the one for Theorem \ref{sdimoppert}.
%
%
%
\end{proof}

\section{Details on Energy Decay and Truncation Stability}\label{somelabelpf}

We first prove the statement made about the relation between truncation stability and energy:

\begin{Lem}

Given the energy  $W_N := \sum_{(\gamma_N,...,\gamma_1) \in \Gamma^{N}}\|U[\gamma_{N}]\circ...\circ U[\gamma_1](f)\|^2_{\ell^2(G_{N})} $ stored in the network at layer $N$,  we have after extending $\Phi_N(f)$ by zero to match dimensions with $\Phi_{N+1}(f)$ that
\begin{equation}
\| \Phi_N(f) - \Phi_{N+1}(f) 	\|^2_{\mathscr{F}_{N + 1}} \leq \left(R^+_{N+1} L^+_{N+1} \right)^2B_{N+1} \cdot W_N.
\end{equation}

\end{Lem}

\begin{proof}
We note 
\begin{align}
&\| \Phi_N(f) - \Phi_{N+1}(f) 	\|^2_{\mathscr{F}_{N + 1}} = \sum\limits_{(\gamma_{N-1},...,\gamma_1)\in \Gamma^{N}}  \|V_{N + 1}\circ U[\gamma_{N}]\circ...\circ U[\gamma_1](f)\|_{\ell^2(G_{N+1})}^2\\
\leq& \left(R^+_{N+1} L^+_{N+1} \right)^2B_{N+1} \sum\limits_{(\gamma_{N-1},...,\gamma_1)\in \Gamma^{N-1}}  \| U[\gamma_{N}]\circ...\circ U[\gamma_1](f)\|_{\ell^2(G_{N+1})}^2.
\end{align}

	\end{proof}

In fact one can prove even more:

\begin{Lem}\label{somelemma}
	The energy $W_N$ stored in layer $N$ satisfies
	\begin{align}
	C_N^-\|f\|^2_{\ell^2(G)} \leq \| \Phi_{N}(f) \|_{\mathscr{F}_N}  + W_N(f) \leq C_N^+\|f\|^2_{\ell^2(G)},
	\end{align}
	with constants $C^-_N := \prod\limits_{i=1}^N \min\left\{ 1,A_i(L_i^-R_i^-)^2\right\}$ and $
	C^+_N := \prod\limits_{i=1}^N \max\left\{ 1,B_i(L_i^+R_i^+)^2\right\}$.

\end{Lem}

\begin{proof}
	\begin{align}
	&\min\left\{ 1,A_1(L_1^-R_1^-)^2\right\}	||f||^2_{\ell^2(G)} \\
	=& A_1(L_1^-R_1^-)^2 ||f||^2_{\ell^2(G)}\\
	=& A_1 ||\rho_1(P_1(f))||^2_{\ell^2(G_1)}\\
	\leq&   \sum\limits_{\gamma_1 \in \Gamma_1}||g_{\gamma_1}(\Delta_1)\rho_1(P_1(f))||^2_{\ell^2(G_1)} + ||\chi_1(\Delta_1) \rho_1(P_1(f))||^2_{\ell^2(G_1)} \\
	=& \sum\limits_{q \in \Gamma^1}||U[q](f)||^2_{\ell^2(G_1)} + ||\chi_1(\Delta_1) \rho_1(P_1(f))||^2_{\ell^2(G_1)}\\
	=& ||\chi_1(\Delta_1) \rho_1(P_1(f))||^2_{\ell^2(G_1)} + W_1(f),
	\end{align}
	and similarly
	\begin{align}
	&||\chi_1(\Delta_1) \rho_1(P_1(f))||^2_{\ell^2(G_1)} + W_1(f)\\
	=& \sum\limits_{q \in \Gamma^1}||U[q](f)||^2_{\ell^2(G_1)} + ||\chi_1(\Delta_1) \rho_1(P_1(f))||^2_{\ell^2(G_1)} \\
	\leq& B_1(L_1^+R^+_1)^2||f||^2_{\ell^2(G)}.
	\end{align}	
	
	This yields the starting point for our induction.
	Now for the inductive step assume the claim holds up until layer $N-1$. Then we have 
	\begin{equation}
	C_{N-1}^-||f||^2_{\ell^2(G)} \leq \sum\limits_{n=1}^{N-1} \left( \sum\limits_{q \in \Gamma^{n-1}}||\chi_n(\Delta_{n})f_q||_{\ell^2(G_{n})}^2\right) + W_{N-1}(f) \leq C_{N-1}^+||f||^2_{\ell^2(G)}.
	\end{equation}
using Notation \ref{cc2}. We note
	\begin{align}
	&\sum\limits_{n=1}^{N} \left( \sum\limits_{q \in \Gamma^{n-1}}||\chi_n(\Delta_{n})\rho_n(P_n(f_q))||_{\ell^2(G_{n})}^2\right) +W_N\\
	=& \sum\limits_{n=1}^{N-1} \left( \sum\limits_{q \in \Gamma^{n-1}}||\chi_n(\Delta_{n})\rho_n(P_n(f_q))||_{\ell^2(G_{n})}^2\right) + \sum\limits_{q \in \Gamma^{N-1}}||\chi_N(\Delta_{N})\rho_N(P_N(f_q))||^2_{\ell^2(G_{N})}\\
	+& \sum\limits_{q \in \Gamma^{N}}||f_q||^2_{\ell^2(G_N)}.
	\end{align}
	Every path $\widetilde q \in \Gamma^N $ may be written as $q=(\gamma_n,q)$, for some $\gamma_n \in \Gamma_n$ and $q \in \Gamma^{N-1}$. Thus we have
	\begin{equation}
	\sum\limits_{q \in \Gamma^{N}}||f_q||^2_{\ell^2(G_N)} = \sum\limits_{q \in \Gamma^{N-1}}\sum\limits_{\gamma_N \in \Gamma_N}||g_{\gamma_N}(\Delta_{N})P_N(\rho_N(f_q))||^2_{\ell^2(G_N)}
	\end{equation}
	Inserting this in the above equation yields
	\begin{align}
	&\sum\limits_{n=1}^{N} \left( \sum\limits_{q \in \Gamma^{n-1}}||\chi_n(\Delta_{n})\rho_n(P_n(f_q))||_{\ell^2(G_{n})}^2\right) +W_N\\
	=& \sum\limits_{n=1}^{N-1} \left( \sum\limits_{q \in \Gamma^{n-1}}||\chi_n(\Delta_{n})\rho_n(P_n(f_q))||_{\ell^2(G_{n})}^2\right)\\
	+& \sum\limits_{q \in \Gamma^{N-1}}\underbrace{\left(||\chi_{N}(\Delta_{N})\rho_N(P_N(f_q))||^2_{\ell^2(G_{n-1})} + \sum\limits_{\gamma_n \in \Gamma_N}||g_{\gamma_N}(\Delta_{N})P_N(\rho_N(f_q)||^2_{\ell^2(G_N)}\right)}_{=: \beta(f_q)}.
	\end{align}
	We have
	\begin{equation}
	( L^-_NR^-_N)^2A_N ||f_q||^2_{\ell^2(G_{n-1})} \leq \beta(f_q) \leq ( L^+_NR^+_N)^2B_N ||f_q||^2_{\ell^2(G_{n-1})},
	\end{equation}
	by the operator frame property. 
	With this we find:
	\begin{align}
	&\min\{1,(L^-_NR^-_N)^2A_N\}\left(\sum\limits_{n=1}^{N-1} \left( \sum\limits_{q \in \Gamma^{n-1}}||\chi_n(\Delta_{n})\rho_n(P_n(f_q))||_{\ell^2(G_{n})}^2\right) +W_{N-1}\right)\\
	\leq&\sum\limits_{n=1}^{N}\sum\limits_{q \in \Gamma^{n-1}}||\chi_n(\Delta_{n})\rho_n(P_n(f_q))||_{\ell^2(G_{n})}^2 + W_{N}\\
	\leq&\max\{1,( L^-_NR^-_N)^2B_N\} \left(\sum\limits_{n=1}^{N-1} \left( \sum\limits_{q \in \Gamma^n}||\chi_n(\Delta_{n})U[q](f)||_{\ell^2(G_{n})}^2\right)+W_{N-1}\right),
	\end{align}
	after unravelling the definition
	\begin{equation}
	W_{N-1}(f) \equiv \sum\limits_{q\in\Gamma^{N}}||f_q||^2_{\ell^2(G_{n-1})} .
	\end{equation}
	The induction hypothesis together with the definition of $C^\pm_N$ now yields the claim.	
	
\end{proof}
With this we now prove our main theorem concerning energy decay.
\begin{Thm} Let $\Phi_\infty$ be a generalized graph scattering transform based on a  an operator sequence $\mathscr{D}_\infty = (P_n,\Delta_n)_{n=1}^\infty$ and a module sequence $\Omega_\infty$ with each $\rho_n(\cdot)\geq 0$. Assume in each layer $n \geq 1$ that there is an eigenvector $\psi_n$ of $\Delta_n$ with solely positive entries; denote the smallest entry by $m_n:=\min_{i \in G_n}\psi_n[i] $. Denote the eigenvalue corresponding to $\psi_n$ by $\lambda_n$. Quantify the 'spectral-gap' opened up at this eigenvalue through neglecting the output-generating function by $\eta_n := \sum_{\gamma_n \in \Gamma_n} |g_{\gamma_n}(\lambda_n)|^2  $ and assume $B_nm_n \geq \eta_n$. We then have
	\begin{align}
	W_N(f) \leq C_N^+\cdot \left[\prod\limits_{n=1}^{N}\left(1-\left(m^2_n-\frac{\eta_n}{B_n}\right)\right)\right]\cdot\|f\|^2_{\ell^2(G)}.
	\end{align}
\end{Thm}
\begin{proof}
	
	Denote the spectral projection (i.e. the orthogonal projection projecting to the space of eigenvectors) onto the eigenspace corresponding to $\lambda_n$ by $P_c^{n}$. 
	
	%
	%
	%
	Then we have
	\begin{align}
	W_N(f) &=  \sum\limits_{q \in \Gamma^{N-1}}\sum\limits_{\gamma_N \in \Gamma_N} ||g_{\gamma_N}(\Delta_{N})\rho_N(P_N(f_q))||^2_{\ell^2(G_{N})}\\
	&=\sum\limits_{q \in \Gamma^{N-1}}\sum\limits_{\gamma_N \in \Gamma_N} ||g_{\gamma_N}(\Delta_{N})(1-P_c^{N})\rho_N(P_N(f_q))||^2_{\ell^2(G_{N})} \\&+\sum\limits_{q \in \Gamma^{N-1}}\sum\limits_{\gamma_N \in \Gamma_N} ||g_{\gamma_N}(\Delta_{N})P_c^{N}\rho_N(P_N(f_q))||^2_{\ell^2(G_{N})} \\
	&\leq \sum\limits_{q \in \Gamma^{N-1}} B_N||(1-P_c^{N})\rho_N(P_N(f_q))||^2_{\ell^2(G_{N})} \\
	&+ \sum\limits_{q \in \Gamma^{N-1}}\eta_N||P_c^{N}\rho_N(P_N(f_q))||^2_{\ell^2(G_{N})}\\
	&\leq\sum\limits_{q \in \Gamma^{N-1}} B_N||(1-P_c^{N})\rho_N(P_N(f_q))||^2_{\ell^2(G_{N})} \\
	&+ \sum\limits_{q \in \Gamma^{N-1}}\eta_N||\rho_N(P_N(f_q))||^2_{\ell^2(G_{N})}.
	\end{align}
	%
	By orthogonality of the spectral projection, we then have
	\begin{equation}
	||(1-P_c^{N})\rho_N(P_n(f_q))||^2_{\ell^2(G_{N})} = 	||\rho_N(P_N(f_q))||^2_{\ell^2(G_{N})} - 	||P_c^{N}\rho_N(P_n(f_q))||^2_{\ell^2(G_{N})}.
	\end{equation}
	Furthermore, we have 
	\begin{equation}
	|\langle\psi^{N},\rho_N(P_n(f_q))\rangle_{\ell^2(G_{N})}|^2\leq  ||P_c^{N}\rho_N(P_n(f_q))||^2_{\ell^2(G_{N})}
	\end{equation}
	with equality if the multiplicity of $\lambda^{N}$ is exactly one.  With this we find
	\begin{align}
	||(1-P_c^{N})\rho_N(P_N(f_q))||^2_{\ell^2(G_{N})} &= ||\rho_N(P_N(f_q))||^2_{\ell^2(G_{N})} - 	||P_c^{N}\rho_N(P_N(f_q))||^2_{\ell^2(G_{N})}\\
	& \leq ||\rho_N(P_N(f_q))||^2_{\ell^2(G_{N})} - |\langle\psi^{N},\rho_N(P_N(f_q))\rangle_{\ell^2(G_{N})}|^2.	\\
	\end{align}
	Since the image of $\rho_N$ is contained in $\mathds{R}_+$ by assumption, we have
	\begin{align}
	|\langle\psi^{N},\rho_N(P_N(f_q))\rangle_{\ell^2(G_{N})}|^2 &=  \left|\sum\limits_{i=1 }^{|G_N|} \rho_N(P_N(f_q))_i(\psi_N)_i \mu_i \right|^2\\
	&\geq \left|  \sum\limits_{i=1 }^{|G_N|}|\rho_N(P_N(f_q))_i| \mu_i\right|^2 \cdot m_N^2\\
	&\geq\left|  \sum\limits_{i=1 }^{|G_N|}|\rho_N(P_N(f_q))_i|^2 \mu_i^2\right| \cdot m_N^2\\
	&\geq\left|  \sum\limits_{i=1 }^{|G_N|}|\rho_N(P_N(f_q))_i|^2 \mu_i\right| \cdot m_N^2\\
	&\geq  ||\rho_N(P_N(f_q))||_{\ell^2(G_{N})}^2 \cdot m_N^2
	\end{align}
	Here the second to last inequality follows since in any finite dimensional vector space, the $1$-norm is larger than the $2$-norm ($||f||_1 \geq||f||_2  $) and all weights are assumed to satisfy $\mu_i \geq 1$. Thus we now know
	\begin{equation}
	||(1-P_c^{N})\rho_N(P_N(f_q))||^2_{\ell^2(G_{N})} \leq \left(1-m_N^2\right)||\rho_N(P_N(f_q))||^2_{\ell^2(G_{N})}.
	\end{equation}
	Inserting this in our estimate for $W_N(f)$  we find 
	
	\begin{align}
	W_N(f) &\leq \left(1-\left(m_N^2-\frac{\eta_n}{B_n}\right)\right) L^+_NR^+_NB_N \cdot W_{N-1}(f)\\
	&\leq C_N^+\ \prod\limits_{n=1}^{N}\left(1-\left(m_N^2-\frac{\eta_n}{B_n}\right)\right)||f||^2_{\ell^2(G)}.
	\end{align}
	
\end{proof}

Taking $N$ to infinity, we know that $C^+_N$ converges to something larger than zero by assumption.\\ For products of the form $\prod\limits_{n=0}^N(1 - q_n)$ with $0\leq q_n < 1$ it is a standard exercise to prove that the limit is non-zero precisely if the sum over the $q_n$ converges.
Combining the above result with Lemma \ref{somelemma}, we obtain as an immediate Corollary:

\begin{Cor}\label{trivcor}
	\normalfont 	In the setting 	of Theorem \ref{EDTM}, the generalized scattering transform satisfies  $\Phi_\infty^{-1}(0) = \{0\}$  if  $C^\pm_N\rightarrow C^\pm$ for some positive constants $C^\pm$  and $\sum_{n=1}^N (m_n-\eta_n/B_n)\rightarrow \infty$ as $N \rightarrow \infty$.
\end{Cor}

\section{Stability of Graph Level Feature Aggregation}\label{Aggregation}
\subsection{General non-linear feature aggregation:}\label{Awildsectionappears}
Our main stability theorem for non-linear feature aggregation is as follows:
\begin{Thm}
	We have 
	\begin{equation}
	\|\Psi_N(f) - \Psi_N(g)\|_{\mathscr{R}_N}  \leq \left(1+ \sum\limits_{n=1}^N\max\{[B_n-1],[B_n(L_n^+R_n^+)^2-1],0\}\prod\limits_{k=1}^{n-1}B_k\right)^\frac12\|f-h\|_{\ell^2(G)}.
	\end{equation} 
	With the conditions and notation of Theorem \ref{sdimoppert} we have 
	\begin{equation}
	\|\Psi_N(f)  -\widetilde{\Psi}_N(f) \|_{\mathscr{R}_N} \leq  \sqrt{2(2^{N}-1)  }\cdot\sqrt{(\max\{B,1/2\})^{N - 1}} \cdot D \cdot \delta \cdot \|f\|_{\ell^2(G)}.
	\end{equation}
	Additionally,  in the setting of Theorem \ref{BigT}, assuming that for each $n \leq N$ the identification operator $J_n$ satisfies $
	\big| \|J_nf\|_{\ell^1(\widetilde G_n)}/\sqrt{\mu_{\widetilde G_n}} - \|f\|_{\ell^1( G_n)}/\sqrt{\mu_{ G_n}}\big| ,\big| \|J_nf\|_{\ell^k(\widetilde G_n)} - \|f\|_{\ell^k( G_n)}\big| \leq \delta \cdot K \cdot \|f\|_{\ell^2( G_n)} $  ($2 \leq k \leq p_n
	$) 
	implies ($\forall f\in \ell^2(G)$)
	\begin{align}
	\|\widetilde{\Psi}_N(J_0f) -  \Psi_N(f) \|_{\mathscr{R}_N}	\leq \sqrt{2}\cdot\sqrt{K^2_N \cdot + K^2} \cdot \delta \cdot \|f\|_{\ell^2(G)}.
	\end{align}
	Furhermore, under the assumptions of Corollary \ref{trivcor} $\Psi_\infty(f)=0$ implies $f=0$.
\end{Thm}
\begin{proof}
	Let $f,h \in \ell^2(G)$. To prove the first two claims, it suffices to prove 
	\begin{align}
	\|\Psi_N(f) - \Psi_N(h)\|_{\mathscr{R}_N} \leq \|\Phi_N(f) - \Phi_N(h)\|_{\mathscr{F}_N},
	\end{align}	
	and 
	\begin{equation}
	\|\Psi_N(f)  -\widetilde{\Psi}_N(f) \|_{\mathscr{R}_N}  \leq \|\Phi_N(f)  -\widetilde{\Phi}_N(f) \|_{\mathscr{F}_N} .
	\end{equation}	
	Both statements follow immediately, as soon as we have proved
	\begin{equation}
	\|N_p^G(f) - N_p^G(h)\|_{\mathds{R}^p } \leq \| f - h\|_{\ell^2(G)}
	\end{equation}
	for arbitrary choices of $p$ and $G$. To this end we note that for $p\geq2$ we have $\|f\|_{\ell^p(G)} \leq \|f\|_{\ell^2(G)}$ by the monotonicity of $p$-norms, while we have $\|f\|_{\ell^1(G)} \leq \sqrt{\mu_G}\cdot\|f\|_{\ell^2(G)}$ by Hölder's inequality. With this we find
	\begin{align}
	\|N_p^G(f) - N_p^G(h)\|_{\mathds{R}^p }^2 &= \frac1p\left(\frac{1}{\mu_G}|\|f\|_{\ell^1(G)}-\|h\|_{\ell^1(G)}|^2 + \sum_{i=2}^{p} |\|f\|_{\ell^i(G)}-\|h\|_{\ell^i(G)}|^2\right) \\
	&\leq\frac1p\left(\frac{1}{\mu_G}|\|f-h\|_{\ell^1(G)}|^2 + \sum_{i=2}^{p} |\|f-h\|_{\ell^i(G)}|^2\right) \\
	&\leq \frac1p\cdot p \cdot |\|f-h\|_{\ell^2(G)}|^2 \\
	&=  \| f - h\|_{\ell^2(G)}.
	\end{align}
	where we have employed the reverse triangle inequality in the first step.
	
	To prove the second claim, we note that we have 
	\begin{align}
	&\|\Psi_N(f) - \widetilde{\Psi}_N(J_0f) \|_{\mathscr{R}_N}^2\\
	=&  \sum\limits_{n=1}^N \left(\sum\limits_{q \in \Gamma^{n-1}} \|N^{G_n}_{p_n}(\underbrace{\chi_n(\Delta_n)\rho_n(P_n((f_q)))}_{=:x_q}) - N^{\widetilde G_n}_{p_n}(\underbrace{\chi_n(\widetilde\Delta_n)\rho_n(P_n(\widetilde f_q))}_{=:\widetilde x_q})\|^2_{\mathds{R}^{p_n}}\right) \\
	\leq& 2\sum\limits_{n=1}^N \left(\sum\limits_{q \in \Gamma^{n-1}}\|N^{\widetilde{G}_n}_{p_n}(J_nx_q) - N^{\widetilde G_n}_{p_n}(\widetilde x_q)\|_{\mathds{R}^{p_n}} \right)\\
	+&2\sum\limits_{n=1}^N \left(\sum\limits_{q \in \Gamma^{n-1}}\|N^{\widetilde{G}_n}_{p_n}(J_nx_q) - N^{ G_n}_{p_n}( x_q)\|_{\mathds{R}^{p_n}} \right)\\
	=& 2	\|\mathscr{J}\Phi_N(f) - \widetilde{\Phi}_N(J_0f) \|_{\mathscr{F}_N}^2\\
	+&2\sum\limits_{n=1}^N \left(\sum\limits_{q \in \Gamma^{n-1}}\|N^{\widetilde{G}_n}_{p_n}(J_nx_q) - N^{ G_n}_{p_n}( x_q)\|_{\mathds{R}^{p_n}} \right).\\
	\end{align}
	Thus it remains to bound the last expression. We have 
	\begin{align}
	&\|N^{\widetilde{G}_n}_{p}(J_nx_q) - N^{ G_n}_{p_n}( x_q)\|_{\mathds{R}^{p_n}} \\
	=&\frac{1}{p_n}\left(\left|\frac{1}{\sqrt{\mu_{G_n}}}\|f\|_{\ell^1(G)}-\frac{1}{\sqrt{\mu_{\widetilde G_n}}}\|J_nf\|_{\ell^1(\widetilde G)}\right|^2 + \sum_{i=2}^{p_n} |\|f\|_{\ell^i(G)}-\|J_nf\|_{\ell^i(\widetilde G)}|^2\right) \\
	\leq &K^2 \cdot \delta^2 \cdot \|x_q\|^2_{\ell^2(G_n)}.
	\end{align}
	By our results of Appendix \ref{sptcy} and since we assume admissibility, we have
	\begin{align}
	\sum\limits_{n=1}^N \sum_{q \in \Gamma^{n-1}} \|x_q\|^2_{\ell^2(G_n)} \leq \|f\|_{\ell^2(G)}^2.
	\end{align}
	Thus in total 
	\begin{align}
	&\|\Psi_N(f) - \widetilde{\Psi}_N(J_0f) \|_{\mathscr{F}_N}^2 \leq  	2 \|\mathscr{J}\Phi_N(f) - \widetilde{\Phi}_N(J_0f) \|_{\mathscr{F}_N}^2 + 2 K \delta \|f\|_{\ell^2(G)},
	\end{align}
	from which our stability claim follows.
	
	It remains to prove that the assumptions of Corollary \ref{trivcor} $\Psi_\infty(f)=0$ imply $f=0$. But since $N^G_p(f)=0$ implies $f=0$, this is clear.
\end{proof}

\subsection{Low-Pass feature Aggregation}\label{J2}
The main assumption we have in this section is that each operator $\Delta_n$ (and $ \widetilde \Delta_n$) has a simple lowest lying eigenvalue equal to zero. We denote the associated eigenvector (determined up to a complex phase) by $\psi_{\Delta_n}$ and the associated spectral projection to the lowest lying eigenvalue by $P_{\Delta_n}$. It acts as 
\begin{equation}
P_{\Delta_n}f \equiv \psi_{\Delta_n}\langle\psi_{\Delta_n},f\rangle_{\ell^2(G_n)}.
\end{equation}

Now we are ready to state our main stability result under these circumstances:
\begin{Thm}\label{scalarstabII}
We have
	\begin{equation}
	\|\Psi^{|\langle\cdot,\cdot\rangle|}_N(f) - \Psi^{|\langle\cdot,\cdot\rangle|}_N(g)\|_{\mathscr{C}_N}  \leq \left(1+ \sum\limits_{n=1}^N\max\{[B_n-1],[B_n(L_n^+R_n^+)^2-1],0\}\prod\limits_{k=1}^{n-1}B_k\right)^\frac12\|f-h\|_{\ell^2(G)}.
	\end{equation} 
	With the conditions and notation of Theorem \ref{sdimoppert} and under the additional assumption $\|(P_{\Delta_n}-P_{\widetilde \Delta_n})\|_{op} \leq K\cdot\delta$ for $n \leq N$ and some $K\geq0$, we have 
	\begin{equation}
	\|\Psi^{|\langle\cdot,\cdot\rangle|}_N(f)  -\widetilde{\Psi}^{|\langle\cdot,\cdot\rangle|}_N(f) \|_{\mathscr{C}_N} \leq  \sqrt{2} \cdot   \sqrt{2(2^{N}-1)(\max\{B,1/2\})^{N - 1} +K^2 } \cdot \delta \cdot \|f\|_{\ell^2(G)}.
	\end{equation}
	In the setting of Theorem \ref{BigT} and under the additional assumption $|\|P_{\Delta_n}f\|_{\ell^2(G_n)}  - \|P_{\widetilde \Delta_n}J_n f\|_{\ell^2(\widetilde G_n)}  |\leq K\delta ||f||_{\ell^2( G_n)} $ for all $f \in \ell^2(G_n)$ ($n\leq N$), we have
	\begin{align}
	\|\widetilde{\Psi}^{|\langle\cdot,\cdot\rangle|}_N(J_0f) -  \Psi^{|\langle\cdot,\cdot\rangle|}_N(f) \|_{\mathscr{C}_N}	\leq \sqrt{2}\cdot\sqrt{K^2_N \cdot + K^2} \cdot \delta \cdot \|f\|_{\ell^2(G)}.
	\end{align}
	
\end{Thm}

\begin{proof}
	Let $f,h \in \ell^2(G)$. To prove the first claim, it suffices to prove 
	\begin{align}
	\|\Psi^{|\langle\cdot,\cdot\rangle|}_N(f) - \Psi^{|\langle\cdot,\cdot\rangle|}_N(h)\|_{\mathscr{C}_N} \leq \|\Phi_N(f) - \Phi_N(h)\|_{\mathscr{F}_N}.
	\end{align}	
	This immediately follows from the fact that for all $f \in \ell^2(G_n)$
	\begin{equation}
	|\langle\psi_{\Delta_n},f\rangle_{\ell^2(G_n)}|^2\leq \|\psi_{\Delta_n}\|^2_{\ell^2(G_n)}\cdot\|f\|^2_{\ell^2(G_n)}
	\end{equation}
	by Hölder's inequality.
	
	The next claim we want to prove is that we have for all $f \in \ell^2(G)$
	\begin{equation}
	\|\Psi^{|\langle\cdot,\cdot\rangle|}_N(f)  -\widetilde{\Psi}^{|\langle\cdot,\cdot\rangle|}_N(f) \|_{\mathscr{C}_N} \leq  \sqrt{2} \cdot   \sqrt{2(2^N-1) +K^2 } \cdot \delta \cdot \|f\|_{\ell^2(G)}.
	\end{equation}

	We note
	
	\begin{align}
	&\|\Psi^{|\langle\cdot,\cdot\rangle|}_N(f)  -\widetilde{\Psi}^{|\langle\cdot,\cdot\rangle|}_N(f) \|^2_{\mathscr{C}_N} \\
	=&  \sum\limits_{n=1}^N \left(\sum\limits_{q \in \Gamma^{n-1}} \left||\langle\psi_{\Delta_n},\underbrace{\chi_n(\Delta_n)\rho_n(P_n(f_q))}_{=:x_q}\rangle_{\ell^2(G_n)}| -|\langle\psi_{\widetilde \Delta_n},\underbrace{\chi_n(\widetilde\Delta_n)\rho_n(P_n(\widetilde f_q))}_{\widetilde x_q}\rangle_{\ell^2( G_n)}|\right|^2\right) \\
	=&\sum\limits_{n=1}^N \left(\sum\limits_{q \in \Gamma^{n-1}} \left|\|P_{\Delta_n}x_q\|_{\ell^2(G_n)} -\|P_{\widetilde \Delta_n}\widetilde x_q\|_{\ell^2( G_n)}\right|^2\right) \\
	\leq & \sum\limits_{n=1}^N \left(\sum\limits_{q \in \Gamma^{n-1}} \|P_{\Delta_n}x_q - P_{\widetilde \Delta_n}\widetilde x_q\|_{\ell^2(G_n)}^2\right) \\
	\leq & 2 \sum\limits_{n=1}^N \left(\sum\limits_{q \in \Gamma^{n-1}} \|P_{\widetilde \Delta_n}(x_q - \widetilde x_q)\|_{\ell^2(G_n)}^2\right) + 2 \sum\limits_{n=1}^N \left(\sum\limits_{q \in \Gamma^{n-1}} \|(P_{\Delta_n}-P_{\widetilde \Delta_n})x_q \|_{\ell^2(G_n)}^2\right)  \\
	\leq & 2 	\|\Phi_N(f) - \Phi_N(h)\|_{\mathscr{F}_N}^2  + 2 \sum\limits_{n=1}^N \left(\sum\limits_{q \in \Gamma^{n-1}} \|(P_{\Delta_n}-P_{\widetilde \Delta_n})x_q \|_{\ell^2(G_n)}^2\right)  \\
	\end{align}
	
	Hence we need to bound  the expression "$\|(P_{\Delta_n}-P_{\widetilde \Delta_n})x_q \|_{\ell^2(G_n)}^2$". We  note
	\begin{align}
	\|(P_{\Delta_n}-P_{\widetilde \Delta_n})x_q \|_{\ell^2(G_n)}^2 &\leq \|(P_{\Delta_n}-P_{\widetilde \Delta_n})\|_{op}\cdot\|x_q \|_{\ell^2(G_n)}^2\\
	& \leq K^2\cdot \delta^2\cdot\|x_q \|_{\ell^2(G_n)}^2 
	\end{align}
	and thus 
	\begin{align}
	&\|\Psi^{|\langle\cdot,\cdot\rangle|}_N(f)  -\widetilde{\Psi}^{|\langle\cdot,\cdot\rangle|}_N(f) \|_{\mathscr{C}_N}^2 \\
	\leq & 2 	\|\Phi_N(f) - \Phi_N(h)\|_{\mathscr{F}_N}^2  + 2 K^2\cdot\delta^2\cdot\sum\limits_{n=1}^N \left(\sum\limits_{q \in \Gamma^{n-1}} \|\chi_n(\Delta_n)\rho_n(P_n((f_q)))\|_{\ell^2(G_n)}^2\right)  \\
	\leq & 2 	\|\Phi_N(f) - \Phi_N(h)\|_{\mathscr{F}_N}^2  + 2 K^2\cdot\delta^2\cdot  \|f\|_{\ell^2(G)}^2  
	\end{align}
	and the claim follows.

	Finally we want to prove
	\begin{align}
	\|\widetilde{\Psi}^{|\langle\cdot,\cdot\rangle|}_N(J_0f) -  \Psi^{|\langle\cdot,\cdot\rangle|}_N(f) \|_{\mathscr{C}_N}	\leq \sqrt{2}\cdot\sqrt{K^2_N \cdot + K^2} \cdot \delta \cdot \|f\|_{\ell^2(G)}.
	\end{align}

	We note
	
	\begin{align}
	&\|\Psi^{|\langle\cdot,\cdot\rangle|^2}_N(f)  -\widetilde{\Psi}^{|\langle\cdot,\cdot\rangle|}_N(f) \|_{\mathscr{C}_N} \\
	=&  \sum\limits_{n=1}^N \left(\sum\limits_{q \in \Gamma^{n-1}} \left| |\langle\psi_{\Delta_n},\underbrace{\chi_n(\Delta_n)\rho_n(P_n((f_q)))}_{=:x_q}\rangle_{\ell^2(G_n)}| -|\langle\psi_{\widetilde \Delta_n},\underbrace{\chi_n(\widetilde\Delta_n)\rho_n(P_n(\widetilde f_q))}_{\widetilde x_q}\rangle_{\ell^2(\widetilde G_n)}|\right|^2\right) \\
	=&\sum\limits_{n=1}^N \left(\sum\limits_{q \in \Gamma^{n-1}} \left|\|P_{\Delta_n}x_q\|_{\ell^2(G_n)} -\|P_{\widetilde \Delta_n}\widetilde x_q\|_{\ell^2(\widetilde  G_n)}\right|^2\right) \\
	\leq &\sum\limits_{n=1}^N \left(\sum\limits_{q \in \Gamma^{n-1}} \left|\|P_{\Delta_n}x_q\|_{\ell^2(G_n)}  - \|P_{\widetilde \Delta_n}J_n x_q\|_{\ell^2(\widetilde G_n)}         +    \|P_{\widetilde \Delta_n}J_n x_q\|_{\ell^2(\widetilde G_n)}     
	-\|P_{\widetilde \Delta_n}\widetilde x_q\|_{\ell^2(\widetilde  G_n)}\right|^2\right) \\
	\leq &  2 \sum\limits_{n=1}^N \left(\sum\limits_{q \in \Gamma^{n-1}} \left|      \|P_{\widetilde \Delta_n}J_n x_q\|_{\ell^2(\widetilde G_n)}     
	-\|P_{\widetilde \Delta_n}\widetilde x_q\|_{\ell^2(\widetilde  G_n)}\right|^2\right)  \\
	+&2 \sum\limits_{n=1}^N \left(\sum\limits_{q \in \Gamma^{n-1}} \left|\|P_{\Delta_n}x_q\|_{\ell^2(G_n)}  - \|P_{\widetilde \Delta_n}J_n x_q\|_{\ell^2(\widetilde G_n)}     \right|^2\right)\\
	\leq &  2 \sum\limits_{n=1}^N \left(\sum\limits_{q \in \Gamma^{n-1}} \left|      \|P_{\widetilde \Delta_n}J_n x_q\|_{\ell^2(\widetilde G_n)}     
	-\|P_{\widetilde \Delta_n}\widetilde x_q\|_{\ell^2(\widetilde  G_n)}\right|^2\right)  \\
	+&2 \sum\limits_{n=1}^N \left(\sum\limits_{q \in \Gamma^{n-1}} \left|\|P_{\Delta_n}x_q\|_{\ell^2(G_n)}  - \|P_{\widetilde \Delta_n}J_n x_q\|_{\ell^2(\widetilde G_n)}     \right|^2\right)\\
	\leq & 2 	\|\mathscr{J}\Phi_N(f) - \widetilde\Phi_N(J_0 f)\|_{\widetilde{\mathscr{F}}_N}^2  + 2 \sum\limits_{n=1}^N \left(\sum\limits_{q \in \Gamma^{n-1}} \left|\|P_{\Delta_n}x_q\|_{\ell^2(G_n)}  - \|P_{\widetilde \Delta_n}J_n x_q\|_{\ell^2(\widetilde G_n)}     \right|^2\right) \\
	\leq & 2 	\|\mathscr{J}\Phi_N(f) - \widetilde\Phi_N(J_0 f)\|_{\widetilde{\mathscr{F}}_N}^2 
	+ 2 \sum\limits_{n=1}^N \left(\sum\limits_{q \in \Gamma^{n-1}} K^2\cdot\delta^2\|x_q\|^2_{\ell^2(G_n)} \right) \\
	\leq&  2 	\|\mathscr{J}\Phi_N(f) - \widetilde\Phi_N(J_0 f)\|_{\widetilde{\mathscr{F}}_N}^2 
	+  2  K^2\cdot\delta^2\|f\|^2_{\ell^2(G)}.  \\
	\end{align}
	which proves the claim. 
\end{proof}

In establishing triviality of the 'kernel', we have to be a tiny bit more careful: 

\begin{Thm} In the setting of of Corollary \ref{trivcor}, assume  that in each layer $n$, the output generating function $\chi_n$ of the underlying scattering transform satisfies $\chi_n(0)\neq0$ and $\chi_n(\lambda_i)=0$ for ordered non-zero eigenvalues $\lambda_2\leq...\leq\lambda_{|G_n|}$ of the operator $\Delta_n$.
	Then $\Psi^{|\langle\cdot,\cdot\rangle|}_\infty(f) = 0$ implies 	$f=0$.
\end{Thm}
\begin{proof}
	Under these assumptions, we do not lose any information by projecting to $\psi_{\Delta_n}$ in each $\ell^2(G_n)$, since the image of $\chi_{n}(\Delta_n)$ is already contained in the one-dimensional space generated by the lowest lying eigenvector 	$\psi_{\Delta_n}$.	
\end{proof}

\section{Details on Higher Order Scattering}\label{HLSTB}
Node signals capture information about nodes in isolation. However, one might also want to analyse or incorporate information about binary, ternary or even higher order relations between nodes, such as distances or angles between nodes representing atoms in a molecule. This can be formalized by considering tensorial input signals: 
\paragraph{Tensorial input:}
A $2$-tensor on a graph $G$, as it was already utilized in Section \ref{HOS}, is simply an element of $\mathds{C}^{|G|\times|G|}$ or -- equivalently -- a map from $G \times G$ to $\mathds C$, since it associates a complex number to each element $(g_1,g_2) \in G \times G$. Since $G \times G$ is precisely the set of (possible) edges $E$, we can equivalently think of $2$-tensors edge-signals. 
A $3$-tensor an element of $\mathds{C}^{|G|\times|G|\times|G|}$ or equivalently a map from $G\times G\times G \equiv G^3$ to $\mathds C$. A  $4$-tensor then is a map from $G^4 \equiv G\times G\times G\times G$ to $\mathds C$ or equivalenlty an element of $\mathds{C}^{|G|\times|G| \times|G| \times|G|}$ and so forth.
Clearly the space of $k$-tensors forms a linear vector space. Addition and scalar multiplication by $\lambda \in \mathds C$ are given by 
\begin{equation}
(f + \lambda g)_{i_1,...,i_k} := f_{i_1,...,i_k} + \lambda g_{i_1,...,i_k}
\end{equation}
with $f$ and $g$ being $k$-tensors.  For fixed $k$, we equip the space of $k$-tensors with an inner product according to 
\begin{equation}
\langle f,g \rangle = \sum^{|G|}_{i_1,...,i_k=1}\overline{f_{i_1,...,i_k}}g_{i_1,...,i_k}
 \mu_{i_1,...,i_k} 
 \end{equation}
  and  denote the resulting inner-product space by $\ell^2(G^k)$.


\paragraph{Operators on Spaces of Tensors:}
Since for fixed $k$ the space  $\ell^2(G^k)$ is simply a $|G|^k$-dimensional complex inner product space, there are exist normal operators $\Delta^k: \ell^2(G^k) \rightarrow \ell^2(G^k) $ on this space. Note that the $k$ in $\Delta^k$ signifies on which space this operator acts. It does not signify that an operator is raised to the $k^{th}$ power.
Setting for example node-weights $\mu_i$ and edge weights $\mu_{ik}$ to one, the adjacency matrix $W$ as well as normalized or un-normalized graph Laplacians constitute self-adjoint operators on $\ell^2(G^2)$, where they act by matrix multiplication.

\paragraph{Higher order Scattering Transforms:}
We can then follow the recipe laid out Section \ref{GGST} in constructing $k^{\textit{th}}$-order scattering transforms; all that we need are a module sequence $\Omega_N$ and an operator sequence $\mathscr{D}^k_N:=(P^k_n,\Delta^k_n)_{n=1}^N$, where now $P^k_n:\ell^2(G^k_{n-1})\rightarrow \ell^2(G^k_{n})$ and $\Delta^k_n:\ell^2(G^k_{n}) \rightarrow \ell^2(G^k_{n})$.\\
\ \\
\begin{minipage}{0.5\textwidth}
	
	\includegraphics[scale=0.34]{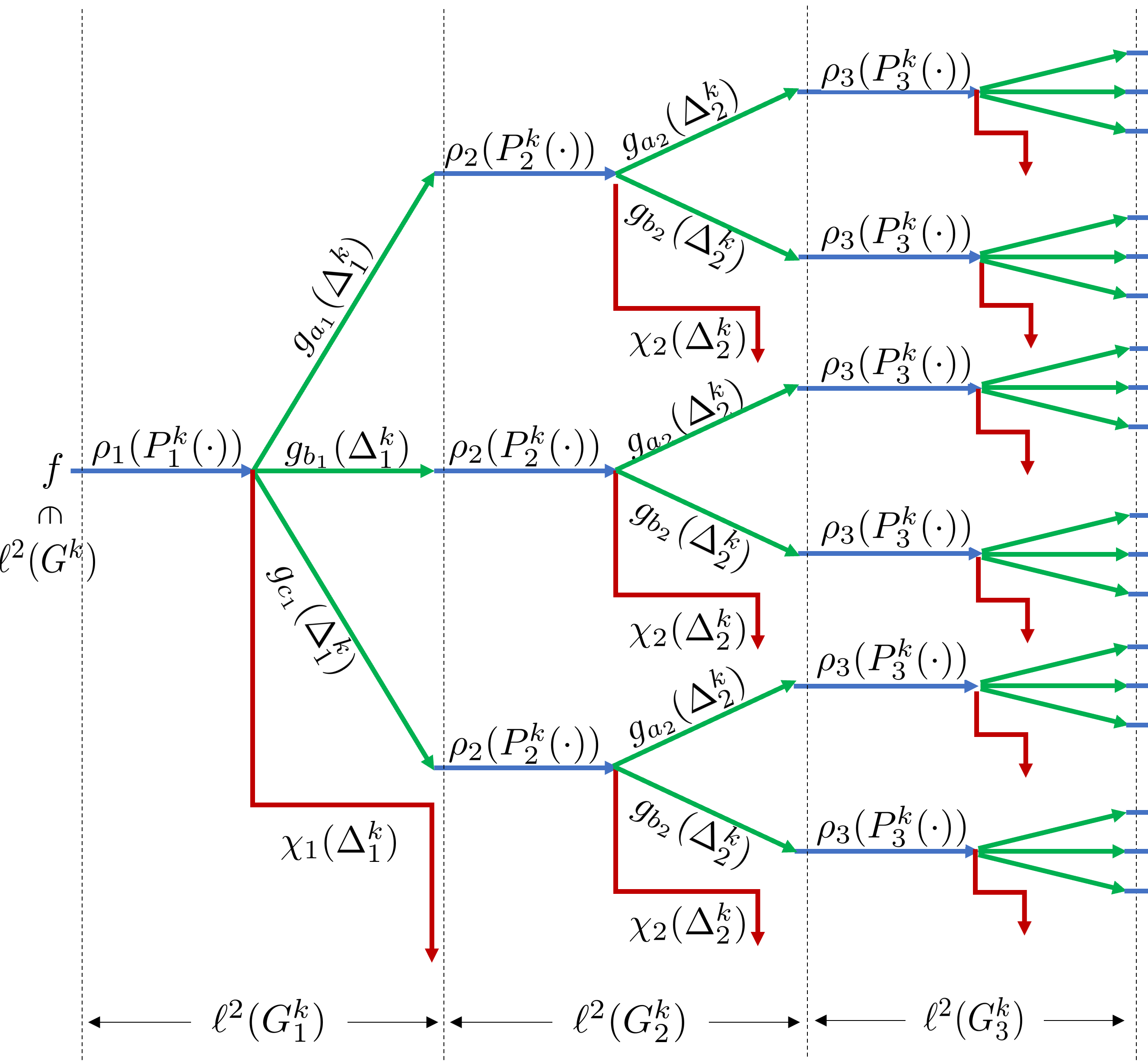}
	\captionof{figure}{Schematic Higher Order Scattering Architecture} 
	\label{SKAII}
\end{minipage}\hfill
\begin{minipage}{0.48\textwidth}
	
	To our initial signal $f \in \ell^2(G^k)$ we first apply the connecting operator $P_1^k$, yielding a signal representation in $\ell^2(G_1^k)$.  Subsequently, we apply the pointwise non-linearity $\rho_1$. Then we apply our graph filters $\{\chi_1(\Delta_1^k)\} \bigcup \{g_{\gamma_1}(\Delta_1^k)\}_{\gamma_1 \in  \Gamma_1}$ to $\rho_1(P^k_1(f))$ yielding the 
	output $V_1(f) := \chi_1(\Delta_1^k)\rho_1(P^k_1(f))$ as well as the intermediate hidden representations $\{U_1[\gamma_1](f) := g_{\gamma_1}(\Delta_1^k)\rho_1(P_1^k(f))\}_{\gamma_1 \in \Gamma_1} $ obtained in the first layer.
	Here we have introduced the \textbf{one-step scattering propagator} $U_n[\gamma_n]: \ell^2(G^k_{n-1}) \rightarrow \ell^2(G^k_{n})$ mapping $f \mapsto g_{\gamma_n}(\Delta_n)\rho_n(P_n(f))$  as well as the \textbf{output generating operator} $V_{n}: \ell^2(G^k_{n-1})\rightarrow \ell^2(G^k_n)$ mapping $f$ to $\chi_n(\Delta^k_n)\rho_n(P^k_n(f))$.  Upon defining the \textbf{set} $\Gamma^{N-1} := \Gamma_{N-1} \times...\times\Gamma_1$ \textbf{of paths} of length $(N-1)$  terminating in layer $N-1$ (with $\Gamma^0$ taken to be the one-element set) and iterating the above procedure, we see that the outputs generated in the $N^{\text{th}}$-layer  are indexed by paths $\Gamma^{N-1}$ terminating in the previous layer.
	
\end{minipage}

We denote the resulting feature map by $\Phi_N^k$
and write $\mathscr{F}^k_N$ for the corresponding feature space. The node-level stability
results of the preceding sections then readily translate to higher order scattering transforms.
%

 As the respective proofs are identical to the corresponding results for the node setting, we do not repeat them here.

\begin{Thm}
	With the notation of Section \ref{clubbedtodeath}, we have for all $f,h\in\ell^2(G^k)$:
	\begin{align}
	\|\Phi^k_N(f) - \Phi^k_N(h)\|^2_{\mathscr{F}^k_N} \leq \left(1+ \sum\limits_{n=1}^N\max\{[B_n-1],[B_n(L_n^+R_n^+)^2-1],0\}\prod\limits_{\ell=1}^{n-1}B_\ell\right)\|f-h\|_{\ell^2(G^k)}^2
	\end{align}
\end{Thm}

\begin{Thm}\label{ktwo}
	Let $\Phi_N$ and $\widetilde{\Phi}_N$ be two  scattering transforms based on the same module sequence $\Omega_N$ and operator sequences
	$\mathscr{D}^k_N,\widetilde{\mathscr{D}}^k_N$  with the same connecting operators ($P^k_n = \widetilde P^k_n$) in each layer. Assume $R^+_n,L^+_n\leq 1$ and $B_n\leq B$ for some $B$ and $n \leq N$. Assume that the respective normal operators satisfy $\|\Delta^k_n - \widetilde \Delta^k_n\|_F \leq \delta$ for some $\delta >0$. Further assume that  the  functions $\{g_{\gamma_n}\}_{\gamma_n \in \Gamma_n}$ and $\chi_n$ in each layer  are Lipschitz continuous with associated Lipschitz constants
	satisfying $L_{\chi_n}^2+\sum_{\gamma_n \in \Gamma_n} L_{g_{\gamma_n}}^2\leq D^2$ for all $n \leq N$ and some $D>0$. Then we have for all $f\in \ell^2(G^k)$
	\begin{equation}
\|\widetilde \Phi_N^k(f)  -\Phi^k_N(f) \|_{\mathscr{F}_N} \leq  \sqrt{2(2^{N}-1)  }\cdot\sqrt{(\max\{B,1/2\})^{N - 1}} \cdot D \cdot \delta \cdot	 \|f\|_{\ell^2(G^k)}
\end{equation}
\end{Thm}

\begin{Thm}\label{k3surface}
	Let $\Phi_N^k, \widetilde{\Phi}_N^k$ be higher order scattering transforms based on a common module sequence $\Omega_N$ and differing operator sequences $\mathscr{D}^k_N, \widetilde{\mathscr{D}}^k_N$.  Assume $R^+_n,L^+_n\leq 1$ and $B_n\leq B$ for some $B$ and $n \geq 0$.
	Assume that there are identification operators $J_n: \ell^2(G^k_n) \rightarrow \ell^2(\widetilde G^k_n) $, $\widetilde{J}_n: \ell^2(\widetilde{G}^k_n) \rightarrow \ell^2( G^k_n) $ ($0\leq n \leq N$) so that the respective signal spaces are $\delta$-unitarily equivalent, the respective normal operators $\Delta^k_n,\widetilde{\Delta}^k_n$ are $\omega$-$\delta$-close as well as bounded (in norm) by $K>0$ and the connecting operators satisfy $\|\widetilde P^k_nJ_{n-1}f - J_n P^k_n f\|_{\ell^2(\widetilde G^k_n)} \leq \delta \|f\|_{\ell^2( G^k_{n-1})}$. For the common module sequence $\Omega_N$ assume  that the non-linearities  satisfy $\| \rho_n(J_{n}f) - J_n \rho_n(f)\|_{\ell^2(\widetilde G^k_n)} \leq \delta \| f\|_{\ell^2(G^k_n)}$ and that the constants $C_{\chi_n}$ and $\{C_{g_{\gamma_n}}\}_{\gamma_n \in \Gamma_N}$	associated through Lemma \ref{cpl} to the functions of the generalized frames in each layer satisfy $ C^2_{\chi_n} + \sum_{\gamma_n\in \Gamma_N} C^2_{g_{\gamma_n}} \leq D^2$ for some $D > 0$. Denote the operator that the family $\{J_n\}_n$ of identification operators induce on $\mathscr{F}^k_N$ through concatenation by
	$\mathscr{J}_N: \mathscr{F}^k_N \rightarrow \widetilde{\mathscr{F}}^k_N$. 
	Then we have with  $K_N = \sqrt{ (8^N-1) (2 D^2 +12B)/7\cdot B^{N-1}}$ if $B > 1/8$  and $K_N = \sqrt{  (2 D^2 +12B)\cdot (1 - B^N)/(1 - B)} $ if $B \leq 1/8$  that
	\begin{align}
	\|\widetilde{\Phi}^k_N(J_0f) - \mathscr{J}_N \Phi^k_N(f) \|_{\widetilde{\mathscr{F}}^k_N}	\leq K_N  \cdot \delta \cdot \|f\|_{\ell^2(G},\ \  \ \forall f \in \ell^2(G^k).
	\end{align}
	If additionally $\|\widetilde P^k_nJ_{n-1}f - J_n P^k_n f\|_{\ell^2(\widetilde G_n)} = 0$ or $ \| \rho_n(J_{n}f) - J_n \rho_n(f)\|_{\ell^2(\widetilde G^k_n)} = 0$ holds in each layer, then we have  $K_N = \sqrt{ (4^N-1) (2 D^2 +4B)/3\cdot B^{N-1}}$ if $B > 1/4$  and $K_N = \sqrt{  (2 D^2 +4B)\cdot(1 - B^N)/(1 - B)} $ if $B \leq 1/4$. If both additional equations hold, we have  $K_N = \sqrt{ (2^N-1) 2 D^2 \cdot B^{N-1}}$ if $B > 1/2$  and $K_N = \sqrt{  2 D^2 \cdot (1 - B^N)/(1 - B)} $ if $B \leq 1/2$. 
\end{Thm}

The map $N^G_p$ introduced in (\ref{NGP}) can also be adapted to aggregate higher-order tensorial features into graph level features:
With 
\begin{equation}
\|f\|_q := \left(\sum_{i_1,...,i_{k} \in G}|f_{i_1,...,i_{k} }|^q \mu_{i_1,...,i_k}\right)^{1/q} 
\end{equation}

 and $\mu_{G^k}:= \sum_{i_1...i_k=1}^{|G|}\mu_{i_1,...,i_k}$, we define 
\begin{equation}
N^{G^k}_p(f) = (\|f\|_{\ell^1(G^k)}/\sqrt{\mu_{G^k}},\|f\|_{\ell^2(G^k)},\|f\|_{\ell^3(G^k)},...,\|f\|_{\ell^p(G^k)})^\top/\sqrt{p}.
\end{equation}
Given a feature map $\Phi^k_N$ with feature space 
\begin{equation}
	 \mathscr{F}_N = \oplus_{n=1}^N \left(  \ell^2(G^k_n)  \right)^{|\Gamma^{n-1}|},
	 \end{equation} 
	 we  obtain a corresponding map $\Psi^k_N$ mapping from $\ell^2(G^k)$ to 
	 \begin{equation}
	 	\mathscr{R}_N = \oplus_{n=1}^N \left(  \mathds{R}^{p_n} \right)^{|\Gamma^{n-1}|}
	 	\end{equation}
	 	 by concatenating  $\Phi^k_N$ with the map that the family of non-linear maps $\{  N^{p_n}_{G^k_n} \}_{n=1}^N$  induces on $\mathscr{F^k}_N$ by concatenation. The resulting map $\Psi^k_N$ again has  stability properties analogous to the node level case:

\begin{Thm}
	Assuming admissibility, we have 
	\begin{equation}
	\|\Psi^k_N(f) - \Psi^k_N(h)\|_{\mathscr{R}_N} \leq \left(1+ \sum\limits_{n=1}^N\max\{[B_n-1],[B_n(L_n^+R_n^+)^2-1],0\}\prod\limits_{\ell=1}^{n-1}B_\ell\right)\|f-h\|_{\ell^2(G^k)}^2
	\end{equation} 
	 for all $f,h \in \ell^2(G)$ .
	With the conditions and notation of Theorem \ref{ktwo} we have
	\begin{equation}
	\|\Psi^k_N(f)  -\widetilde{\Psi}^k_N(f) \|_{\mathscr{R}_N} \leq \sqrt{2(2^{N}-1)  }\cdot\sqrt{(\max\{B,1/2\})^{N - 1}} \cdot D \cdot \delta \cdot	 \|f\|_{\ell^2(G^k)}.
	\end{equation}
	Additionally,  in the setting of Theorem \ref{k3surface}, assuming that for each $n \leq N$ the identification operator $J_n$ satisfies $
	\big| \|J_nf\|_{\ell^1(\widetilde G^k_n)}/\sqrt{\mu_{\widetilde G^k_n}} - \|f\|_{\ell^1( G^k_n)}/\sqrt{\mu_{ G^k_n}}\big| ,\big| \|J_nf\|_{\ell^r(\widetilde G^k_n)} - \|f\|_{\ell^r( G^k_n)}\big| \leq \delta \cdot K \cdot \|f\|_{\ell^2( G^k_n)} $  for $2 \leq r \leq p_n
	$
	implies ($\forall f\in \ell^2(G^k)$)
	\begin{align}
	\|\widetilde{\Psi}_N(J_0f) -  \Psi_N(f) \|_{\mathscr{R}_N}	\leq \sqrt{2}\cdot\sqrt{K^2_N  + K^2} \cdot \delta \cdot \|f\|_{\ell^2(G^k)}.
	\end{align}
\end{Thm}

As the proofs here are virtually the same as for the corresponding results in previous sections -- essentially only replacing $G$ by $G^k$, we omit a repetition of them here.

\section{Additional Details on Experiments}\label{AddEx}
Here we provide additional details on utilized scattering architectures, training procedures, datasets and (performance of) other methods our approach is being compared to. Irrespective of task, our models are trained on an NVIDIA DGX A100 architecture utilizing between two and eight NVIDIA Tesla A100 GPUs with 80GB memory each. Running $10$-fold cross validation for the respective experiments took at most $71$ hours (which was needed for social network graph classification on REDDIT-$12$K).

\subsection{Social Network Graph Classification}
\paragraph{Datasets:}

The data we are working with is taken from \cite{socialdatasets}. In particular this work introduced six social network datasets extracted from from scientific collaborations (COLLAB), movie collaborations (IMDB-B, IMDB-M) and Reddit discussion threads (REDDIT-B, REDDIT-$5$K, REDDIT-$12$K). Data is anonymised and contains no content that might be considered offensive. Each graph carries a class label, and the goal is to predict this label. Some basic properties of these datasets are listed in Table \ref{sonchar} below.

\begin{table}[h!]
	\begin{center}
		\scalebox{0.85}{
			\begin{tabular}{ p{3cm}|p{1.5cm}p{1.5cm}p{1.5cm}p{1.8cm}p{1.9cm}p{2.0cm}  }
				\hline
				Attributes: 			& COLLAB 	&IMDB- B&IMDB-M&		 REDDIT-B & REDDIT-5K&REDDIT-12K\\
				\hline
				Graphs   & 5K  		&1K&   1.5K& 		2K	&	5K &12K\\
				Nodes&  			 372.5K    &19.8K  &19.5K&		859.2K	&2.5M &4.7M\\
				Edges &				49.1M 		& 386.1K&  395.6&		4M	& 11.9M & 21.8M\\
				Maximum Degree    &2k 			& 540&  352& 		12.2K	&8K &12.2K\\
				Minimum Degree	&   4 			 & 4&	4& 			4	&4 & 4\\
				Average Degree	& 263  			& 39  &40&			9 	&9 & 9\\
				Target Labels	& 3  			& 2  &3&			2 	&5 & 11\\
				Disconnected  Graphs & No			& No  &No&			Yes 	&Yes & Yes\\
				\hline
			\end{tabular}
		}
	\end{center}
	\caption{Social Network Dataset Characteristics}
	\label{sonchar}
\end{table}

\ \\
These datasets contain graph structures, however they don't contain associated weights or graph signals. Having unspecified weights simply means that the adjacency matrix $W$ from which we construct the  graph Laplacian
\begin{equation}
\mathcal{L} = D - W
\end{equation}
on which our operator $\Delta$ is based simply has each entry corresponding to an edge set to unity. If no edge is present between vertices $i$ and $j$, the entry $W_{ij}$ is set to zero. 
It remains to solve the problem of the missing input signals. Our strategy is to generate signals reflecting the geometry of the underlying graph. We do this by utilizing features that associate to each node a number that characterizes its role or importance within its local environment or within the entire graph.
We briefly describe them here:
\begin{enumerate}
	\item \underline{\textbf{Degree}}: The degree of a node is the number of edges incident at this node.
	\item \underline{\textbf{Eccentricity}}: For a connected graph, the eccentricity of a node is the maximum distance from this node to all other nodes. On a disconnected graph it is not defined.
	\item \underline{\textbf{Clustering}}: For unweighted graphs the clustering  $c(u)$ of a node $u$ is the fraction of possible triangles through that node that actually exist. It is calculated as
	\begin{equation}
	c(u) = \frac{2T(u)}{\text{deg}(u)(\deg(u)-1)}.
	\end{equation}
	\item \underline{\textbf{Number of triangles}}: The number of triangles containing the given node as a vertex.
	\item \underline{\textbf{Core number}}: A k-core is a maximal subgraph that only contains nodes of degree k or more.	
	The core number of a node is the largest value k of a k-core containing that node.
	\item \underline{\textbf{Clique number}}:  A clique is a subset of vertices of an undirected graph such that every two distinct vertices in the clique are adjacent. This input assigns the number of cliques the nodes participates in to each node.
	\item \underline{\textbf{Pagerank}}: This returns the PageRank of the respective nodes in the graph.	
	PageRank computes a ranking of the nodes in the graph based on the structure of the edges. Originally it was designed as an algorithm to rank web pages.
\end{enumerate}
For the first three datasets listed in Table \ref{sonchar} we utilize all listed input features. For the latter three datasets we have to refrain from using eccentricity as an input signal, as these datasets contain graphs that have multiple non-connected graph components. 

\paragraph{Scattering Architecture:}
We chose a generalized scattering architecture of depth $N=4$. As normal-operators, we utilize in each layer the un-normalized graph Laplacian $\mathcal{L} = D - W$ scaled by its largest eigenvalue ($\Delta = \mathcal{L}/\lambda_{\textit{max}}(\mathcal{L})$).  Filters are chosen as  $\frac12(\sin(\pi/2\cdot \Delta), [\cos(\pi/2\cdot \Delta) - \psi_\Delta\psi_\Delta^\top], \sin(\pi\cdot \Delta), [\cos(\pi\cdot \Delta) - \psi_\Delta\psi_\Delta^\top]) $,
  which allows to specify the output generating function solely by demanding $\chi(0)=1$ and $\chi(\lambda) = 0$ on all other eigenvalues of $\Delta$.
 Here $\psi_\Delta$ is the normalized vector of all ones (satisfying $\Delta\psi_\Delta=0$). Connecting operators are chosen as the identity, while we set $\rho_{n\geq 1}(\cdot)= |\cdot|$.  We note that for connected graphs, this recovers Architecture I of Fig. \ref{BothArchitectures}. On disconnected graphs (as they can appear in the REDDIT datasets), we however do not account for  vectors other than $\psi_\Delta$ in  the lowest-lying eigenspace of the graph Laplacian.
This scattering architecture is then applied to each of these input signal individually. For each input signal, this returns a feature vector with $1+4+16+64 = 85$ entries. These individual feature vectors are then concatenated into one final composite feature vector for each graph. Concerning applicable theoretical results, we note the following:

\paragraph{Training Procedure:}
%
%
%
%
%
We train RBF kernel support vector classifiers on our composite scattering features.
We fix  $\epsilon = 0.1$. The hyperparameter $\gamma$ scaling the exponent is chosen from 
\begin{equation}
G_{\text{pool}} := \{0.00001, 0.0001, 0.001, 0.01, 0.1, 1, 10, 100\},
\end{equation}
while we pick the  $C$ that controls  the error our slack variables introduce among
\begin{equation}
C_{\text{pool}} := \{0.001, 0.01, 0.1, 1, 10, 25, 50, 100, 1000\}.
\end{equation}
We chose these parameters in agreement with the choices of \cite{gao2019geometric} to facilitate comparison between the two works.\\
We could simply implement the training of the RBF-classifier on our composite scattering features by dividing each social-network dataset into $10$ folds, then iteratively choosing one fold for testing and among the remaining $9$ folds randomly choosing one for validation (i.e. for tuning the hyperparameters). 
Instead, following \cite{gao2019geometric} (whose code is released under an Apache license and on which we partially built), we take a slightly different approach: We still randomly split our dataset into $10$ folds. Among the 10 folds, we iteratively pick one for testing. Say we have picked the $n^{\text th}$ fold for testing. Then there are $9$ remaining folds. We iteratively pick the $m_n^{\text th}$ (with $1 \leq m_n \leq 9$) of the remaining $9$ folds for choosing hyperparameters. This leaves $8$ folds on which we train our model for each choice of hyper parameter  in $C_{\text{pool}} \times G_{\text{pool}}$. The resulting classifiers are all evaluated on the $m_n^{\text th}$ fold. The one that performs best is retained as classifier $m_n$. As $m_n$ varies between $1$ and $9$ (still for fixed $n$), this yields a set  $\{f_{m_n}:\ 1\leq m_n\leq 9\}$ of nine classifiers. From these we build the classifier $f_n$, whose classification result is obtained from a majority vote among the nine classifiers in $\{f_{m_n}:\ 1\leq m_n\leq 9\}$.  Then we evaluate the performance of $f_n$ on the $n^{\text th}$ fold to obtain the $n^{\text th}$ estimation of how well our  model performs. As $n$ varies from one to ten, we built the mean and variance of the performances of the classifiers $f_n$ on the $n^{\text th}$ fold expressed as the percentage of correct classifications.

\paragraph{Reference Methods:}
To allow for a comparison of our results to the literature, typical classification accuracies for graph algorithms on social network datasets are displayed in Table \ref{classify}. Following the standard format of reporting classification accuracies, they are presented in the format (Accuracy $\pm$ standard deviation). If results are not reported for a dataset, we denote this as not available (N/A). The first three rows of Table \ref{classify} display results for graph kernel methods; namely Weisfeiler-Lehman graph kernels (WL, \cite{WL}), Graphlet kernels (Graphlet, \cite{Graphlet})
 and deep graph kernels (DGK, \cite{DGK}). The subsequent rows display results for geometric deep
learning algorithms: Deep graph convolutional neural networks (DGCNN,\cite{DGCNN}),
  Patchy-san (PSCN (with k=10), \cite{PSCN}), 
   recurrent neural network autoencoders  (S2S-N2N-PP, \cite{S2S}) and graph isomorphism networks (GIN \cite{GIN}). These  results are taken from \cite{gao2019geometric}. 
  Additionally we compare with P-Poinc \cite{kyriakis2021learning}, which embeds nodes into a hyperbolic space (the Poincare ball, to be precise),  GSN-e \cite{bouritsas2022improving} which combines message passing with structural features  extracted via subgraph isomorphism and WKPI-kC \cite{zhao2019learning} which utilizes a weighted kernel within its metric learning framework.
The second to last row (GS-SVM \cite{gao2019geometric}) provides a result that is also based on a method that combines a static scattering architecture with a support vector machine. Its filters are based on graph wavelets built from differences between lazy random walks that have propagated at different time scales.

\subsection{Regression of Quantum Chemical Energies}

\paragraph{Dataset:}
The dataset we consider is the QM$7$ dataset, introduced in \cite{blum, rupp}. This dataset contains descriptions of $7165$ organic molecules, each with up to seven heavy atoms, with all non-hydrogen atoms being considered heavy. A molecule is represented by its Coulomb matrix $C^{\text{Clmb}}$, whose off-diagonal elements
\begin{equation}\label{offdiag}
C^{\text{Clmb}}_{ij}	 = \frac{Z_iZ_j}{|R_i-R_j|}
\end{equation}
correspond to the Coulomb-repulsion between atoms $i$ and $j$, while diagonal elements
encode a polynomial fit of atomic energies to nuclear charge \cite{rupp}:
\begin{equation}
C^{\text{Clmb}}_{ii} = \frac12 Z_i^{2.4}
\end{equation}

For each atom in any given molecular graph, the individual Cartesian coordinates $R_i$ and the atomic charge $Z_i$ are also accessible individually. 
To each molecule an atomization energy - calculated via density functional theory - is associated. The objective is to predict this quantity, the performance metric is mean absolute error. Numerically, atomization energies are negative numbers in the range $-600$ to $-2200$. The associated unit is $[\textit{kcal/mol}]$.\\

\paragraph{Scattering Architecture:}
Off-diagonal entries in the Coulomb Matrix clearly represent an inverse distance. A weight of zero can then heuristically be thought of as the inverse distance between two infinitely separated atoms. After calculating the degree matrix $D$ associated to $C$, we obtain the graph Laplacian once more as $\mathcal L = D - C $ and set our normal operator to
\begin{equation}
\Delta = \frac{\mathcal{L}}{\lambda_{\text{max}}(\mathcal{L})}.
\end{equation}
If we continuously vary the distances in (\ref{offdiag}), staying clear of zero,  then the adjacency matrix and hence the graph Laplacian $\mathcal L$  varies continuously. As long as we avoid complete degeneracy, the largest eigenvalue $\lambda_{\text{max}}(\mathcal L)$ will remain positive. This implies that our normal operator $\Delta$ varies continuously under changes of the inter-atomic distances, which implies that our feature vector also varies continuously, as distances are changed.
%
%
Connecting operators are set to the identity, while non-linearities are fixed to $\rho_{n \geq 1}(\cdot) = |\cdot|$.  Filters are chosen as  $(\sin(\pi/2\cdot \Delta),$ $ \cos(\pi/2\cdot \Delta) , \sin(\pi\cdot \Delta), \cos(\pi\cdot \Delta))  $ acting 
through matrix multiplication. The output generating functions are set to the identity as well. Graph level features are aggregated via the map $N^{E}_5(\cdot)$ of Section \ref{HOS}; slightly modified to neglect the normalizing factor in the first entry for improved convenience in numerical implementability.
As weights $\mu_{ij}$ for our second-order feature space are set to unity and molecular graphs in QM$7$ contain at most $23$-molecules, we note that  $\sqrt{\mu_{G^2}}\leq\sqrt{23^2}=23$. Going through the proofs of our graph-level stability results, we see that they remain valid after multiplying each
stability constant by $23$. The Coulomb matrix (divided by a factor of $10$ as this empirically improved performance) is then also utilized as an edge level input signal.
Node level features are obtained by applying the above architecture to the node level information provided by the respective atomic charges $\{Z_i\}$ on each graph. We aggregate to graph level features using $N^G_5$ (cf. Section \ref{aggro1}), again neglecting the normalizing factor in the first entry for improved convenience in implementing. The network depth is set to $N=4$ in both cases. We then concatenate graph level features obtained from node- and edge level input into a composite scattering feature vector.

\paragraph{Training Procedure:}

The QM$7$ dataset comes with a precomputed partition into five subsets; each containing a representative amount of heavy and light molecules covering the entire complexity range of  QM$7$. To allow for $10$-fold cross validation, we further dissect each of these subsets into two smaller datasets, one containing graphs indexed by an even number, one containing graphs indexed by an odd number. On these $10$-subsets, we then perform 10-fold cross validation. 
Among the 10 folds, we iteratively pick one for testing. Say we have picked the $n^{\text th}$ fold for testing. Then there are $9$ remaining folds. We iteratively pick the $m_n^{\text th}$ (with $1 \leq m_n \leq 9$) of the remaining $9$ folds for choosing hyperparameters. This leaves $8$ folds on which we train our model for each choice of hyper parameter  in $C_{\text{pool}} \times G_{\text{pool}}$. This yields $8$ regression models, which we average to built our final predictor for the $n^{\text th}$ run. This mean absolute error of this predictor is then evaluated on the  $n^{\text th}$ fold which was retained for testing. As $n$ varies from one to ten, we built the mean and variance of the performances of the generated regression models. 
We chose $\log$-linear equidistant hyperparameters  from

\begin{equation}
G_{\text{pool}} := \{0.00003,0.0003,0.003,0.03,0.3,3,30\},
\end{equation}
and
\begin{equation}
C_{\text{pool}} := \{400000,40000,4000,400,40,4,0.4\}.
\end{equation}

%
%

\paragraph{Reference Methods:}

We comprehensively evaluate our method against $11$ popular baselines and state of the art approaches. Among these methods are graph convolutional methods such as  GraphConv \cite{Kipf}, Weave \cite{kearnes2016molecular} or SchNet \cite{sch}.
MPNN \cite{mpnncm}  and its variant DMPNN \cite{ALMR} are models considering edge features during message passing. AttentiveFP \cite{pushing} is an extension of the graph attention framework, while N-Gram \cite{ngram} is a pretrained method. Results for these methods as well as for GROVER are taken from \cite{grover}.
PhysChem \cite{quantmet} learns molecular representations
via fusing physical and chemical information. Deep Tensor Neural Networks (DTNN \cite{molnet}) are adaptable extensions of the Coulomb Matrix featurizer mapping atom numbers to trainable embeddings which are then updated based on distance information and other (node-level) atomic features. Finally Path-Augmented Graph Transformer Networks (PATGN, \cite{patgn}) exploit the connectivity structure of the data in a global attention mechanism.

\end{document}